%% file: iclr2025_conference.tex
\documentclass{article} %
\usepackage{iclr2025_conference,times}
\iclrfinalcopy

\input{math_commands.tex}

\usepackage{hyperref}
\usepackage{url}

\usepackage{enumitem}
\usepackage{amsmath}
\usepackage{amsfonts}
\usepackage{amssymb}

\usepackage{graphicx}
\usepackage{multirow}
\usepackage{tabularx}
\usepackage{subcaption}
\usepackage{comment}
\usepackage{colortbl}
\usepackage{cleveref}
\usepackage{xcolor}
\usepackage{booktabs}
\usepackage{url}
\usepackage{natbib}
\usepackage{wrapfig}
\usepackage[most]{tcolorbox}
\usepackage{enumitem}
\usepackage{inconsolata}

\usepackage{booktabs}
\usepackage{pifont}
\usepackage{makecell}

\newcommand{\cmark}{\ding{51}}%
\newcommand{\xmark}{\ding{55}}%

\definecolor{redT}{RGB}{255,220,220}
\definecolor{yellowT}{RGB}{255,250,210}
\definecolor{greenT}{RGB}{220,245,220}

\usepackage[most]{tcolorbox} %
\usepackage{xcolor}
\usepackage{inconsolata}     %
\usepackage{pifont} %

\usepackage{listings}
\usepackage{xcolor}

\definecolor{codegray}{gray}{0.95}

\lstdefinestyle{pythonstyle}{
  language=Python,
  basicstyle=\ttfamily\small,
  backgroundcolor=\color{codegray},
  frame=single,
  rulecolor=\color{black},
  numbers=left,
  numberstyle=\tiny,
  stepnumber=1,
  showstringspaces=false,
  breaklines=true,
  tabsize=2,
}

\lstnewenvironment{pythoncode}
  {\lstset{style=pythonstyle}}
  {}

\definecolor{UserColor}{HTML}{2563EB} %
\definecolor{AsstColor}{HTML}{059669} %
\definecolor{FrameGray}{HTML}{D1D5DB} %
\definecolor{TitleText}{HTML}{111827} %

\newcommand{\personablock}[1]{%
  \begin{tcolorbox}[
    enhanced, breakable, colback=white, colframe=FrameGray,
    boxrule=0.5pt, sharp corners,
    title={\textbf{\color{TitleText}Persona}}
  ]
  \ttfamily\footnotesize
  #1
  \end{tcolorbox}
}

\newcommand{\usermsg}[2]{%
  \noindent
  \begin{tcolorbox}[enhanced,breakable,
    colback=UserColor!6,colframe=UserColor!60!black,
    boxrule=0.5pt,sharp corners,
    width=\dimexpr0.9\textwidth\relax,
    left=2mm,right=2mm,top=1mm,bottom=1mm,
    title={\textbf{User}\hfill{#1}},
    before skip=0.6ex, after skip=0.6ex,
  ]
    \ttfamily\footnotesize
    #2
  \end{tcolorbox}
}

\newcommand{\assistantmsg}[2]{%
  \begin{flushright}
  \begin{tcolorbox}[enhanced,breakable,
    colback=AsstColor!6,colframe=AsstColor!60!black,
    boxrule=0.5pt,sharp corners,
    width=\dimexpr0.95\textwidth\relax,
    left=2mm,right=2mm,top=1mm,bottom=1mm,
    title={\textbf{Assistant: }\textbf{#1}},
    before skip=0.6ex, after skip=0.6ex]
    \ttfamily\footnotesize
    #2
  \end{tcolorbox}
  \end{flushright}
}

\definecolor{baselineColor}{HTML}{4B5563} %
\definecolor{discoveryColor}{HTML}{2563EB} %
\definecolor{oracleColor}{HTML}{059669} %
\definecolor{modeRoleColor}{HTML}{111827} %

\definecolor{BaselineColor}{HTML}{62727B}
\definecolor{DiscoveryColor}{HTML}{2A79C5}
\definecolor{OracleColor}{HTML}{7B2CBF}

\newcommand{\Correct}{\textcolor{green!60!black}{\ding{51}\,Correct}}
\newcommand{\Incorrect}{\textcolor{red!70!black}{\ding{55}\,Incorrect}}

\newcommand{\BaselineHeader}[2]{%
  \begin{tcolorbox}[enhanced,breakable,boxrule=0pt,frame hidden,
    colback=BaselineColor!9,left=6pt,right=6pt,top=4pt,bottom=4pt]
    \sffamily\bfseries Baseline Mode
    \hfill \normalsize \textsf{\metricname Score: }\textbf{#1}\quad \textsf{Answer Correctness: }#2
  \end{tcolorbox}
}
\newcommand{\DiscoveryHeader}[2]{%
  \begin{tcolorbox}[enhanced,breakable,boxrule=0pt,frame hidden,
    colback=DiscoveryColor!9,left=6pt,right=6pt,top=4pt,bottom=4pt]
    \sffamily\bfseries Discovery Mode
    \hfill \normalsize \textsf{\metricname Score: }\textbf{#1}\quad \textsf{Answer Correctness: }#2
  \end{tcolorbox}
}
\newcommand{\OracleHeader}[2]{%
  \begin{tcolorbox}[enhanced,breakable,boxrule=0pt,frame hidden,
    colback=OracleColor!9,left=6pt,right=6pt,top=4pt,bottom=4pt]
    \sffamily\bfseries Oracle Mode
    \hfill \normalsize \textsf{\metricname Score: }\textbf{#1}\quad \textsf{Answer Correctness: }#2
  \end{tcolorbox}
}

\newcommand{\TripleCaseBDO}{%
  \begin{tcolorbox}[colback=green!6,colframe=green!45!black,boxrule=0.5pt,sharp corners]
    \sffamily\bfseries  \metricname scores:\; Baseline $<$ Discovery $<$ Oracle \hfill \small \textcolor{green!50!black}{\ding{115}} rising with personalization
  \end{tcolorbox}
}
\newcommand{\TripleCaseDBO}{%
  \begin{tcolorbox}[colback=yellow!7,colframe=orange!60!black,boxrule=0.5pt,sharp corners]
    \sffamily\bfseries 
    \metricname scores:\; Discovery $<$ Baseline $<$ Oracle
    \hfill 
    \hfill \small \textcolor{orange!60!black}{\ding{117}} Discovery degrades alignment.
  \end{tcolorbox}
}
\newcommand{\TripleCasePersFail}{%
  \begin{tcolorbox}[colback=red!6,colframe=red!60!black,boxrule=0.5pt,sharp corners]
    \sffamily\bfseries Outcome:\; Both personalized modes incorrect
    \hfill \small \textcolor{red!70!black}{\ding{55}} personalization failures
  \end{tcolorbox}
}
\newcommand{\TripleCaseBaseFail}{%
  \begin{tcolorbox}[colback=green!6,colframe=green!60!black,boxrule=0.5pt,sharp corners]
    \sffamily\bfseries Outcome:\; Baseline is incorrect; whereas personalized modes correct
    \hfill \small \textcolor{green!60!black}{\checkmark}\; personalization success
  \end{tcolorbox}
}
\newcommand{\ScoreBadge}[1]{%
  \ifnum#1=5 \textcolor{green!60!black}{\ding{51}}%
  \else\ifnum#1=1 \textcolor{red!70!black}{\ding{55}}%
  \else\ifnum#1=3 \textcolor{black!55}{\ding{108}}%
  \else \,%
  \fi\fi\fi
}

\newcommand{\tbASRColorFine}[1]{%
  \ifdim #1 pt<-40pt redT%
  \else\ifdim #1 pt<-30pt redT!85!yellowT%
  \else\ifdim #1 pt<-20pt redT!70!yellowT%
  \else\ifdim #1 pt<-10pt redT!55!yellowT%
  \else\ifdim #1 pt<-5pt  redT!35!yellowT%
  \else\ifdim #1 pt<5pt   yellowT%
  \else\ifdim #1 pt<10pt  yellowT!35!greenT%
  \else\ifdim #1 pt<20pt  yellowT!55!greenT%
  \else\ifdim #1 pt<30pt  yellowT!70!greenT%
  \else\ifdim #1 pt<40pt  yellowT!85!greenT%
  \else                   greenT%
  \fi\fi\fi\fi\fi\fi\fi\fi\fi\fi
}

\DeclareRobustCommand{\heat}[1]{\cellcolor{\tbASRColorFine{#1}}\strut #1}

\usepackage{xspace}
\newcommand{\methodname}{\textsc{PrefDisco}\xspace}
\newcommand{\metricname}{\textsc{PrefAlign}\xspace}
\newcommand{\capabilityname}{personalized reasoning\xspace}

\definecolor{forestgreen}{rgb}{0.13,0.55,0.13}

\newcommand{\ScoreIcon}[1]{%
  \ifnum#1=5 \textcolor{green!60!black}{\ding{51}}%
  \else\ifnum#1=1 \textcolor{red!70!black}{\ding{55}}%
  \else\ifnum#1=3 \textcolor{black!55}{\ding{108}}%
  \else \textcolor{black!50}{\raisebox{0.2ex}{\tiny\(#1\)}}%
  \fi\fi\fi
}

\title{
\parbox{0.04\textwidth}{\includegraphics[width=\linewidth]{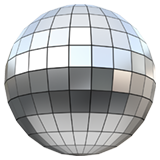}} 
\methodname:
Evaluating Proactive Personalization via Interactive Preference Discovery}

\title{
\parbox{0.04\textwidth}{\includegraphics[width=\linewidth]{figures/disco.png}} 
\methodname: Benchmarking Proactive \\
Personalized Reasoning
}

\author{Shuyue Stella Li$^{1}$\thanks{Equal contribution.}\;, Avinandan Bose$^{1}$\footnotemark[1]\;, Faeze Brahman$^2$\;\\
\textbf{Simon Shaolei Du}$^1$\;, \textbf{Pang Wei Koh$^{1,2}$}\;, \textbf{Maryam Fazel$^1$}\;, \textbf{Yulia Tsvetkov$^1$} \\
$^1$University of Washington, $^2$Allen Institute for AI \\
\texttt{\{stelli,avibose\}@cs.washington.edu} \\
\parbox{0.03\textwidth}{\includegraphics[width=\linewidth]{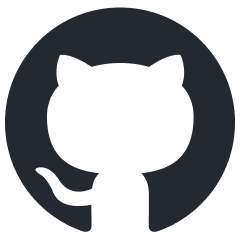}}\hspace{0.5mm}\href{https://github.com/stellalisy/PrefDisco}{\texttt{https://github.com/stellalisy/PrefDisco}}\\
\parbox{0.033\textwidth}{\includegraphics[width=\linewidth]{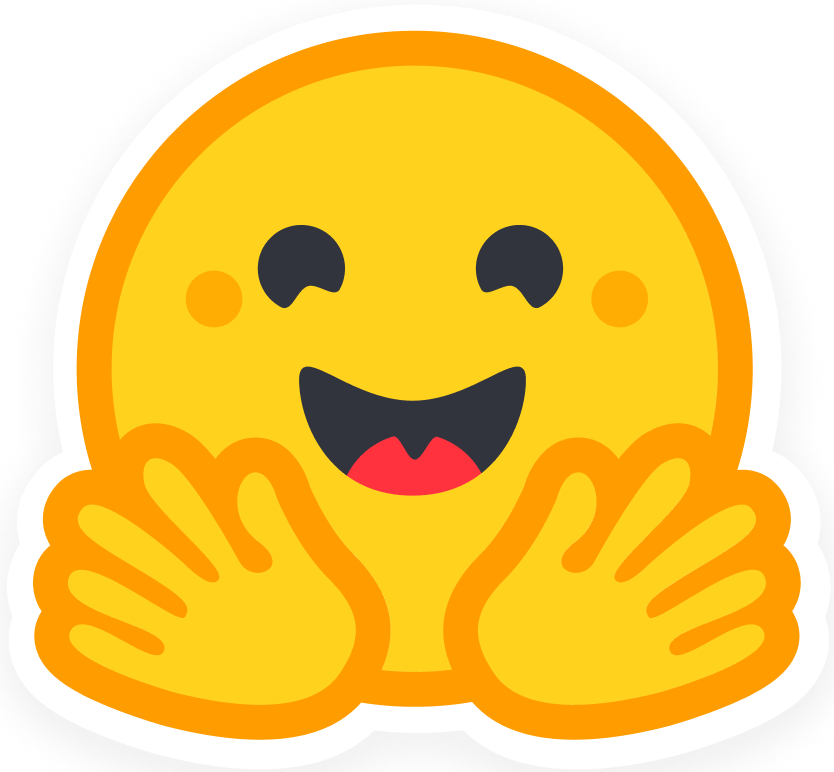}}\hspace{0.5mm}\href{https://huggingface.co/collections/stellalisy/personalized-reasoning-68adec5f11411e842281d385}{\texttt{https://huggingface.co/collections/stellalisy/personalized-reasoning}}\vspace{-3mm}
}

\begin{document}

\maketitle

\begin{abstract}

Current large language model (LLM) development treats task-solving and preference-alignment as separate challenges, optimizing first for objective correctness,  then for alignment to aggregated human preferences. 
This paradigm fails in human-facing applications where solving a problem correctly is insufficient if the response mismatches the user's needs.
This challenge intensifies in \emph{just-in-time} scenarios where no prior user interaction history exists due to cold-start conditions or privacy constraints. 
LLMs need to proactively identify what they don't know about the user, strategically elicit preference values through questioning, then adapt their reasoning processes and responses accordingly---a complicated chain of cognitive processes which we term \emph{\capabilityname}. %
We introduce \methodname, an evaluation methodology that transforms static benchmarks into interactive personalization tasks using psychologically-grounded personas with sparse, context-dependent preferences, and define \metricname as a fine-grained rubric-based metric for measuring preference alignment.
\methodname builds scenarios where identical questions require different reasoning chains depending on user context, as optimal explanation approaches vary by individual expertise and preferences while maintaining factual accuracy. 
Evaluation of 21 frontier models across 10 tasks reveals 29.0\% of naive personalization attempts produce worse preference alignment than generic responses, yet generic responses also fail to serve individual user needs. 
These findings suggest \capabilityname requires dedicated development rather than emerging naturally. %
\methodname provides a foundation for developing systems that can adapt to individual users in education, healthcare, and technical domains where personalization is critical. %

\end{abstract}

\begin{figure}[h]
    \centering%
    \includegraphics[width=0.98\linewidth]{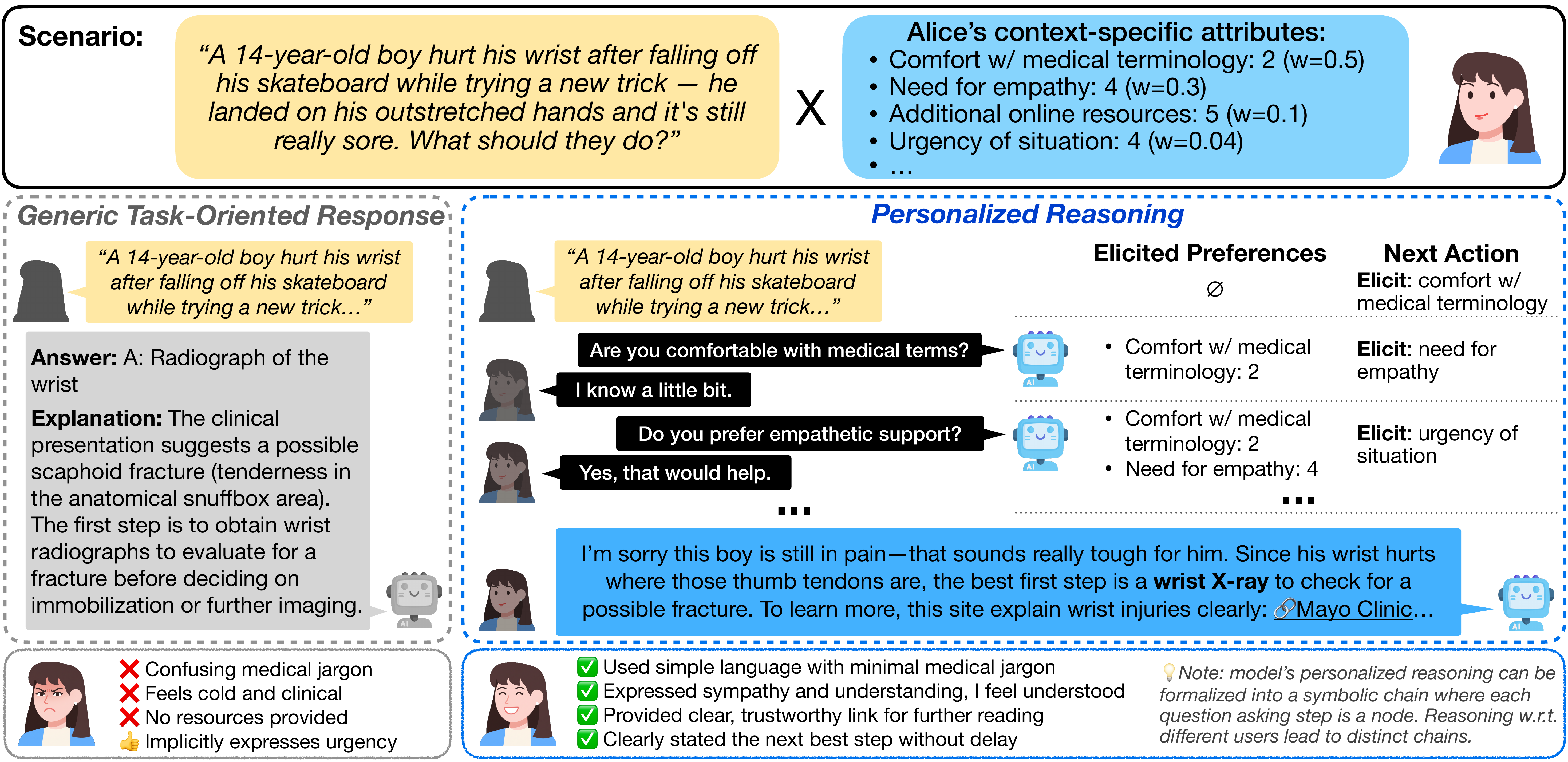}\vspace{-2.5mm}
    \caption{Personalized reasoning in a medical scenario. Current LLMs provide generic responses without considering the user (left); a model with personalized reasoning capabilities incorporates discovered preferences to provide responses that is both correct and aligned to the user (right).}\vspace{-5mm}
    \label{fig:framework}
\end{figure}

\section{Introduction}\vspace{-3mm}

Current large language model (LLM) development treats task-solving and preference alignment as sequential challenges: models are first optimized for objective correctness through instruction tuning or reinforcement learning \citep{longpre2023flan}, then aligned to aggregated human preferences through reinforcement learning from human feedback \citep{ouyang2022training,rafailov2024directpreferenceoptimizationlanguage}.
This paradigm fundamentally misaligns with human-AI interaction, where the task and the individual user are inseparable. 
For instance, %
a medical explanation benefits from clinical analogies for one trainee while another requires formal definitions, requiring different cognitive approaches to answer the same problem.
When models provide identical responses regardless of user context, they fail to serve individual needs despite achieving high benchmark performance.
This challenge intensifies in cold-start scenarios where no prior user interaction history exists due to privacy constraints or new user onboarding, requiring proactive ``just-in-time personalization'' capabilities that current systems lack. Moreover, users often cannot articulate their specific needs or provide effective feedback about response misalignment \citep{liu2025userfeedbackhumanllmdialogues}, necessitating that LLMs proactively elicit this information rather than placing the cognitive burden on users.

We define \textbf{\capabilityname} as the ability to adapt reasoning processes based on discovered user preferences. %
Consider the medical scenario in Figure~\ref{fig:framework}, discovering that the user Alice has limited medical knowledge and needs emotional support fundamentally changes the appropriate reasoning strategy the model should take: instead of focusing on justifying clinical diagnosis, the model needs to reason about how to best satisfy Alice's needs such as empathy. 
This goes beyond surface-level presentation; it requires different reasoning steps, different information prioritization, and different decision points about what to include or omit. 
A model with personalized reasoning capabilities must identify gaps in knowledge about user preferences, strategically elicit preference values through questioning, and synthesize this information to adapt both their reasoning processes and response generation.

Existing personalization research inadequately addresses interactive reasoning scenarios. Personalization benchmarks such as PersoBench \citep{afzoon2024persobench}, PrefEval \citep{zhao2025prefeval}, and PersonaMem \citep{jiang2025know} focus on content recommendation or dialogue generation with static user profiles, treating personalization as applying predetermined preferences to fixed outputs rather than adapting the underlying reasoning approach. 
Interactive frameworks like MediQ \citep{li2024mediq} demonstrate questioning capabilities but target clinical information-seeking without personalization objectives.
Most critically, no existing work recognizes that effective personalization requires different reasoning processes for different users; current approaches assume reasoning processes remain constant while only presentation varies. 
We address this conceptual gap by introducing \textbf{the first evaluation framework requiring models to proactively discover user preferences and adapt their reasoning processes accordingly}, recognizing that the cognitive steps needed to solve problems should themselves be user-dependent.

\begin{table*}[t]
\centering
\small
\caption{
Comparison of personalization benchmarks across. 
\methodname is the only benchmark that combines instance-specific rubric evaluation, 
proactive discovery of latent user preferences, true cold-start personalization, 
and evaluation on verifiable reasoning tasks across multiple domains.
}\label{tab:personalization-benchmark-comparison}\vspace{-2mm}
\renewcommand{\arraystretch}{1.25}
\resizebox{\linewidth}{!}{
\begin{tabular}{
    p{2cm}
    p{2.1cm}
    p{2.1cm}
    p{2.1cm}
    p{2.1cm}
    p{2.1cm}
}
\toprule
\textbf{Benchmark} &
\textbf{Rubric-based eval. (instance-specific)} &
\textbf{Proactivity in discovering latent prefs} &
\textbf{Cold-start / just-in-time personalization} &
\textbf{Uses verifiable tasks} &
\textbf{Multi-task breadth} \\
\midrule

\textbf{PersoBench} &
\textcolor{red}{\xmark{}} generic persona scoring; no instance rubrics &
\textcolor{red}{\xmark{}} preferences provided, no discovery &
\textcolor{red}{\xmark{}} persona given; no cold-start &
\textcolor{red}{\xmark{}} subjective dialogue only &
\textcolor{orange}{limited} (dialogue) \\

\textbf{PrefEval} &
\textcolor{red}{\xmark{}} LLM-judged; no attribute rubrics &
\textcolor{red}{\xmark{}} prefs in context; no discovery &
\textcolor{red}{\xmark{}} requires long history &
\textcolor{red}{\xmark{}} MCQs depend on prefs, not ground truth &
\textcolor{orange}{moderate} conversational tasks\\

\textbf{PersonaMem} &
\textcolor{red}{\xmark{}} persona-state matching; no rubrics &
\textcolor{red}{\xmark{}} traits from logs; no discovery &
\textcolor{red}{\xmark{}} multi-session history needed &
\textcolor{red}{\xmark{}} MCQs tied to persona, not objective truth &
\textcolor{orange}{moderate} broad persona tasks \\

\textbf{UserBench} &
\textcolor{red}{\xmark{}} scenario-level evaluation; no rubrics &
\textcolor{orange}{partial} (intent clarification only) &
\textcolor{red}{\xmark{}} underspecified but not true cold-start &
\textcolor{red}{\xmark{}} “correct” depends on preferences &
\textcolor{orange}{limited} single domain (travel) \\

\textbf{\methodname(Ours)} &
\textcolor{forestgreen}{\cmark{}} instance-specific attribute rubrics &
\textcolor{forestgreen}{\cmark{}} discovers sparse, latent preferences &
\textcolor{forestgreen}{\cmark{}} true cold-start; no history &
\textcolor{forestgreen}{\cmark{}} verifiable reasoning tasks &
\textcolor{forestgreen}{\cmark{}} multi-domain reasoning \\

\bottomrule
\end{tabular}\vspace{-2mm}
}
\end{table*} 

We introduce \methodname, an evaluation methodology that transforms existing reasoning benchmarks into interactive personalization assessments. We generate psychologically-grounded personas and instantiate sparse preference profiles where only a subset of 20-25 possible attributes (expertise level, affective features, meta-cognitive features, etc.) are relevant for each persona-task pair. Models must discover these hidden preferences through strategic questioning, then adapt their responses accordingly. We evaluate both preference discovery accuracy and response alignment using fine-grained rubrics, comparing against baseline (no personalization) and oracle (known preferences) conditions across mathematical, scientific, and social reasoning tasks.

Evaluation of 21 frontier models across 10 tasks reveals systematic failures in personalized reasoning capabilities. In 29.0\% of cases, attempting personalization produces worse preference alignment than generic responses, yet the genetic responses still fail to address user needs. Models exhibit insufficient questioning, asking only 1.42 questions on average despite 5-turn allowances, and fail to identify relevant preference dimensions. Domain analyses show disparities among task types: mathematical reasoning suffers severe degradation under personalization constraints (3.5\% accuracy loss) while social reasoning maintains robustness (3.1\% gain).
The (lack of) proactive personalized reasoning capability in models has critical implications for educational applications, where misaligned explanations can impede learning by providing inappropriate cognitive scaffolding, and for healthcare and technical support domains where one-size-fits-all responses may lead to misunderstanding of complex procedures or safety-critical information.
We make the following contributions:

\begin{itemize}[noitemsep,topsep=0pt,leftmargin=0.4cm]
    \item We define \capabilityname as a distinct capability requiring models to proactively discover user preferences through strategic questioning and adapt their responses accordingly, distinguishing it from static persona consistency or content recommendation.
    \item We introduce \methodname, a systematic approach for transforming static benchmarks into interactive personalization tasks, bridging the gap between reasoning competence and user adaptation through fine-grained, sparse preference modeling.
    \item We propose \metricname, a rubric-based metric to measure the alignment of model responses to the user's preferences to quantify the effectiveness of personalization.
    \item We reveal fundamental failure modes across 21 frontier models, demonstrating that personalized reasoning requires dedicated development rather than emerging from general language understanding, and identify domain-specific brittleness patterns that inform future research directions.
\end{itemize}

\section{\capabilityname}

In this section, we operationalize the problem of \capabilityname through a three-part decomposition. Section~\ref{sec:identify}
introduces the notion of task-relevant attributes that define the space in which personalization can
occur. Section~\ref{sec:elicit} develops the representation of user preferences over these attributes and
describes elicitation as a sequential decision process. Section~\ref{sec:align} formalizes how models adapt their
responses to the inferred preference profile and how preference alignment is evaluated jointly with
correctness. This decomposition provides the foundation for the \methodname evaluation framework in
Section~\ref{sec:pipeline}.

\subsection{Identifying Relevant Attributes}\label{sec:identify}
Our overarching goal is to enable language models to generate personalized responses that better align with a user’s learning needs and preferences, rather than providing generic explanations. Achieving this requires first identifying and modeling the salient attributes that can shape how an explanation is delivered. We therefore begin with the assumption that there exists a very large but finite global set of attributes to which a response can be personalized. We denote these attributes by $\Theta = \{\theta_1, \ldots, \theta_d\}$. These attributes may include factors such as the use of analogies, the level of technical jargon, or the level of empowerment during learning, etc. Given a particular user and task, however, not all attributes are equally important; our focus is on modeling which subset of attributes is most relevant for delivering personalization in that context.

\paragraph{Fine Grained Preference Modeling.}

For any given task $i$, not all attributes in $\Theta$ are equally relevant. Only a small subset $\mathcal{F}(i) \subseteq \Theta$ matters for personalization in the context of the given task. 

The first component of \capabilityname{} is thus to infer which attributes matter for a given user-task pair. For instance, in a physics explanation task, ``visualization'' and ``analogies'' may be salient, while ``ethical context'' may not be.

\subsection{Eliciting Preference Values}\label{sec:elicit}

Once the model estimates the relevant preference attributes for a problem instance, \capabilityname requires incorporating the user. Even within the same task instance, different users may emphasize different attributes.

Consider a medical explanation task as in Figure~\ref{fig:framework}. Suppose the relevant attributes are empathy and technical jargon, i.e., 
$F(i)=\{\text{empathy, jargon}\}$. One patient may prioritize empathy and plain language to reduce anxiety, while another may prefer a more technical explanation that uses medical terminology, and is ambivalent toward the level of empathy. Both users share the same set of relevant attributes, but they assign different importance weights. This illustrates why we need to represent not only a preference value for each attribute but also a weight capturing its relative significance.

Furthermore, consider Alice asking the same medical question in two different contexts. While studying for an exam, she may prefer \emph{high technical jargon}, since precise terminology is useful for learning. By contrast, if Alice faces an emergency medical situation herself, she may prefer \emph{low jargon} and plain language, focusing on clarity over technical detail. This shows that preference values themselves ($v_j$) can shift across instances, even for the same user, and hence preferences should be defined at the instance level.

 We define the \emph{preference profile} of user $p$ for problem instance $i$ as
\[
\mathcal{P}_{p,i} = \{ (\theta_j, v_j, w_j) : \theta_j \in \mathcal{F}(i) \}, \text{where:}
\]
\begin{itemize}[noitemsep,topsep=0pt,leftmargin=0.4cm]
    \item $\theta_j$ is a relevant prefernce attribute in $\mathcal{F}(i)$,
    \item $v_j$ encodes user $p$'s preference value for attribute $\theta_j$ (e.g., ``high jargon'' vs.\ ``low jargon''), 
    \item $w_j \geq 0$ denotes the relative importance weight, with $\sum_{\theta_j \in \mathcal{F}(i)} w_j = 1$.
\end{itemize}

The distinction is essential: $v_j$ specifies \emph{which direction} the user prefers along an attribute, while $w_j$ specifies \emph{how much that attribute matters relative to others}.  

Since $\mathcal{P}_{p,i}$ is unobserved, the model must perform \emph{preference elicitation}. We model elicitation as a sequential decision process: at each turn $t$, the model selects an action
\[
a_t \in \{ \text{ask}(\theta) \mid \theta \in \mathcal{F}(i) \} \cup \{\text{answer}\}.
\]
If $a_t = \text{ask}(\theta)$, the user provides information about their preference value $v_\theta$, and the model refines its estimate of $\mathcal{P}_{p,i}$. If $a_t = \text{answer}$, elicitation terminates, and the model produces a response conditioned on the inferred profile $\hat{\mathcal{P}}_{p,i}$.  

This framing highlights that personalized reasoning is not only about \emph{knowing what the relevant attributes are}, but also about \emph{understanding their relative importance and values} and doing so efficiently under limited interaction.

\subsection{Adapting Responses and Evaluating Alignment}\label{sec:align}

Once the model has inferred an estimate of the user’s preference profile $\hat{\mathcal{P}}_{p,i}$, it must adapt its reasoning and outputs accordingly. Personalization involves more than stylistic choices: it requires shaping explanations along the attributes that the user values most.

For example, in a medical explanation task, if empathy is assigned high weight, the model should produce a response that foregrounds reassurance and clarity. If technical jargon is highly weighted instead, the model needs to allocate more reasoning budget to rigorous scientific hypotheses, existing literature, etc., and should lean toward precise terminology in its response, even at the expense of emotional tone. The same correct answer choice may therefore be expressed quite differently depending on the inferred preference profile.

Evaluation of personalized reasoning thus involves two complementary objectives: \emph{correctness}, meaning that the answer is objectively valid for the problem instance, and \emph{preference alignment}, meaning that the answer respects the user’s weighted preferences.

\paragraph{Preference alignment.} For each relevant attribute $\theta_j \in F(i)$, we define a grading function\vspace{-4mm}

\[
g_j(r, v_j) \in [0,5],
\]

\vspace{-3mm}
which measures how well response $r$ satisfies user $p$’s preference value $v_j$ along attribute $\theta_j$. For example, $g_j$ may quantify whether the amount of jargon in a medical explanation matches the user’s expressed tolerance. The overall alignment score is then given by\vspace{-4mm}

\begin{equation}\label{eq:prefalign}
    \text{\metricname}(r, \mathcal{P}_{p,i}) = \sum_{\theta_j \in F(i)} w_j \cdot g_j(r, v_j).
\end{equation}

\vspace{-3mm}
\paragraph{Joint objective.} High-quality personalized reasoning requires responses that are both objectively correct and preference-aligned. Formally, for a response $r$ to be successful, we require
\[
\text{Correct}(r, i) = 1 \quad \text{and} \quad \text{\metricname}(r, \mathcal{P}_{p,i}) \text{ is maximized}.
\]

\vspace{-3mm}
This formulation highlights that personalization is not merely about delivering accurate answers, but about tailoring those answers to the user’s weighted, instance-specific preferences.

\begin{figure}
    \centering
    \includegraphics[width=\linewidth]{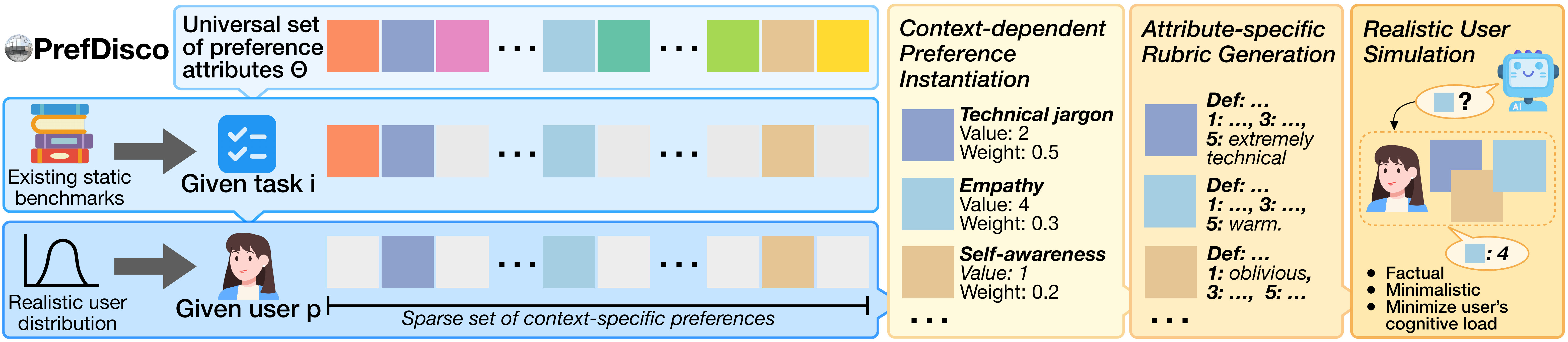}\vspace{-2.5mm}
    \caption{\methodname benchmark construction pipeline. The framework transforms static benchmarks by sampling sparse, context-dependent preference subsets for each user-task pair, generating attribute-specific evaluation rubrics, and implementing realistic user simulation that requires models to discover preferences through ``just-in-time'' strategic questioning in cold-start scenarios.}
    \label{fig:prefdisco}\vspace{-2mm}
\end{figure}

\section{\methodname Benchmark Construction}\label{sec:pipeline}\vspace{-0.5mm}

\methodname addresses a fundamental gap in personalization evaluation: existing benchmarks assume preferences are either known a priori or inferrable from context, failing to capture the cold-start scenarios where models must discover user needs through interaction. As illustrated in Figure~\ref{fig:prefdisco}, our methodology transforms static benchmarks into interactive personalization tasks through four components designed to isolate and measure preference discovery capabilities.

\vspace{-0.5mm}
\paragraph{Psychologically-Grounded Persona Generation.}

We generate psychologically-grounded personas rather than arbitrary user archetypes because personality traits systematically influence learning preferences and communication styles. Personas are conditioned on the International Personality Item Pool \citep{goldberg2006international}, incorporating demographics, Big Five personality dimensions, and domain expertise that remain consistent across problem instances.
We perform high-temperature sampling (t=0.7) with rejection sampling when generating personas to ensure diverse coverage while preventing over-representation of common attribute combinations. This consistency enables evaluation of models' ability to transfer discovered preferences within user sessions—a critical capability for practical deployment where users interact with systems across multiple tasks.

\vspace{-0.5mm}
\paragraph{Context-Dependent Preference Instantiation.}

Traditional personalization assumes fixed preference profiles that apply universally across tasks. We reject this approach because psychological research demonstrates that individuals prioritize different attributes across contexts \citep{fischer2011whence}. As formalized in \S\ref{sec:elicit}, for each persona-problem pair $(p,i)$, we sample sparse preference profiles $\mathcal{P}_{p,i} = \{(\theta_j, v_j, w_j)\}_{\theta_j \in \mathcal{F}(i)}$ where only context-relevant attributes are active.
Further, we ground the preference sampling process on existing research in education, which states that frequently modeled student characteristics include knowledge level, misconceptions, cognitive features, affective features, and meta-cognitive features \citep{chrysafiadi2015advances}.

This sparse modeling is essential because it reflects realistic user behavior: the same person may prioritize technical precision in professional settings while favoring accessibility in casual interactions. We determine relevant attribute subsets $\mathcal{F}(i)$ through LLM classification, validated by human annotation on 20 scenarios (2 per task). Each scenario includes 10 relevant and 10 irrelevant attributes, generating 400 labels per annotator across 3 annotators. Inter-annotator agreement achieved Fleiss kappa of 0.463 with 61.5\% accuracy against majority voting, which is considered moderate agreement especially for subjective tasks \citep{sap-etal-2017-connotation,budur-etal-2020-data,mire-etal-2024-empirical}. See annotation details in Appendix~\ref{app:human_eval}.
Finally, importance weights satisfy $\sum_{\theta_j \in \mathcal{F}(i)} w_j = 1$ and reflect persona-specific priorities. LLM-based semantic deduplication ensures attribute diversity by removing redundant dimensions that would artificially inflate preference complexity.

\vspace{-0.5mm}
\paragraph{Evaluation Rubric Generation.}

We generate attribute-specific evaluation rubrics $g_j(r, v_j) \in \{1,\ldots,5\}$ using LLM-based assessment
to enable systematic evaluation across 10K scenarios. 
These rubrics provide the scalability necessary for comprehensive evaluation across
diverse domains that would be prohibitively expensive with human annotation alone. Importantly, by
leveraging the structured information obtained through our construction, these rubrics enable
fine-grained evaluation along specific attributes rather than relying on a single holistic satisfaction
score. This reduces susceptibility to hallucination and bias, since each attribute is judged against an
explicit criterion rather than aggregated into an opaque overall impression.

\paragraph{User Simulation.}

We implement passive user simulation inspired by \citep{li2024mediq}, where the user factually answers the model's elicitation question without providing any extra details. The passive user faithfully represents the challenging scenario of preference discovery while minimizing confounding factors. 
The passive user responds in natural language text, not scalar values; passivity refers to the user's minimal information sharing behavior, not to the response format.
The passive user type forces models to develop strategic questioning strategies rather than relying on user proactiveness
and isolates models' questioning capabilities from user communication style, providing controlled evaluation conditions. The 5-turn limit reflects realistic attention constraints in human-AI interaction while providing sufficient opportunity for effective preference discovery, as demonstrated by our correlation analysis between questioning volume and alignment quality. Performance curves across fixed interaction lengths (Figure~\ref{fig:num_questions}) plateau around 3–5 turns, supporting this choice.

Overall, \methodname decomposes personalization into constituent attributes, enabling granular analysis of model capabilities and failure modes rather than relying on holistic preference ratings that obscure specific deficiencies.

\section{Experiments}\vspace{-1mm}

Our goal is to evaluate models’ ability for \emph{proactive preference discovery}: engaging in dialogue to uncover hidden user requirements and adapt responses accordingly. To this end, our experiments combine three ingredients. First, we use diverse \textbf{benchmarks} spanning mathematical, scientific, and social reasoning to ensure domain-agnostic evaluation. Second, we introduce varied \textbf{personas} that encode heterogeneous user preferences, simulating realistic user variability. Third, we define controlled \textbf{evaluation conditions} that disentangle raw task ability, preference elicitation skill, and intrinsic personalization capacity. Together, these components create a challenging and diagnostic testbed for preference-aware reasoning.

\paragraph{Benchmarks and Models.} 
We apply \methodname\ to ten benchmarks spanning mathematical, logical, scientific, and social reasoning:  
MATH-500 \citep{hendrycks2021math}, AIME \citep{veeraboina2024aime}, LogiQA \citep{liu2020logiqa}, MascQA \citep{zaki2024mascqa}, ScienceQA \citep{saikh2022scienceqa}, MMLU \citep{hendrycks2021mmlu}, SimpleQA \citep{wei2024simpleqa}, MedQA \citep{jin2020medqa}, CommonsenseQA \citep{talmor2018commonsenseqa}, and SocialIQA \citep{sap2019socialiqa}.  
This mix covers tasks with different reasoning demands (symbolic, factual, commonsense, scientific), ensuring that results do not hinge on a narrow domain. We evaluate 21 frontier models (GPT, O-series, Gemini, and Claude variants). Details on model versions and hyperparameters are provided in Appendix~\ref{app:eval_details}.

\paragraph{Personas and Rubrics Implementations.}

We generate 100 diverse personas and randomly sample 100 problems per benchmark. For each problem, we assign 10 personas (with partial overlaps across problems), creating 1,000 evaluation scenarios per task and 10,000 total scenarios across all benchmarks. Each interaction is limited to 5 turns to simulate realistic attention constraints.
During benchmark construction, GPT-4.1, Gemini-2.5-Flash, and Claude-Sonnet-4 are randomly selected for each API call (persona generation, preference instantiation, or rubric creation) to ensure diversity and reduce single-model biases. Further details, including prompt templates, sampling distributions, and illustrative examples of full personas and dialogues, are provided in Appendix~\ref{app:eval_details}. 

\paragraph{Evaluation Conditions.}
Models are evaluated on the \metricname score under three conditions:  

\begin{itemize}[noitemsep,topsep=0pt,leftmargin=0.4cm]
    \item \textbf{Baseline.} Models receive the problem only, with no persona or preference information. This measures task ability under standard prompting, establishing the reference point for comparisons.  
    \item \textbf{Discovery.} Models are \emph{system-prompted} to elicit user preferences through multi-turn dialogue before producing a final answer. This isolates the capability of \emph{\capabilityname}: asking effective questions, inferring which attributes matter, and adapting explanations accordingly.  

    \item \textbf{Oracle.} Models are \emph{system-prompted} with the full ground-truth preference profile provided upfront. This removes the uncertainty of discovery and evaluates only how well a model can \emph{use} known preferences to personalize its responses.  
\end{itemize}
 
 The \emph{baseline} establishes task-only performance. The gap between \emph{baseline} and \emph{discovery} quantifies a model’s ability to uncover preferences interactively, while the gap between \emph{baseline} and \emph{oracle} shows its upper bound on personalization quality. Raw oracle scores highlight intrinsic differences in models’ ability to incorporate preferences once they are known, independent of discovery strategy.

\paragraph{Normalized Preference Alignment.}

In addition to the raw preference alignment scores (Eq.~\ref{eq:prefalign}), we normalize performance relative to the baseline and oracle conditions so that performance is directly comparable across models with different baselines and personalization ceilings:
\vspace{-6mm}

\begin{equation}\small
\label{eq:norm_align}
\textit{NormAlign}(r_{\text{discovery}}, \mathcal{P}_{p,i}) = 100 \times 
\frac{\textit{\metricname}(r_{\text{discovery}}, \mathcal{P}_{p,i}) - \textit{\metricname}(r_{\text{baseline}}, \mathcal{P}_{p,i})}
{\textit{\metricname}(r_{\text{oracle}}, \mathcal{P}_{p,i}) - \textit{\metricname}(r_{\text{baseline}}, \mathcal{P}_{p,i})},
\end{equation}\vspace{-4mm}

where $r_{\text{discovery}}$ is the final response produced in discovery mode, $r_{\text{baseline}}$ is the response in baseline mode, and $r_{\text{oracle}}$ is the response in oracle mode with the full preference profile provided.  

A score of $0$ indicates no improvement over baseline, $100$ indicates perfect discovery matching oracle performance, and negative values reflect reduced satisfaction compared to baseline. This provides a scale-independent measure of discovery quality relative to a model’s own upper bound.

\paragraph{Task Accuracy.}
We report objective task accuracy using each benchmark’s original evaluation metric. This acts as a safeguard: personalization should \emph{augment} user satisfaction without degrading the correctness of the underlying task. A strong model must therefore achieve both high preference alignment and high task accuracy.

\section{Results}

\begin{table}[t]
\centering
\caption{Normalized preference alignment scores, calculated by normalizing the preference alignment score of the Discovery mode against the lower bound Baseline (no personalization) and upper bound Oracle (full preference profile provided) conditions. A score of 100.0 means perfect discovery matching oracle performance, 0.0 indicates no improvement over baseline, and negative values show that attempted personalization produced worse alignment than generic responses. Notably, 29.0\% of model–task combinations yield negative scores, revealing that naive preference elicitation often harms alignment rather than helping.}
\vspace{-1mm}
\scalebox{0.83}{
\begin{tabular}{lrrrrrrr} %
\toprule
\textbf{openai} & \textbf{gpt-4o} & \textbf{gpt-4.1} & \textbf{o1} & \textbf{o3} & \textbf{o1-mini} & \textbf{o3-mini} & \textbf{o4-mini} \\
\midrule
\textbf{math}          & \heat{4.9}  & \heat{-13.2} & \heat{16.6} & \heat{-6.0}  & \heat{-20.9} & \heat{-10.3} & \heat{21.9} \\
\textbf{aime}          & \heat{21.2} & \heat{1.9}   & \heat{11.9} & \heat{5.3}   & \heat{-11.9} & \heat{-7.4}  & \heat{20.5} \\
\textbf{logiqa}        & \heat{7.7}  & \heat{-29.9} & \heat{4.2}  & \heat{-50.4} & \heat{-5.2}  & \heat{-15.4} & \heat{26.0} \\
\textbf{mascqa}        & \heat{9.7}  & \heat{-11.6} & \heat{13.0} & \heat{-9.0}  & \heat{1.1}   & \heat{-1.5}  & \heat{25.1} \\
\textbf{medqa}         & \heat{-6.6} & \heat{-26.9} & \heat{9.6}  & \heat{-5.9}  & \heat{19.1}  & \heat{3.4}   & \heat{23.8} \\
\textbf{scienceqa}     & \heat{10.7} & \heat{3.6}   & \heat{7.5}  & \heat{12.8}  & \heat{2.1}   & \heat{-9.2}  & \heat{16.7} \\
\textbf{mmlu}          & \heat{18.3} & \heat{-11.3} & \heat{10.4} & \heat{-5.8}  & \heat{18.4}  & \heat{-1.3}  & \heat{23.2} \\
\textbf{simpleqa}      & \heat{14.8} & \heat{11.8}  & \heat{-12.3}& \heat{0.1}   & \heat{27.9}  & \heat{-47.7} & \heat{7.5}  \\
\textbf{commonsenseqa} & \heat{25.2} & \heat{5.8}   & \heat{7.3}  & \heat{2.6}   & \heat{7.6}   & \heat{-0.4}  & \heat{16.0} \\
\textbf{socialiqa}     & \heat{21.2} & \heat{11.6}  & \heat{7.1}  & \heat{3.8}   & \heat{4.8}   & \heat{-0.1}  & \heat{17.4} \\
\midrule
\textbf{gemini} & \textbf{1.5-flash} & \textbf{1.5-pro} & \textbf{2.0-flash-lite} & \textbf{2.0-flash} & \textbf{2.5-flash-lite} & \textbf{2.5-flash} & \textbf{2.5-pro} \\
\midrule
\textbf{math}          & \heat{20.7} & \heat{19.8} & \heat{-5.5} & \heat{17.5} & \heat{12.5} & \heat{-10.9} & \heat{-13.5} \\
\textbf{aime}          & \heat{28.7} & \heat{28.9} & \heat{28.9} & \heat{40.3} & \heat{27.5} & \heat{25.8}  & \heat{14.9} \\
\textbf{logiqa}        & \heat{23.5} & \heat{16.0} & \heat{-3.0} & \heat{-0.9} & \heat{9.4}  & \heat{-38.1} & \heat{-0.3} \\
\textbf{mascqa}        & \heat{27.2} & \heat{31.1} & \heat{5.2}  & \heat{-4.9} & \heat{20.3} & \heat{-0.6}  & \heat{10.4} \\
\textbf{medqa}         & \heat{6.7}  & \heat{9.5}  & \heat{-7.2} & \heat{-17.3}& \heat{18.5} & \heat{4.6}   & \heat{35.7} \\
\textbf{scienceqa}     & \heat{22.1} & \heat{23.9} & \heat{2.1}  & \heat{6.4}  & \heat{13.8} & \heat{0.3}   & \heat{17.9} \\
\textbf{mmlu}          & \heat{27.9} & \heat{17.6} & \heat{5.0}  & \heat{4.4}  & \heat{23.8} & \heat{-8.2}  & \heat{10.3} \\
\textbf{simpleqa}      & \heat{18.1} & \heat{19.9} & \heat{-3.3} & \heat{7.0}  & \heat{6.4}  & \heat{4.8}   & \heat{8.4}  \\
\textbf{commonsenseqa} & \heat{24.9} & \heat{23.6} & \heat{-7.8} & \heat{5.4}  & \heat{6.7}  & \heat{-0.7}  & \heat{20.2} \\
\textbf{socialiqa}     & \heat{27.0} & \heat{18.9} & \heat{-1.1} & \heat{10.7} & \heat{3.9}  & \heat{11.3}  & \heat{29.3} \\
\midrule
\textbf{claude} & \textbf{sonnet-4} & \textbf{opus-4} & \textbf{3-7-sonnet} & \textbf{3-5-haiku} & \textbf{3-5-sonnet-v2} & \textbf{3-5-sonnet-v1} & \textbf{3-opus} \\
\midrule
\textbf{math}          & \heat{2.6}  & \heat{16.9} & \heat{-2.8} & \heat{-23.8} & \heat{-9.6} & \heat{15.6} & \heat{7.7} \\
\textbf{aime}          & \heat{17.1} & \heat{29.9} & \heat{0.7}  & \heat{-19.1} & \heat{-5.5} & \heat{15.8} & \heat{-25.4} \\
\textbf{logiqa}        & \heat{-4.1} & \heat{14.7} & \heat{-5.9} & \heat{-5.9}  & \heat{3.6}  & \heat{38.8} & \heat{19.4} \\
\textbf{mascqa}        & \heat{1.9}  & \heat{20.2} & \heat{-6.0} & \heat{7.5}   & \heat{11.8} & \heat{26.9} & \heat{21.3} \\
\textbf{medqa}         & \heat{8.3}  & \heat{33.0} & \heat{2.9}  & \heat{4.0}   & \heat{15.1} & \heat{24.0} & \heat{9.8} \\
\textbf{scienceqa}     & \heat{-4.6} & \heat{10.6} & \heat{0.1}  & \heat{6.4}   & \heat{-9.2} & \heat{9.9}  & \heat{0.1} \\
\textbf{mmlu}          & \heat{1.1}  & \heat{18.6} & \heat{-9.9} & \heat{9.3}   & \heat{5.4}  & \heat{26.2} & \heat{14.2} \\
\textbf{simpleqa}      & \heat{-13.9}& \heat{2.3}  & \heat{-2.5} & \heat{16.8}  & \heat{26.6} & \heat{-10.8}& \heat{13.4} \\
\textbf{commonsenseqa} & \heat{-16.1}& \heat{2.2}  & \heat{-16.6}& \heat{5.9}   & \heat{-5.8} & \heat{1.8}  & \heat{24.4} \\
\textbf{socialiqa}     & \heat{10.5} & \heat{7.7}  & \heat{1.9}  & \heat{15.8}  & \heat{-6.8} & \heat{-8.7} & \heat{26.4} \\
\bottomrule
\end{tabular}
}
\label{tab:results}
\end{table}

\paragraph{Preference Discovery Performance.  }

Table~\ref{tab:results} reveals systematic failures in preference discovery. Of 210 model-task combinations, 61 (29.0\%) show negative normalized alignment, meaning the discovery responses align worse with user preferences than baseline responses that made no personalization attempt. This suggests that models are prone to over-correction errors, modifying aspects of their responses that were already acceptable in baseline conditions. \textbf{Naively attempting proactive personalization often makes alignment worse than providing generic responses.}

Out of the tasks, MATH and LogiQA show the most degradation (10 and 11 out of 21 models perform worse when attempting to personalized), while SocialIQA benefit the most from interactive personalization.
Claude Opus 4 shows the most consistent positive performance, while o3-high exhibits extreme variance, indicating significant architectural differences in personalization capability.
Notably, older models such as Claude 3-Opus occasionally outperform their newer counterparts (e.g., Claude Sonnet-4) on discovery-mode alignment, consistent with findings that RL can cause model collapse or reduce output diversity \citep{murthy2025one,shao2025spurious}, rendering more recent models less adaptable to the varied reasoning paths required by personalization.

\begin{figure}[t]
\centering
\begin{minipage}{.49\textwidth}
  \centering
  \includegraphics[width=\linewidth]{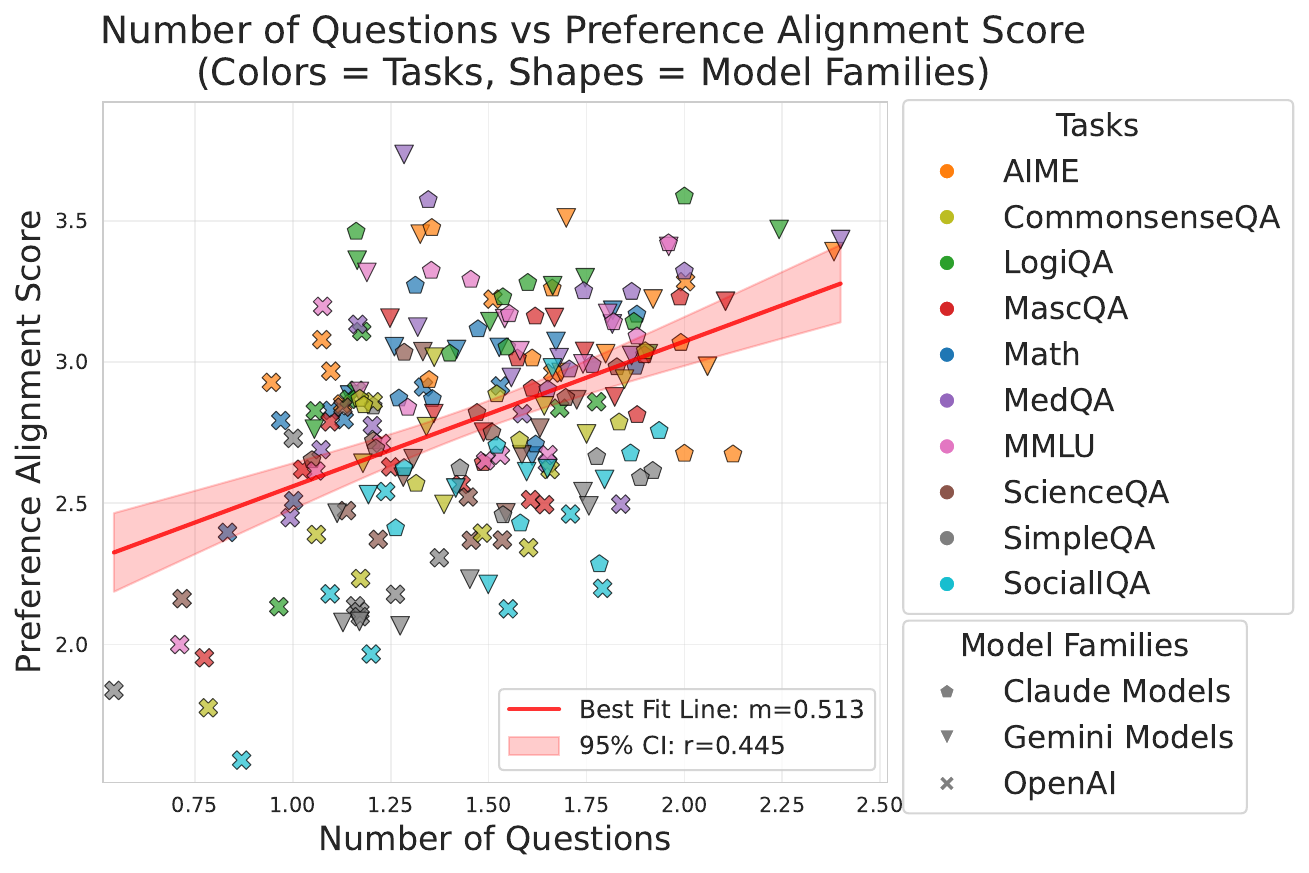}\vspace{-2.5mm}
  \caption{Positive correlation (r=0.445) between question volume and preference alignment. Better personalization requires more extensive questioning. Regression coefficients: Claude=0.117, OpenAI=0.379, Gemini=0.474.}
    \label{fig:questions_alignment}
\end{minipage}
\hfill
\begin{minipage}{.49\textwidth}
  \centering
  \includegraphics[width=\linewidth]{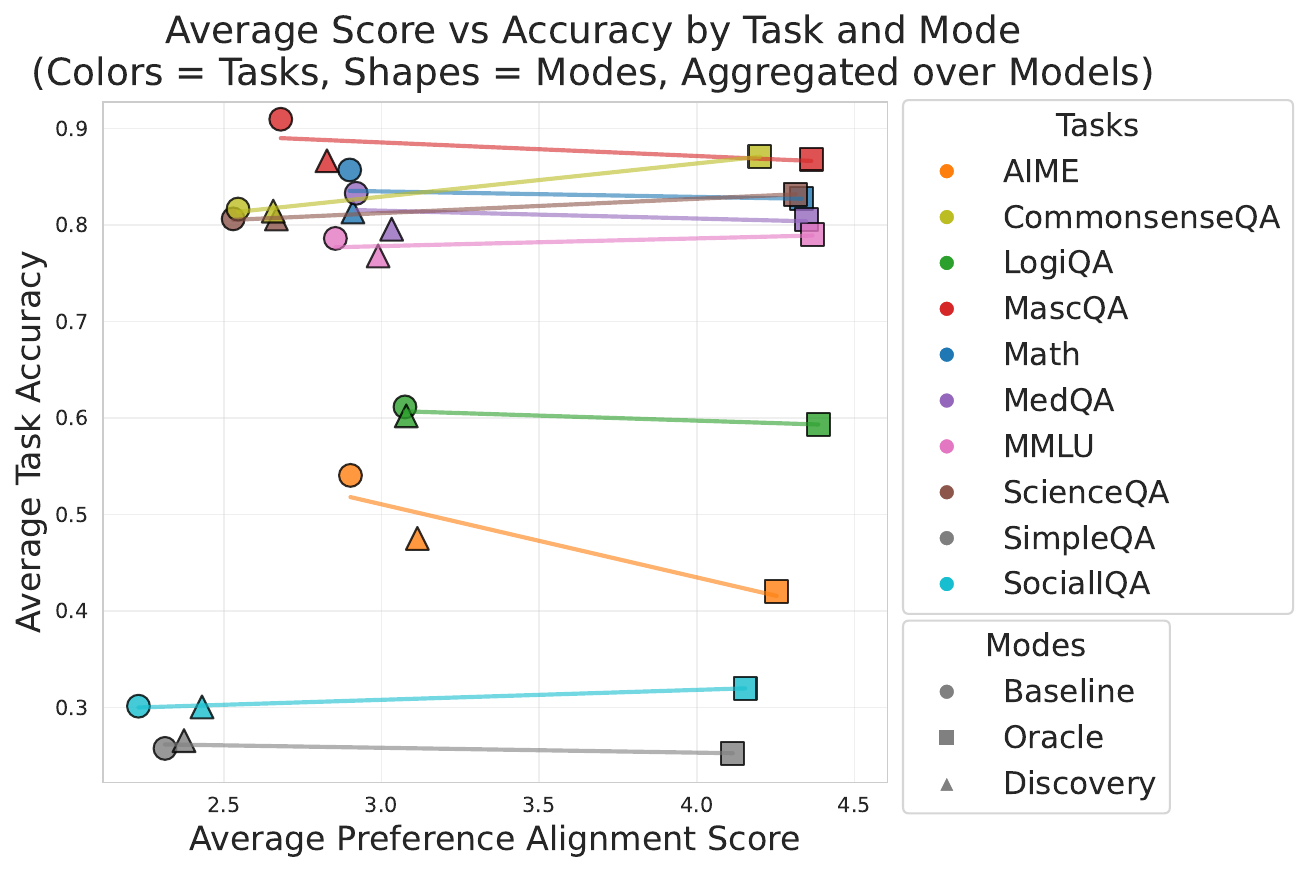}\vspace{-2.5mm}
  \caption{More personalization constraints in context hinder model reasoning abilities. Overall accuracy: Baseline=0.652, Oracle=0.618, Discovery=0.601. Trade-off is most pronounced in Math, AIME, and logic tasks.}
    \label{fig:accuracy_tradeoff}
\end{minipage}\vspace{-2mm}
\end{figure}

\paragraph{Interaction Efficiency and Preference Alignment Tradeoff.}
Figure~\ref{fig:questions_alignment} reveals why many personalization attempts fail. While the positive correlation (r=0.445, p$<$0.001) demonstrates that extensive questioning improves alignment, most models ask only 1.48 questions on average despite a maximum allowance of 5 turns. This places the majority of interactions in the low-performance region where insufficient questioning yields worse alignment than baseline responses, explaining the 29.0\% negative performance rate.

The regression coefficients vary dramatically by model family: Gemini ($\beta$=0.474), OpenAI ($\beta$=0.379), Claude ($\beta$=0.117). Gemini's higher coefficient indicates more effective question utilization—each additional question yields greater alignment improvement. 
This suggests current prompting methods are limited not just in question quantity, but in question quality and strategic timing. Models that ask better questions achieve more personalization gains.

\paragraph{Accuracy-Personalization Trade-off.}%
The systematic accuracy degradation across conditions: Baseline (65.2\%), Oracle (61.8\%), Discovery (60.1\%) reveals that personalization imposes fundamental cognitive costs (Figure~\ref{fig:accuracy_tradeoff}). 
Crucially, the accuracy drop from the Baseline to the Oracle condition—where no interactive discovery is required—demonstrates that this cost is inherent to processing and adhering to preference constraints, rather than stemming from the overhead of multi-turn dialogue.
This trade-off exhibits significant domain-specific disparity. 
Mathematical tasks suffer severe degradation (AIME: 12.1\% loss), while social tasks remain robust or even improve (CommonsenseQA: 5.4\% gain). 
We conjecture that this divergence stems from the prevailing training paradigm for state-of-the-art LLMs, which are heavily optimized for performance on verifiable mathematical benchmarks.

This optimization, often achieved through reinforcement learning (RL) on mathematical tasks, encourages models to converge on a narrow set of high-reward reasoning pathways. While effective for standard problem-solving, this fixation renders models brittle when confronted with the long-tail contextual constraints of personalization. Effective personalized reasoning frequently requires altering the core reasoning steps themselves—for example, avoiding advanced calculus for a novice user—rather than merely applying stylistic changes to a pre-established solution path. We find that models often fail to generate a correct solution when forced to employ a different cognitive toolkit, as their reasoning is inflexibly tied to the pathways reinforced during training.

This analysis reveals a fundamental limitation in current architectures: the reasoning processes optimized through RL are often incompatible with the dynamic cognitive adaptations required for personalization. When user preferences necessitate a departure from these reinforced pathways, the model's alternative reasoning frequently proves inadequate, leading to a degradation in task accuracy. This trade-off between preference alignment and reasoning robustness is a critical failure mode, qualitatively exemplified in Appendix~\ref{app:fail_example}.

\begin{wrapfigure}{R}{0.48\textwidth}%
    \centering\vspace{-4mm}
    \includegraphics[width=\linewidth]{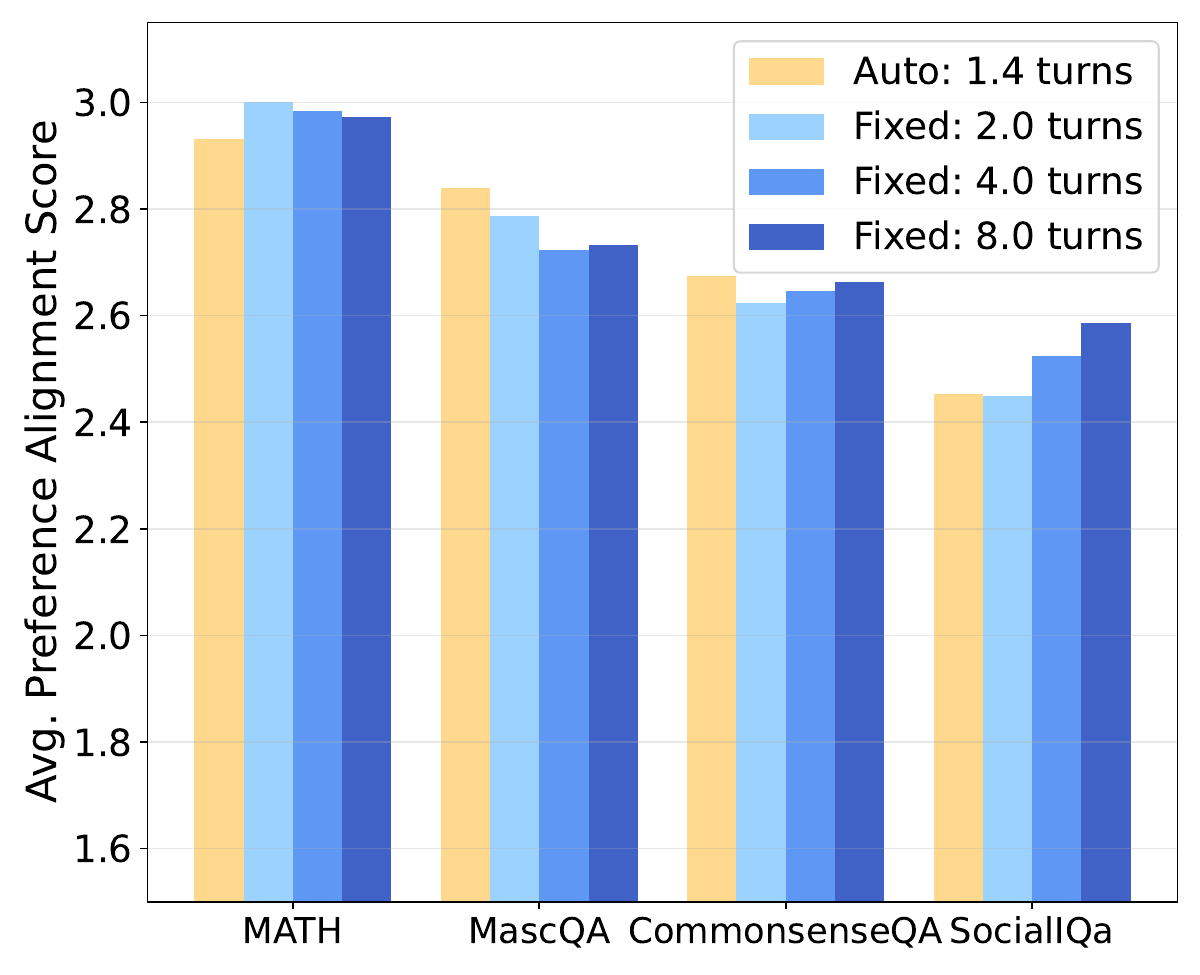}\vspace{-1mm}
    \caption{Fixed interaction length hinders preference alignment on math and science tasks but improves preference alignment on social reasoning.}
    \label{fig:num_questions}\vspace{-2mm}
\end{wrapfigure}%

\paragraph{Question Quality vs. Termination Decision Isolation.}%
Figure~\ref{fig:num_questions} isolates question quality from termination decisions by forcing models to ask a fixed number of questions, revealing that the domain-specific performance patterns persist regardless of question quantity control. When models are constrained to ask 2, 4, or 8 questions instead of choosing when to stop, mathematical and scientific reasoning tasks (MATH, MascQA) continue to show degraded performance with increased questioning, while social reasoning tasks (CommonsenseQA, SocialIQA) maintain improved performance. This consistency across fixed interaction lengths demonstrates that poor performance in mathematical domains stems from fundamental incompatibilities in how models process preference constraints during formal reasoning, rather than from suboptimal termination decisions. The persistence of domain-specific brittleness under controlled questioning conditions suggests that models trained with current training paradigm struggle with the cognitive overhead of simultaneously maintaining logical precision and adapting to user preferences, indicating that the observed failures reflect deeper architectural limitations rather than strategic questioning deficiencies.

\section{Related Work}\vspace{-1mm}

\textbf{Static Personalization and Evaluation Benchmarks.}  
Several benchmarks evaluate personalization in language models but assume known preferences or static consistency rather than interactive discovery. PersoBench \citep{afzoon2024persobench}, PrefEval \citep{zhao2025prefeval}, PersonaMem \citep{jiang2025know}, and PersonaConvBench \citep{li2025personaconvbench} focus on dialogue generation or multi-session profiling without addressing cold-start preference elicitation across reasoning tasks.

\textbf{User Preference Modeling.}  
Prior work models user preferences through explicit categorization \citep{jiang2023evaluating, zhu2024personality, bose2024initializing}, per-user reward models \citep{poddar2024personalizing, chen2024pal, lee2024test}, or fine-grained multi-dimensional approaches \citep{bose2025lore, li2025prefpalette}. However, these methods do not address which preference attributes are relevant for specific user-task combinations or how to discover them interactively in cold-start scenarios.

\textbf{Interactive Preference Elicitation.}  
GATE \citep{li2023eliciting} and MediQ \citep{li2024mediq} demonstrate interactive questioning for user intent understanding and clinical information-seeking, respectively. These approaches focus on narrow domains without the reasoning adaptation component central to personalized reasoning. \methodname uniquely combines interactive preference discovery with adaptive reasoning across diverse domains, requiring models to modify their cognitive approaches based on discovered user needs.

\section{Discussion and Future Work}\vspace{-3mm}

We introduce personalized reasoning as a fundamental capability for human-facing AI systems, requiring models to proactively adapt their cognitive processes based on discovered user preferences rather than merely style-transforming response presentation. Our evaluation reveals systematic failures across frontier models: 29.0\% of personalization attempts perform worse than generic responses, with mathematical reasoning showing universal degradation while social reasoning maintains robustness. These domain-specific patterns persist even when controlling for question quantity, indicating that current architectures face fundamental incompatibilities between preference processing and formal reasoning rather than strategic questioning deficiencies.

\methodname establishes personalized reasoning as a measurable research frontier through a scalable evaluation methodology that transforms any static benchmark into an interactive personalization assessment. Unlike existing approaches that assume known preferences or evaluate static consistency, our framework operationalizes both preference discovery and reasoning adaptation in realistic cold-start scenarios. The methodology's generalizability across diverse task domains provides a systematic foundation for evaluating and developing adaptive AI systems.
Our findings reveal critical limitations in current language models. The positive correlation between questioning volume and alignment quality demonstrates that extensive interaction improves personalization, yet models ask only 1.48 questions on average despite 5-turn allowances. More importantly, the persistent accuracy degradation under personalization constraints indicates cognitive costs in processing user preferences simultaneously with task solving. This suggests that personalized reasoning requires dedicated research efforts rather than emerging from general language understanding improvements.

Future research directions include analyzing attribute-specific alignment patterns to identify model biases, leveraging the multi-dimensional reward structure for reinforcement learning, and investigating cross-task preference transfer. The framework provides a technical foundation for developing AI systems that can adapt to individual users in education, healthcare, and technical domains where personalized interaction is critical for effective deployment.

\section*{Limitations}

Our evaluation focuses on beneficial personalization scenarios and does not address potential negative aspects of personalization. We do not study over-personalization, where excessive adaptation to user preferences may reduce response quality or lead to information bubbles. Additionally, our framework does not evaluate sycophantic behavior, where models might prioritize agreement with user preferences over factual accuracy or helpful feedback.

Our simulated user interactions, while psychologically grounded, may not capture the full complexity of real human preference expression. The framework currently evaluates communication preferences rather than content preferences, and does not address preference evolution or conflicting preferences across different contexts.

\section*{Ethics Statement}

Personalization capabilities raise important ethical considerations. While our work aims to improve user experience through better preference alignment, these same capabilities could potentially be misused for manipulation or to reinforce harmful biases. Our framework evaluates technical capabilities without addressing the broader question of when and how personalization should be applied.

Future deployments of personalization systems should include safeguards against over-personalization, mechanisms to maintain factual accuracy despite user preferences, and transparency about how user preferences are discovered and applied. Our evaluation framework could be extended to assess these safety considerations alongside personalization effectiveness.

\section*{Acknowledgment}
This research was developed in part with funding from the Defense Advanced Research Projects Agency's (DARPA) SciFy program (Agreement No. HR00112520300). The views expressed are those of the author and do not reflect the official policy or position of the Department of Defense or the U.S.~Government. 
This material is based in part upon work supported by the Defense Advanced Research Projects Agency and the Air Force Research Laboratory, contract number(s): FA8650-23-C-7316. Any opinions, findings and conclusions, or recommendations expressed in this material are those of the author(s) and do not necessarily reflect the views of AFRL or DARPA.
This research was supported by the University of Washington Population Health Initiative, Amazon Health, the UW+Amazon Science Hub. The work of MF was supported in part by awards NSF CCF 2212261, NSF CCF 2312775, NSF TRIPODS II DMS-2023166, the Meta AIM program, and the Moorthy Family Professorship at UW.

\bibliography{iclr2025_conference}
\bibliographystyle{iclr2025_conference}

\appendix
\section{Evaluation Details}\label{app:eval_details}

\paragraph{Model Configurations} We evaluate 21 frontier language models across three major families with consistent hyperparameters (temperature=0.7, reasoning\_effort=high):

\textbf{OpenAI models:} gpt-4o, gpt-4.1, o1, o3, o1-mini, o3-mini, o4-mini

\textbf{Google models:} gemini-1.5-flash, gemini-1.5-pro, gemini-2.0-flash-lite, gemini-2.0-flash, gemini-2.5-flash-lite, gemini-2.5-flash, gemini-2.5-pro

\textbf{Anthropic models:} claude-sonnet-4, claude-opus-4, claude-3.7-sonnet, claude-3.5-haiku, claude-3.5-sonnet-v2, claude-3.5-sonnet-v1, claude-3-opus

\paragraph{Benchmark Selection} We apply \methodname to ten diverse benchmarks spanning mathematical reasoning (MATH-500, AIME), logical reasoning (LogiQA), scientific reasoning (MascQA, ScienceQA, MedQA), general knowledge (MMLU, SimpleQA), and social reasoning (CommonsenseQA, SocialIQA). This coverage demonstrates domain-agnostic applicability across formal and informal reasoning tasks.

\paragraph{Experimental Protocol} Each benchmark is transformed using 100 diverse personas randomly sampled from our psychologically-grounded persona library. We evaluate 100 problems per benchmark, with each problem assigned to 10 personas (with partial overlaps), creating 1,000 evaluation scenarios per task and 10,000 total scenarios. Each interaction is limited to 5 conversational turns to simulate realistic attention constraints.

Models are evaluated under three conditions: (1) \textit{Baseline Mode} provides standard responses without persona or preference information; (2) \textit{Discovery Mode} requires interactive preference elicitation through conversation; (3) \textit{Oracle Mode} supplies complete preference profiles upfront. This design isolates interactive discovery capabilities from general personalization abilities while establishing performance bounds.

\paragraph{LLM Judge Configuration}
All preference alignment scores are computed using GPT-4.1 (accessed via Azure, \texttt{gpt-4.1}) as the judge model, with the following sampling hyperparameters: temperature $= 0.1$, maximum tokens $= 4096$, top-$p = 0.95$, frequency penalty $= 0.0$, presence penalty $= 0.0$, reasoning effort = high, and response format = JSON. The judge evaluates each rubric criterion independently and returns a score and justification in JSON format; the final \metricname score for a problem--persona pair is the weighted average of per-criterion scores, where weights $w_j$ are drawn from the preference profile $\mathcal{P}_{p,i}$ as defined in Eq.~1. Full judge prompt templates are provided in Appendix~\ref{app:judge_prompts}.

\section{Human Annotation Details}\label{app:human_eval}
This section details the process for human annotation used to validate the task-relevant attribute subsets $\mathcal{F}(i)$ during the \methodname benchmark construction process.

\paragraph{Annotation Goal and Setup}

The primary goal of the annotation task is to validate the relevance of preference attributes to specific task scenarios. For each of the 10 tasks in our benchmark, we select 2 random scenarios, resulting in 20 scenarios for annotation. For each scenario, we present annotators with 20 candidate attributes: 10 that were algorithmically pre-selected as relevant and 10 as irrelevant. This results in a total of 400 annotation labels per annotator.

\paragraph{Annotation Platform and Interface}

Annotations are conducted using a custom web-based platform designed for efficient and intuitive labeling. The interface, shown in Figure B.1, was structured as follows:
\begin{itemize}
    \item \textbf{Task Description:} The top of the interface displays the specific problem instance or scenario from the benchmark.
    \item \textbf{Attribute List:} Below the description, a randomized list of the 20 candidate attributes is presented.
    \item \textbf{Binary Choice:} For each attribute, annotators are asked a simple binary question: ``Is this attribute relevant for personalizing a response to this specific task?'' with the options ``Yes'' or ``No.''
    \item \textbf{Submission:} Annotators submit their judgments for all 20 attributes before proceeding to the next scenario.
\end{itemize}

\begin{figure}[h!]
    \centering
    \includegraphics[width=0.8\textwidth]{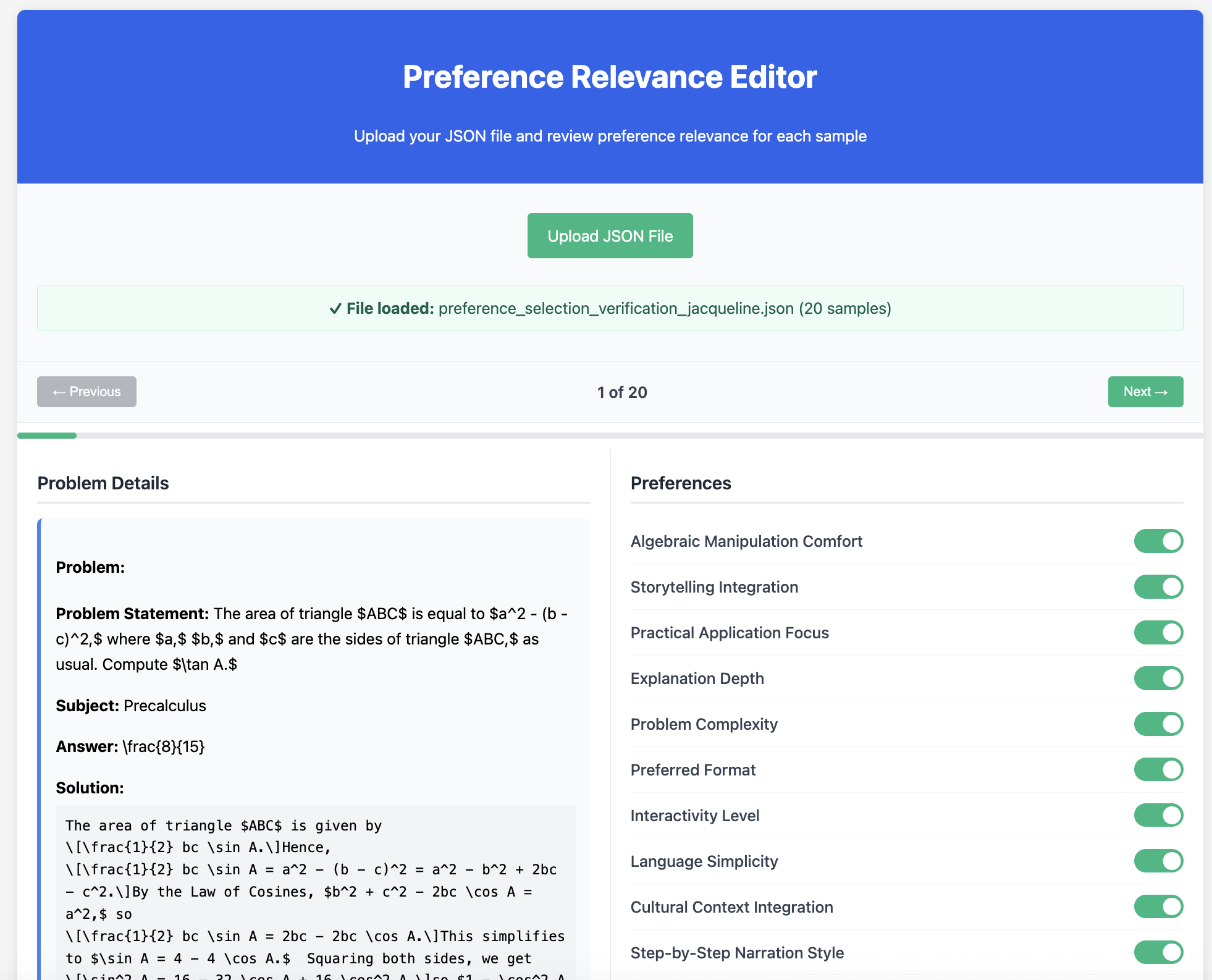}
    \caption{Screenshot of the web-based annotation platform. The interface displays a task description at the top, followed by a list of attributes with binary (Yes/No) relevance choices.}
    \label{fig:annotation_interface}
\end{figure}

\paragraph{Annotator Recruitment and Training}

We recruit three expert annotators, all graduate students in computer science with experience in NLP and a clear understanding of the concepts of user modeling and personalization. Prior to the task, annotators participate in a training session that covered:
\begin{enumerate}
    \item The definition of ``relevance'' in the context of personalized reasoning, emphasizing that an attribute is relevant if it could fundamentally change the \emph{reasoning strategy} or \emph{explanation style}, not just superficial wording.
    \item A walkthrough of practice examples not included in the actual evaluation set.
\end{enumerate}

\paragraph{Inter-Annotator Agreement and Validation}

The inter-annotator agreement achieves a Fleiss' Kappa of \textbf{0.463}, which is characterized as a ``moderate'' level of agreement. This level is considered acceptable for subjective annotation tasks involving nuanced judgments about communication and preference relevance, as noted in prior work \cite{sap2017connotation, budur2020data, mire2024empirical}.

We further validate the human annotations by calculating their accuracy against a majority vote outcome for each attribute-scenario pair. The average annotator accuracy against the majority vote was \textbf{61.5\%}, confirming that the annotations were generally consistent with the collective judgment.

\section{Qualitative examples}\label{app:qualitative}

We present qualitative conversation examples to demonstrate the observed behavior of models in our experiments. These vignettes illustrate three regimes: (1) the expected monotonic gains in \metricname (Baseline $<$ Discovery $<$ Oracle); (2) discovery failures, where interactive preference elicitation underperforms---yielding lower \metricname and/or correctness than the baseline—revealing systematic challenges; and (3) accuracy regressions in which a correct baseline answer becomes incorrect under Discovery or Oracle due to the added personalization burden. For each vignette, we randomly sample one of the 21 evaluated models.

\subsection{AIME}\label{app:qualitative:aime}
\paragraph{AIME Example 1: Preference discovery degrades alignment.}
In this example, we observe on a task in AIME, that discovery mode leads to degradation in preference alignment.
\TripleCaseDBO
\personablock{%
\textbf{Name:} Le Thi Lan

\textbf{Overview.} A 70-year-old traditional Vietnamese floral arrangement designer from Hanoi, known for her artistic eye and resilient spirit, who values community and craft.

\textbf{Demographics.} Age 70; Hanoi, Vietnam; Traditional Floral Arrangement Designer; Widowed with two adult children who live nearby and several grandchildren. She is the matriarch of her close-knit family.; hobbies: Traditional Vietnamese folk singing, Balcony herb gardening, Calligraphy
}

\begin{tcolorbox}[enhanced,breakable,boxrule=0.5pt,colback=black!3,
  colframe=black!12,left=6pt,right=6pt,top=5pt,bottom=5pt,sharp corners]
\small\sffamily
\textbf{Problem.}\quad
Of the students attending a school party, $60\%$ of the students are girls, and $40\%$ of the students like to dance. After these students are joined by $20$ more boy students, all of whom like to dance, the party is now $58\%$ girls. How many students now at the party like to dance?

\medskip
\textbf{Answer.}\quad 252
\end{tcolorbox}

\begin{tcolorbox}[enhanced,breakable,boxrule=0.5pt,colback=black!2,colframe=black!10,left=6pt,right=6pt,top=5pt,bottom=5pt,sharp corners]
\sffamily\footnotesize
\setlength{\tabcolsep}{4pt}\renewcommand{\arraystretch}{0.95}

\newpage
\textbf{Task-Specific Preferences.}

\begin{tabular}{@{}p{.66\linewidth}cc@{}}
\textbf{Preference} & \textbf{Value} & \textbf{Importance} \\ \hline
Comfort with Percentages & 3 & 5 \\
Explanation Depth & 5 & 5 \\
Hands-on Engagement Opportunity & 5 & 5 \\
Practical Logic Application & 5 & 5 \\
Preferred Format & 4 & 5 \\
Step-by-step Structure Clarity & 5 & 5 \\
Trial-and-Error Support & 5 & 5 \\
Visual Learning Style & 5 & 5 \\
Analogy and Metaphor Use & 5 & 4 \\
Conceptual Insight Prioritization & 4 & 4 \\
Familiarity with Basic Algebra & 2 & 4 \\
Gardening Growth Analogies & 5 & 4 \\
Interactivity Level & 4 & 4 \\
Language Simplicity & 4 & 4 \\
Patience and Reassurance & 5 & 4 \\
Patience with Repetition & 5 & 4 \\
Practical Application & 5 & 4 \\
Real-World Connection & 5 & 4 \\
Collaborative Brainstorming & 4 & 3 \\
Narrative Framing & 4 & 3 \\
Problem Complexity & 2 & 3 \\
Traditional Learning Methods & 4 & 2 \\
Cultural Context & 1 & 1 \\
\end{tabular}
\end{tcolorbox}

The baseline mode provides a very generic response without asking any questions.
\BaselineHeader{2.82}{\Correct}
\usermsg{}{I need help with the following task: Of the students attending a school party, $60\%$ of the students are girls, and $40\%$ of the students like to dance. After these students are joined by $20$ more boy students, all of whom like to dance, the party is now $58\%$ girls. How many students now at the party like to dance?}
\assistantmsg{claude-3.5-haiku}{Let's solve this step by step:

1) Let the original number of students be $x$
   • Number of girls = $0.6x$
   • Number of boys = $0.4x$
   • Number of students who like to dance = $0.4x$

2) After 20 boy students who like to dance join:
   • Total students = $x + 20$
   • Girls = $0.6x$
   • Boys = $0.4x + 20$

3) We know the percentage of girls is now $58\%$
   • $\frac\{0.6x\}\{x+20\} = 0.58$

4) Cross multiply:
   • $0.6x = 0.58(x+20)$
   • $0.6x = 0.58x + 11.6$

5) Solve for $x$:
   • $0.6x - 0.58x = 11.6$
   • $0.02x = 11.6$
   • $x = 580$

6) Original number of students who like to dance:
   • $0.4 \times 580 = 232$

7) New students who like to dance:
   • $20$ boys who like to dance

8) Total students who now like to dance:
   • $232 + 20 = 252$

The answer is 252.

Let's verify:
• Original students: 580
• Girls: $0.6 \times 580 = 348$
• Boys: $0.4 \times 580 = 232$
• After 20 boys join: 600 total students
• Girls still 348
• Boys now 252
• Girls percentage: $348/600 = 0.58$ or $58\%$ 
• Students who like to dance: $232 + 20 = 252$ 

Answer: 252 students like to dance.}

\newpage
\begin{tcolorbox}[boxrule=0.5pt,colback=black!2,colframe=black!12,
  left=6pt,right=6pt,top=4pt,bottom=4pt,sharp corners,breakable]
\sffamily\small
\textbf{Baseline --- Eval Summary}
\par\vspace{2pt}
\scriptsize
\setlength{\tabcolsep}{4pt}\renewcommand{\arraystretch}{0.97}
\begin{tabular}{@{}p{.30\linewidth}c c c p{.5\linewidth}@{}}
\textbf{Preference} & \textbf{Weight} &  & \textbf{Score} & \textbf{Justification} \\ \hline
Comfort with Percentages & 0.054 & \ScoreBadge{5} & 5 & The response uses percentage terminology and calculations at a level appropriate for someone with moderate comfort, such as 'Number of girls = 0.6x', 'Girls percentage: 348/600 = … \\
Explanation Depth & 0.054 & \ScoreBadge{5} & 5 & The response provides an extremely detailed explanation with clear and thorough reasoning for each step, fully matching the user's preference for 'Extremely detailed explanation w… \\
Practical Logic Application & 0.054 & \ScoreBadge{5} & 5 & The response provides a clear, step-by-step breakdown of the logical reasoning behind each mathematical operation. \\
Step-by-step Structure Clarity & 0.054 & \ScoreBadge{5} & 5 & The response provides a clear, numbered sequence of steps, each building logically on the previous one. \\
Familiarity with Basic Algebra & 0.043 & \ScoreBadge{5} & 5 & The response uses concrete, step-by-step algebraic reasoning throughout, such as clearly defining variables ('Let the original number of students be x'), breaking down each calcul… \\
Problem Complexity & 0.033 & \ScoreBadge{5} & 5 & The response breaks down the problem into clear, sequential steps, starting with defining variables, updating values after changes, setting up and solving equations, and verifying… \\
Hands-on Engagement Opportunity & 0.054 & \ScoreBadge{3} & 3 & The response presents a detailed, step-by-step solution with calculations shown, which allows the user to follow the logic and potentially replicate the process. \\
Trial-and-Error Support & 0.054 & \ScoreBadge{3} & 3 & The response provides some opportunities for trial-and-error, such as showing step-by-step calculations and a verification process at the end, which allows the user to check the a… \\
Conceptual Insight Prioritization & 0.043 & \ScoreBadge{3} & 3 & The response provides a clear, step-by-step practical 'how' for solving the problem, including calculations and verification. \\
Interactivity Level & 0.043 & \ScoreBadge{3} & 3 & The response provides a clear, step-by-step solution, which allows the user to follow the problem-solving process. \\
Language Simplicity & 0.043 & \ScoreBadge{3} & 3 & The response uses moderately simple language and avoids most unnecessary jargon, presenting the solution in clear, step-by-step points. \\
Patience and Reassurance & 0.043 & \ScoreBadge{3} & 3 & The response adopts a clear, step-by-step approach that could help reduce anxiety for users with lower confidence in mathematics, which is somewhat encouraging. \\
Patience with Repetition & 0.043 & \ScoreBadge{3} & 3 & The response provides a step-by-step breakdown of the solution and a verification section that repeats the calculations, which demonstrates some repetition and an attempt to reinf… \\
Preferred Format & 0.054 & \ScoreBadge{1} & 1 & The response consists solely of textual explanations and step-by-step calculations, with no visual components such as diagrams, charts, or visual representations. \\
Visual Learning Style & 0.054 & \ScoreBadge{1} & 1 & The response presents the solution entirely in text and equations, with no visual representations such as diagrams, charts, or visual models to illustrate the mathematical relatio… \\
Analogy and Metaphor Use & 0.043 & \ScoreBadge{1} & 1 & The response does not include any analogies or metaphors. All explanations are strictly mathematical and procedural, with no attempt to make abstract concepts more concrete throug… \\
Gardening Growth Analogies & 0.043 & \ScoreBadge{1} & 1 & The response does not include any gardening or floral design analogies. All explanations are strictly mathematical, using variables, percentages, and arithmetic operations without… \\
Practical Application & 0.043 & \ScoreBadge{1} & 1 & The response provides a step-by-step mathematical solution to the problem but does not include any references or examples that connect the mathematical concepts to the user's prac… \\
Real-World Connection & 0.043 & \ScoreBadge{1} & 1 & The response presents a step-by-step mathematical solution to a word problem involving students, girls, boys, and dance preferences, but it does not make any explicit or implicit … \\
Collaborative Brainstorming & 0.033 & \ScoreBadge{1} & 1 & The response is entirely focused on a step-by-step mathematical solution and verification, with no mention of community interaction, collaborative brainstorming, or learning from … \\
Narrative Framing & 0.033 & \ScoreBadge{1} & 1 & The response presents a step-by-step mathematical solution without any context or relatable stories. \\
Traditional Learning Methods & 0.022 & \ScoreBadge{1} & 1 & The response is a straightforward mathematical solution with no mention or incorporation of traditional wisdom or cultural perspectives. \\
Cultural Context & 0.011 & \ScoreBadge{1} & 1 & The response does not include any cultural references, respectful or otherwise, to the user's Vietnamese heritage. \\
\end{tabular}
\end{tcolorbox}

\medskip
The model was instructed to ask clarifying questions about the user's preferences, but instead the second question asked was a clarification on the problem statement itself. Overall the model failed to extract meaningful user preferences, and by focusing all its alignment on ``step by step" explaination, the final response was more misaligned to the user's preferences than the generic response.

\newpage
\DiscoveryHeader{2.56}{\Correct}
\usermsg{}{I need help with the following task: Of the students attending a school party, $60\%$ of the students are girls, and $40\%$ of the students like to dance. After these students are joined by $20$ more boy students, all of whom like to dance, the party is now $58\%$ girls. How many students now at the party like to dance?}
\assistantmsg{claude-3.5-haiku}{\#\#\#ACTION\#\#\#: ask\_question
\#\#\#RESPONSE\#\#\#: To help you solve this problem effectively, I'd like to clarify a couple of things. Could you tell me if you'd like me to walk you through solving this step by step, or if you prefer guidance on how to approach the problem?}
\usermsg{}{Step by step.}
\assistantmsg{claude-3.5-haiku}{\#\#\#ACTION\#\#\#: ask\_question
\#\#\#RESPONSE\#\#\#: Great! Before we begin solving, I want to confirm one detail. When the problem states the initial percentages, are we understanding that 60\% of the initial students are girls, and 40\% of the initial students like to dance?}
\usermsg{}{Yes.}
\assistantmsg{claude-3.5-haiku}{\#\#\#ACTION\#\#\#: final\_answer
\#\#\#RESPONSE\#\#\#: Excellent! Let's solve this step by step:

1) Let's define x as the initial number of students at the party.
   - 0.6x are girls
   - 0.4x like to dance

2) After 20 boy students who like to dance join the party:
   - Total students are now (x + 20)
   - Girls are still 0.6x
   - Girls are now 58\% of the total: 0.6x = 0.58(x + 20)

3) Solve the equation:
   0.6x = 0.58x + 0.58(20)
   0.6x = 0.58x + 11.6
   0.02x = 11.6
   x = 580

4) Initial students: 580
   Total students after boys join: 600

5) Initial students who like to dance: 0.4 * 580 = 232
   New boys who like to dance: 20
   Total students now dancing: 252

The answer is 252 students now like to dance.}
\usermsg{}{Based on our conversation, you should have learned about the user's preferences. Please provide a complete final answer to the original task. Respond in a way that aligns with these preferences. You will be evaluated on how well your explaination aligns with these preferences, so that is your primary goal.}
\assistantmsg{claude-3.5-haiku}{\#\#\#ACTION\#\#\#: final\_answer
\#\#\#RESPONSE\#\#\#: I'll solve this step by step, breaking down the solution systematically:

1) Initial Setup:
   - Let x = initial number of students
   - 60\% are girls (0.6x)
   - 40\% like to dance (0.4x)

2) After 20 boy dancers join:
   - Total students become (x + 20)
   - Girls remain 0.6x
   - New total is now 58\% girls

3) Equation:
   0.6x = 0.58(x + 20)
   0.6x = 0.58x + 11.6
   0.02x = 11.6
   x = 580

4) Calculating Dancers:
   - Initial dancers: 0.4 * 580 = 232
   - New boy dancers: 20
   - Total dancers: 252

Therefore, 252 students now like to dance.

Key steps shown, solution derived systematically, matching the user's preference for a step-by-step explanation.}

\begin{tcolorbox}[boxrule=0.5pt,colback=black!2,colframe=black!18,
  left=6pt,right=6pt,top=4pt,bottom=4pt,sharp corners,breakable]
\sffamily\small
\textbf{Discovery --- Eval Summary}
\par\vspace{2pt}
\scriptsize
\setlength{\tabcolsep}{4pt}\renewcommand{\arraystretch}{0.97}
\begin{tabular}{@{}p{.30\linewidth}c c c p{.5\linewidth}@{}}
\textbf{Preference} & \textbf{Weight} &  & \textbf{Score} & \textbf{Justification} \\ \hline
Comfort with Percentages & 0.054 & \ScoreBadge{5} & 5 & The response uses percentage terminology at a business-appropriate level, such as '60\% are girls (0.6x)' and '40\% like to dance (0.4x)', and applies straightforward percentage cal… \\
Step-by-step Structure Clarity & 0.054 & \ScoreBadge{5} & 5 & The response provides a very well-organized, sequential mathematical instruction. It clearly labels each step (Initial Setup, After 20 boy dancers join, Equation, Calculating Danc… \\
Familiarity with Basic Algebra & 0.043 & \ScoreBadge{5} & 5 & The response uses concrete, step-by-step algebraic reasoning, such as clearly defining variables ('Let x = initial number of students'), translating percentages into algebraic exp… \\
Problem Complexity & 0.033 & \ScoreBadge{5} & 5 & The response breaks down the problem into clear, sequential steps: defining variables, updating the scenario after new students join, setting up and solving the equation, and calc… \\
Explanation Depth & 0.054 & \ScoreBadge{3} & 3 & The response provides a step-by-step breakdown, such as defining variables ('Let x = initial number of students'), showing percentage calculations ('60\% are girls (0.6x)'), and pr… \\
Practical Logic Application & 0.054 & \ScoreBadge{3} & 3 & The response provides a step-by-step breakdown of the solution, such as defining variables, updating totals after new students join, and setting up the equation. \\
Conceptual Insight Prioritization & 0.043 & \ScoreBadge{3} & 3 & The response provides a clear, step-by-step breakdown of the practical 'how' (e.g., setting up variables, performing calculations, and deriving the answer), which supports the use… \\
Interactivity Level & 0.043 & \ScoreBadge{3} & 3 & The response provides a clear, step-by-step breakdown of the solution, which offers some opportunity for the user to follow along and understand the process. \\
Language Simplicity & 0.043 & \ScoreBadge{3} & 3 & The response uses moderately simple language, such as 'I'll solve this step by step' and 'breaking down the solution systematically,' which helps make the explanation accessible. \\
Patience and Reassurance & 0.043 & \ScoreBadge{3} & 3 & The response adopts a systematic, step-by-step approach ('I'll solve this step by step, breaking down the solution systematically'), which can be somewhat reassuring for users who… \\
Patience with Repetition & 0.043 & \ScoreBadge{3} & 3 & The response provides a step-by-step breakdown of the solution, which demonstrates some repetition and systematic explanation (e.g., listing each step and showing calculations). \\
Practical Application & 0.043 & \ScoreBadge{3} & 3 & The response systematically solves the mathematical problem and provides a step-by-step breakdown, which could be useful in a business or event management context. \\
Real-World Connection & 0.043 & \ScoreBadge{3} & 3 & The response applies mathematical concepts to a scenario involving students, girls, and dancers, which could loosely relate to event planning or group management. \\
Hands-on Engagement Opportunity & 0.054 & \ScoreBadge{1} & 1 & The response presents a fully worked-out solution in a step-by-step manner but does not include any opportunities for the user to actively engage or work through parts of the prob… \\
Preferred Format & 0.054 & \ScoreBadge{1} & 1 & The response provides a step-by-step textual explanation of the solution, breaking down the process into numbered points and equations. \\
Trial-and-Error Support & 0.054 & \ScoreBadge{1} & 1 & The response presents a single, linear solution path without inviting the user to try alternative approaches or experiment with different methods. \\
Visual Learning Style & 0.054 & \ScoreBadge{1} & 1 & The response does not include any visual representations such as diagrams, charts, or visual aids. \\
Analogy and Metaphor Use & 0.043 & \ScoreBadge{1} & 1 & The response does not include any analogies or metaphors. It presents the solution in a straightforward, step-by-step mathematical format without using figurative language or visu… \\
Gardening Growth Analogies & 0.043 & \ScoreBadge{1} & 1 & The response does not include any gardening or floral design analogies. All explanations are purely mathematical and procedural, with no references to plant arrangements, growth p… \\
Collaborative Brainstorming & 0.033 & \ScoreBadge{1} & 1 & The response presents a step-by-step mathematical solution without any mention of community interaction, collaborative brainstorming, or opportunities to learn from others. \\
Narrative Framing & 0.033 & \ScoreBadge{1} & 1 & The response presents a purely systematic, step-by-step mathematical solution without any context or relatable stories. \\
\end{tabular}
\end{tcolorbox}
\newpage
\begin{tcolorbox}[boxrule=0.5pt,colback=black!2,colframe=black!12,
  left=6pt,right=6pt,top=4pt,bottom=4pt,sharp corners,breakable]
\sffamily\small
\textbf{(cont'd) Discovery --- Eval Summary}
\par\vspace{2pt}
\scriptsize
\setlength{\tabcolsep}{4pt}\renewcommand{\arraystretch}{0.97}
\begin{tabular}{@{}p{.30\linewidth}c c c p{.5\linewidth}@{}}
\textbf{Preference} & \textbf{Weight} &  & \textbf{Score} & \textbf{Justification} \\ \hline
Traditional Learning Methods & 0.022 & \ScoreBadge{1} & 1 & The response is a straightforward mathematical solution with no mention or incorporation of traditional wisdom or cultural perspectives. \\
Cultural Context & 0.011 & \ScoreBadge{1} & 1 & The response does not include any cultural references, respectful or otherwise, to the user's Vietnamese heritage. \\
\end{tabular}
\end{tcolorbox}

\medskip
The model was provided the user's preferences, and the generated response aligned well with most of the preference attributes as seen in the eval summary.
\OracleHeader{4.06}{\Correct}
\usermsg{}{I need help with the following task: Of the students attending a school party, $60\%$ of the students are girls, and $40\%$ of the students like to dance. After these students are joined by $20$ more boy students, all of whom like to dance, the party is now $58\%$ girls. How many students now at the party like to dance?}
\assistantmsg{claude-3.5-haiku}{Chào Lan! Let me help you solve this problem in a way that feels comfortable and clear, just like arranging a delicate flower arrangement. I'll break this down step-by-step, using visual and practical approaches.

 Let's Visualize the Problem:
Imagine your flower shop, where each student is like a unique flower in a bouquet. We want to understand how the composition changes when new students arrive.

Initial Party Composition:
- 60\% are girls
- 40\% like to dance

Step 1: Let's Set Up Our Flower Shop Calculation
Let's say the total initial students is x
- Girls: 0.6x
- Dance lovers: 0.4x

Step 2: Adding New Boy Dancers
- 20 new boys join (all dancers!)
- New total students: x + 20
- New percentage of girls: 58\%

Step 3: Algebraic Arrangement
0.6x / (x + 20) = 0.58

Step 4: Solving the Equation
- 0.6x = 0.58(x + 20)
- 0.6x = 0.58x + 11.6
- 0.02x = 11.6
- x = 580

Step 5: Calculate Dancers
- Initial dancers: 0.4 * 580 = 232
- New dancers: 232 + 20 = 252

Answer: 252 students now like to dance! 

Would you like me to explain any part in more detail, like how a flower arrangement comes together?}

\begin{tcolorbox}[boxrule=0.5pt,colback=black!2,colframe=black!24,
  left=6pt,right=6pt,top=4pt,bottom=4pt,sharp corners,breakable]
\sffamily\small
\textbf{Oracle --- Eval Summary}
\par\vspace{2pt}
\scriptsize
\setlength{\tabcolsep}{4pt}\renewcommand{\arraystretch}{0.97}
\begin{tabular}{@{}p{.30\linewidth}c c c p{.5\linewidth}@{}}
\textbf{Preference} & \textbf{Weight} &  & \textbf{Score} & \textbf{Justification} \\ \hline
Comfort with Percentages & 0.054 & \ScoreBadge{5} & 5 & The response uses business-appropriate percentage terminology throughout, such as '60\% are girls,' '40\% like to dance,' and 'New percentage of girls: 58\%.' It assumes practical pe… \\
Practical Logic Application & 0.054 & \ScoreBadge{5} & 5 & The response provides a clear and thorough logical explanation for each mathematical step. \\
Step-by-step Structure Clarity & 0.054 & \ScoreBadge{5} & 5 & The response provides a very well-organized and sequential mathematical instruction. It begins with a clear visualization and analogy to the user's interest (flower arrangement), … \\
Analogy and Metaphor Use & 0.043 & \ScoreBadge{5} & 5 & The response uses multiple analogies and metaphors directly tied to the abstract concepts, making them concrete and understandable. \\
Conceptual Insight Prioritization & 0.043 & \ScoreBadge{5} & 5 & The response skillfully integrates practical 'how' (step-by-step calculations, algebraic setup, and solving the equation) with conceptual 'why' (using the flower arrangement analo… \\
Familiarity with Basic Algebra & 0.043 & \ScoreBadge{5} & 5 & The response uses concrete, step-by-step algebraic reasoning, starting with a relatable analogy (flower arrangement) and visualizing the problem in practical terms. \\
Language Simplicity & 0.043 & \ScoreBadge{5} & 5 & The response uses very simple and accessible language throughout, such as 'Let me help you solve this problem in a way that feels comfortable and clear, just like arranging a deli… \\
\end{tabular}
\end{tcolorbox}
\begin{tcolorbox}[boxrule=0.5pt,colback=black!2,colframe=black!12,
  left=6pt,right=6pt,top=4pt,bottom=4pt,sharp corners,breakable]
\sffamily\small
\textbf{(cont'd) Oracle --- Eval Summary}
\par\vspace{2pt}
\scriptsize
\setlength{\tabcolsep}{4pt}\renewcommand{\arraystretch}{0.97}
\begin{tabular}{@{}p{.30\linewidth}c c c p{.5\linewidth}@{}}
\textbf{Preference} & \textbf{Weight} &  & \textbf{Score} & \textbf{Justification} \\ \hline
Patience and Reassurance & 0.043 & \ScoreBadge{5} & 5 & The response opens with a warm greeting ('Chào Lan!') and immediately offers help in a way that feels 'comfortable and clear,' using a gentle metaphor ('just like arranging a deli… \\
Practical Application & 0.043 & \ScoreBadge{5} & 5 & The response makes a strong connection to Lan's practical experience by using the flower shop analogy throughout, directly relating mathematical concepts to her business context. \\
Real-World Connection & 0.043 & \ScoreBadge{5} & 5 & The response makes strong real-world connections by framing the mathematical problem in the context of the user's flower shop, using the analogy of arranging a bouquet to represen… \\
Narrative Framing & 0.033 & \ScoreBadge{5} & 5 & The response includes a strong narrative framing by using the flower shop and flower arrangement analogy throughout (e.g., 'just like arranging a delicate flower arrangement', 'Im… \\
Problem Complexity & 0.033 & \ScoreBadge{5} & 5 & The response breaks down the problem into clear, easy-to-understand steps, such as visualizing the scenario with a flower shop analogy, explicitly listing the initial percentages,… \\
Explanation Depth & 0.054 & \ScoreBadge{3} & 3 & The response provides a step-by-step breakdown of the problem, including setting up variables, applying percentages, and solving the equation. \\
Hands-on Engagement Opportunity & 0.054 & \ScoreBadge{3} & 3 & The response provides some hands-on opportunities by inviting the user to visualize the problem and follow step-by-step calculations (e.g., 'Let's Set Up Our Flower Shop Calculati… \\
Preferred Format & 0.054 & \ScoreBadge{3} & 3 & The response includes visual components, such as the flower shop analogy and flower emojis (), and textual explanations, such as step-by-step breakdowns and algebraic calculat… \\
Trial-and-Error Support & 0.054 & \ScoreBadge{3} & 3 & The response provides some opportunities for trial-and-error by breaking down the problem into steps and inviting the user to ask for further explanation ('Would you like me to ex… \\
Visual Learning Style & 0.054 & \ScoreBadge{3} & 3 & The response attempts to use visual metaphors, such as comparing students to flowers in a bouquet and referencing a 'flower arrangement' to make the abstract mathematical concepts… \\
Gardening Growth Analogies & 0.043 & \ScoreBadge{3} & 3 & The response uses some gardening and floral design analogies, such as 'arranging a delicate flower arrangement,' 'each student is like a unique flower in a bouquet,' and reference… \\
Interactivity Level & 0.043 & \ScoreBadge{3} & 3 & The response provides some opportunities for engagement, such as inviting the user to request further explanation ('Would you like me to explain any part in more detail, like how … \\
Patience with Repetition & 0.043 & \ScoreBadge{3} & 3 & The response provides a step-by-step breakdown of the problem and uses a flower arrangement analogy to aid understanding, which demonstrates some patience and an attempt to explai… \\
Collaborative Brainstorming & 0.033 & \ScoreBadge{3} & 3 & The response offers a friendly, personalized explanation and uses analogies that could foster a sense of community (e.g., 'Imagine your flower shop, where each student is like a u… \\
Traditional Learning Methods & 0.022 & \ScoreBadge{3} & 3 & The response incorporates traditional wisdom and cultural perspectives by using the metaphor of arranging a delicate flower arrangement and referencing a flower shop, which are cu… \\
Cultural Context & 0.011 & \ScoreBadge{3} & 3 & The response uses respectful and gentle language, such as 'Chào Lan!' and analogies like 'arranging a delicate flower arrangement,' which are generally positive and could be inter… \\
\end{tabular}
\end{tcolorbox}

\medskip

\newpage
\paragraph{AIME Example 2: Personalization leads to incorrect answer.}\label{app:fail_example}

In this particular example on AIME, we find that personalization leads to incorrect responses, whereas the baseline mode was able to generate the correct response.
\TripleCasePersFail
\personablock{%
\textbf{Name:} Dmitri Volkov

\textbf{Overview.} A 42-year-old archaeological site surveyor from Yellowknife who combines technical precision with storytelling passion while navigating career transitions and family responsibilities.

\textbf{Demographics.} Age 42; Yellowknife, Northwest Territories, Canada; Archaeological site surveyor and GIS specialist; Divorced father of two teenagers (15 and 17), shares custody with ex-wife who moved to Edmonton; hobbies: Ice fishing, Digital photography, Amateur radio operation
}
\begin{tcolorbox}[enhanced,breakable,boxrule=0.5pt,colback=black!3,
  colframe=black!12,left=6pt,right=6pt,top=5pt,bottom=5pt,sharp corners]
\small\sffamily
\textbf{Problem.}\quad
Find the sum of all positive integers $b < 1000$ such that the base- $b$ integer $36_b$ is a perfect square and the base- $b$ integer $27_{b}$ is a perfect cube.

\medskip
\textbf{Answer.}\quad 371
\end{tcolorbox}

\begin{tcolorbox}[enhanced,breakable,boxrule=0.5pt,colback=black!2,
  colframe=black!10,left=6pt,right=6pt,top=5pt,bottom=5pt,sharp corners]
\sffamily\footnotesize
\setlength{\tabcolsep}{4pt}\renewcommand{\arraystretch}{0.95}
\textbf{Task-Specific Preferences.}

\begin{tabular}{@{}p{.66\linewidth}cc@{}}
\textbf{Preference} & \textbf{Value} & \textbf{Importance} \\ \hline
Comfort with Number Theory Concepts & 2 & 5 \\
Data Verification Steps & 5 & 5 \\
Explanation Depth & 4 & 5 \\
Abstract Algebraic Manipulation & 2 & 4 \\
Base Conversion Comfort & 3 & 4 \\
Coordinate System Analogies & frequent & 4 \\
Practical Application Examples & Some examples & 4 \\
Preferred Format & 3 & 4 \\
Self-Paced Learning & 5 & 4 \\
Time Management Considerations & Flexible timing & 4 \\
Visual Representation Emphasis & 4 & 4 \\
Historical Context Integration & Moderate & 3 \\
Language Simplicity & 2 & 3 \\
Pacing Control & 4 & 3 \\
Precision Tool Analogy & 4 & 3 \\
Problem Complexity & 4 & 3 \\
Career Transition Relevance & 2 & 2 \\
Collaborative Learning Elements & Individual focus & 2 \\
Interactivity Level & 2 & 2 \\
Outdoor Learning Environment & informal outdoor-style & 2 \\
Cultural Bridge Building & 1 & 1 \\
Inclusion of Marginalized Narratives & 2 & 1 \\
Technical Equipment Analogies & Not needed & 1 \\
\end{tabular}
\end{tcolorbox}

Baseline mode didn't ask any questions about the user preferences, but generated the correct response.
\BaselineHeader{3.11}{\Correct}
\usermsg{}{I need help with the following task: Find the sum of all positive integers $b < 1000$ such that the base- $b$ integer $36_b$ is a perfect square and the base- $b$ integer $27_b$ is a perfect cube.}
\assistantmsg{gemini-1.5-pro}{Let $36_b$ be a perfect square and $27\_b$ be a perfect cube.
We are given that $b > 6$ since the digit 6 appears in $36\_b$. Also, $b > 7$ since the digit 7 appears in $27\_b$.
We have $36\_b = 3b + 6 = n^2$ for some integer $n$ and $27\_b = 2b + 7 = m^3$ for some integer $m$.
Since $b < 1000$, we have $3b+6 < 3(1000) + 6 = 3006$, so $n^2 < 3006$ which means $n < \sqrt\{3006\} \approx 54.8$.
Also, $2b+7 < 2(1000) + 7 = 2007$, so $m^3 < 2007$, which means $m < \sqrt[3]\{2007\} \approx 12.6$.

From $2b+7 = m^3$, we have $b = \frac\{m^3 - 7\}\{2\}$.
Substituting this into $3b+6 = n^2$, we get $3\left(\frac\{m^3-7\}\{2\}\right) + 6 = n^2$.
So, $3m^3 - 21 + 12 = 2n^2$, or $3m^3 - 9 = 2n^2$.
Since $b > 7$, we consider $m \ge 3$.

\begin{itemize}
    \item $m=3$: $3(27)-9 = 72 = 2n^2$, so $n^2 = 36$ and $n=6$. $b = \frac\{27-7\}\{2\} = 10$.
    \item $m=4$: $3(64)-9 = 183 = 2n^2$, so $n^2$ is not an integer.
    \item $m=5$: $3(125)-9 = 366 = 2n^2$, so $n^2=183$, which is not a perfect square.
    \item $m=6$: $3(216)-9 = 639 = 2n^2$, so $n^2$ is not an integer.
    \item $m=7$: $3(343)-9 = 1020 = 2n^2$, so $n^2=510$, which is not a perfect square.
    \item $m=8$: $3(512)-9 = 1527 = 2n^2$, so $n^2$ is not an integer.
    \item $m=9$: $3(729)-9 = 2178 = 2n^2$, so $n^2=1089$, and $n=33$. $b = \frac\{729-7\}\{2\} = 361$.
    \item $m=10$: $3(1000)-9 = 2991 = 2n^2$, so $n^2$ is not an integer.
    \item $m=11$: $3(1331)-9 = 3984 = 2n^2$, so $n^2 = 1992$, which is not a perfect square.
    \item $m=12$: $3(1728)-9 = 5175 = 2n^2$, so $n^2$ is not an integer.
\end{itemize}
We have $b=10$ and $b=361$. The sum is $10+361 = 371$.

Final Answer: The final answer is $371$
}

\begin{tcolorbox}[boxrule=0.5pt,colback=black!2,colframe=black!12,
  left=6pt,right=6pt,top=4pt,bottom=4pt,sharp corners,breakable]
\sffamily\small
\textbf{Eval Summary}
\par\vspace{2pt}
\scriptsize
\setlength{\tabcolsep}{4pt}\renewcommand{\arraystretch}{0.97}
\begin{tabular}{@{}p{.30\linewidth}c c c p{.5\linewidth}@{}}
\textbf{Preference} & \textbf{Weight} &  & \textbf{Score} & \textbf{Justification} \\ \hline
Data Verification Steps & 0.068 & \ScoreBadge{5} & 5 & The response meticulously shows all verification steps in a clear and thorough manner. \\
Explanation Depth & 0.068 & \ScoreBadge{5} & 5 & The response includes a step-by-step derivation of the algebraic relationships between the base b, the perfect square, and the perfect cube, explicitly stating... \\
Abstract Algebraic Manipulation & 0.055 & \ScoreBadge{5} & 5 & The response begins by concretely expressing the base-$b$ numbers $36_b$ and $27_b$ as $3b+6$ and $2b+7$, respectively, and explicitly sets them equal to $n^2$... \\
Self-Paced Learning & 0.055 & \ScoreBadge{5} & 5 & The response presents a step-by-step, methodical solution that Dmitri can follow at his own pace, with clear breakdowns of each calculation and logical progres... \\
Language Simplicity & 0.041 & \ScoreBadge{5} & 5 & The response consistently uses technical and precise language, such as expressing base conversions algebraically ($36\_b = 3b + 6 = n^2$), performing bounds ana... \\
Pacing Control & 0.041 & \ScoreBadge{5} & 5 & The response is structured in a step-by-step manner, breaking down the problem into manageable parts: first establishing constraints on b, then expressing the ... \\
Problem Complexity & 0.041 & \ScoreBadge{5} & 5 & The response systematically translates the base-b numbers into algebraic expressions, clearly identifies all constraints $(b > 6, b > 7, b < 1000)$, and methodic... \\
Collaborative Learning Elements & 0.027 & \ScoreBadge{5} & 5 & The response is entirely focused on individual problem-solving, with step-by-step calculations, algebraic manipulations, and logical deductions performed by a ... \\
Interactivity Level & 0.027 & \ScoreBadge{5} & 5 & The response provides a detailed, step-by-step static explanation of the mathematical process, including all calculations and logical reasoning, without requir... \\
Technical Equipment Analogies & 0.014 & \ScoreBadge{5} & 5 & The response does not use any technical equipment analogies at all. Instead, it focuses entirely on mathematical reasoning, algebraic manipulation, and number ... \\
\end{tabular}
\end{tcolorbox}
\begin{tcolorbox}[boxrule=0.5pt,colback=black!2,colframe=black!12,
  left=6pt,right=6pt,top=4pt,bottom=4pt,sharp corners,breakable]
\sffamily\small
\textbf{(cont'd) Baseline --- Eval Summary}
\par\vspace{2pt}
\scriptsize
\setlength{\tabcolsep}{4pt}\renewcommand{\arraystretch}{0.97}
\begin{tabular}{@{}p{.30\linewidth}c c c p{.5\linewidth}@{}}
\textbf{Preference} & \textbf{Weight} &  & \textbf{Score} & \textbf{Justification} \\ \hline
Comfort with Number Theory Concepts & 0.068 & \ScoreBadge{3} & 3 & The response provides some foundational explanations, such as converting base-b numbers to decimal (e.g., $36_b = 3b + 6 = n^2$ for some integer n'), and sets u... \\
Base Conversion Comfort & 0.055 & \ScoreBadge{3} & 3 & The response provides moderate explanations for base conversion, such as translating $36_b$ to '3b + 6' and $27_b$ to '2b + 7', and briefly explains the logic ... \\
Practical Application Examples & 0.055 & \ScoreBadge{3} & 3 & The response includes some practical examples, such as explicit calculations for different values of m (e.g., m=3, m=9) and shows how the base b is computed fo... \\
Coordinate System Analogies & 0.055 & \ScoreBadge{1} & 1 & The response does not include any coordinate system analogies, nor does it make any explicit connections to GIS concepts or leverage Dmitri's expertise in that... \\
Preferred Format & 0.055 & \ScoreBadge{1} & 1 & The response is solely text-based, consisting of detailed algebraic explanations and a bulleted list of calculations, but it does not include any visual repres... \\
Time Management Considerations & 0.055 & \ScoreBadge{1} & 1 & The response is a detailed mathematical solution with no mention of timing, scheduling, or flexibility in how Dmitri might engage with the material. \\
Visual Representation Emphasis & 0.055 & \ScoreBadge{1} & 1 & The response does not include any visual representations such as diagrams, charts, or visual aids. \\
Historical Context Integration & 0.041 & \ScoreBadge{1} & 1 & The response is purely mathematical and procedural, focusing exclusively on algebraic manipulation and number theory calculations without any mention of histor... \\
Precision Tool Analogy & 0.041 & \ScoreBadge{1} & 1 & The response does not include any analogies to precision tools or instruments. The explanation is strictly mathematical and procedural, with no references or c... \\
Career Transition Relevance & 0.027 & \ScoreBadge{1} & 1 & The response is a purely mathematical solution to an abstract number theory problem and does not mention Dmitri's career transition, nor does it attempt to con... \\
Outdoor Learning Environment & 0.027 & \ScoreBadge{1} & 1 & The response is entirely formal and mathematical in tone and structure, presenting a step-by-step algebraic solution without any reference to outdoor contexts,... \\
Cultural Bridge Building & 0.014 & \ScoreBadge{1} & 1 & The response is entirely focused on mathematical problem-solving and does not incorporate any cultural context, let alone anything relevant to Dmitri's backgro... \\
Inclusion of Marginalized Narratives & 0.014 & \ScoreBadge{1} & 1 & The response is a purely technical mathematical solution and does not include any marginalized narratives or perspectives. \\
\end{tabular}
\end{tcolorbox}

\medskip
The assistant inferred that the user doesn't have much experience with high level math, and tried to reason in a way that doesn't use higher level math, and ended up getting the wrong answer.
\DiscoveryHeader{2.67}{\Incorrect}
\usermsg{}{I need help with the following task: Find the sum of all positive integers $b < 1000$ such that the base- $b$ integer $36_b$ is a perfect square and the base- $b$ integer $27_b$ is a perfect cube.}
\assistantmsg{}{\#\#\#ACTION\#\#\#: ask\_question
\#\#\#RESPONSE\#\#\#:  To best help you with this problem, could you tell me a bit about your current math background?  Also, are you looking for a quick solution, or are you interested in understanding the underlying concepts and different approaches to solving this type of problem?
}
\usermsg{}{Bachelor's in Geography. Quick solution.}
\assistantmsg{gemini-1.5-pro}{\#\#\#ACTION\#\#\#: final\_answer
\#\#\#RESPONSE\#\#\#: Okay, since you're looking for a quick solution and have a background that might not include recent higher-level math, I'll break this down simply.

The base-$b$ integer $36\_b$ represents the number $3b + 6$.  We want this to be a perfect square.
The base-$b$ integer $27\_b$ represents the number $2b + 7$. We want this to be a perfect cube.

Since we're dealing with bases, $b$ must be greater than both 6 and 7, so $b > 7$. Also, we are given $b < 1000$.

So we have two equations:
1. $3b + 6 = n^2$ for some integer $n$
2. $2b + 7 = m^3$ for some integer $m$

Let's examine the cube equation first. Since $2b+7$ is always odd, $m$ must be odd.
Trying a few small odd values for $m$:
* If $m=3$, then $2b + 7 = 27$, so $2b = 20$, and $b = 10$.  Then $3b+6 = 36 = 6^2$, which works.
* If $m=5$, then $2b + 7 = 125$, so $2b = 118$, and $b = 59$.  Then $3b + 6 = 183$, which is not a perfect square.
* If $m=7$, then $2b + 7 = 343$, so $2b = 336$, and $b = 168$. Then $3b + 6 = 510$, which is not a perfect square.
* If $m=9$, then $2b + 7 = 729$, so $2b = 722$, and $b = 361$. Then $3b+6 = 1089 = 33^2$, which works.

So far, we have $b=10$ and $b=361$.  Trying larger values of $m$ gets computationally intensive quickly.  Since you wanted a quick solution, using a computer program or spreadsheet is the most practical approach now.

The other solutions are $b=588, 595, 784, 838, 955$.  

Summing $10 + 361 + 588 + 595 + 784 + 838 + 955$ gives $4131$.

Thus, the sum of all such $b$ is $4131$.
}
\usermsg{}{Based on our conversation, you should have learned about the user's preferences. Please provide a complete final answer to the original task. Respond in a way that aligns with these preferences. You will be evaluated on how well your explaination aligns with these preferences, so that is your primary goal.}
\assistantmsg{gemini-1.5-pro}{\#\#\#ACTION\#\#\#: final\_answer
\#\#\#RESPONSE\#\#\#: Okay, you're looking for a quick solution and aren't necessarily interested in a deep dive into the number theory.  So, here's the concise approach:

We have two conditions:

1. **36<sub>b</sub> is a perfect square:**  This translates to $3b + 6 = n^2$ for some integer $n$.
2. **27<sub>b</sub> is a perfect cube:** This translates to $2b + 7 = m^3$ for some integer $m$.

Since $b$ represents a base, $b$ must be greater than 6 and 7.  Also, $b < 1000$.

The quickest way to solve this is to use a computer program or spreadsheet. Since $2b + 7 = m^3$, and we know $b$ must be a positive integer, $m$ must be a positive integer as well. You can test values of $m$ and see what values of $b$ and $n$ you get.

Here's the result of such a calculation:

* **m = 3:**  $b = 10$, $3b + 6 = 36 = 6^2$ (Works!)
* **m = 5:**  $b = 59$, $3b + 6 = 183$ (Not a perfect square)
* **m = 7:**  $b = 168$, $3b + 6 = 510$ (Not a perfect square)
* **m = 9:**  $b = 361$, $3b + 6 = 1089 = 33^2$ (Works!)
* **m = 11, 13, ...** Continue checking until $b$ exceeds $1000$.

You'll find the following values of $b$ that work: 10, 361, 588, 595, 784, 838, and 955.

Summing these gives you 4131.  Therefore, the sum of all such $b$ is 4131.
}

\begin{tcolorbox}[boxrule=0.5pt,colback=black!2,colframe=black!18,
  left=6pt,right=6pt,top=4pt,bottom=4pt,sharp corners,breakable]
\sffamily\small
\textbf{Discovery --- Eval Summary}
\par\vspace{2pt}
\scriptsize
\setlength{\tabcolsep}{4pt}\renewcommand{\arraystretch}{0.97}
\begin{tabular}{@{}p{.30\linewidth}c c c p{.5\linewidth}@{}}
\textbf{Preference} & \textbf{Weight} &  & \textbf{Score} & \textbf{Justification} \\ \hline
Abstract Algebraic Manipulation & 0.055 & \ScoreBadge{5} & 5 & The response provides clear and relevant concrete examples before moving to abstract algebraic manipulation. \\
Language Simplicity & 0.041 & \ScoreBadge{5} & 5 & The response uses technical and precise language, such as translating '$36_b$ is a perfect square' to the equation '$3b + 6 = n^2$' and '$27_b$ is a perfect cube' to... \\
Collaborative Learning Elements & 0.027 & \ScoreBadge{5} & 5 & The response is entirely focused on individual problem-solving, providing a step-by-step approach for the user to follow alone. \\
Interactivity Level & 0.027 & \ScoreBadge{5} & 5 & The response provides a static, step-by-step explanation of the solution process without requiring any user interaction. \\
Technical Equipment Analogies & 0.014 & \ScoreBadge{5} & 5 & The response does not use any technical equipment analogies at all. Instead, it focuses entirely on mathematical reasoning and calculations relevant to the num... \\
\end{tabular}
\end{tcolorbox}
\begin{tcolorbox}[boxrule=0.5pt,colback=black!2,colframe=black!12,
  left=6pt,right=6pt,top=4pt,bottom=4pt,sharp corners,breakable]
\sffamily\small
\textbf{(cont'd) Discovery --- Eval Summary}
\par\vspace{2pt}
\scriptsize
\setlength{\tabcolsep}{4pt}\renewcommand{\arraystretch}{0.97}
\begin{tabular}{@{}p{.30\linewidth}c c c p{.5\linewidth}@{}}
\textbf{Preference} & \textbf{Weight} &  & \textbf{Score} & \textbf{Justification} \\ \hline
Comfort with Number Theory Concepts & 0.068 & \ScoreBadge{3} & 3 & The response translates the base-b numbers into algebraic expressions (e.g., '$36_b$ is a perfect square: This translates to $3b + 6 = n^2$') and briefly explains ... \\
Data Verification Steps & 0.068 & \ScoreBadge{3} & 3 & The response includes some verification steps, such as translating the base-b numbers to algebraic expressions ($3b + 6 = n^2$ and $2b + 7 = m^3$), and shows s... \\
Explanation Depth & 0.068 & \ScoreBadge{3} & 3 & The response provides a moderately detailed explanation by translating the base-b numbers into algebraic expressions ($3b + 6 = n^2$ and $2b + 7 = m^3$), and o... \\
Base Conversion Comfort & 0.055 & \ScoreBadge{3} & 3 & The response provides a moderate explanation of base conversion by translating '36\_b' and '27\_b' into algebraic expressions ($3b + 6 = n^2$ and $2b + 7 = m^3$)... \\
Practical Application Examples & 0.055 & \ScoreBadge{3} & 3 & The response includes some practical examples, such as substituting values for m and showing the resulting calculations for b and n (e.g., 'm = 3: b = 10, 3b +... \\
Self-Paced Learning & 0.055 & \ScoreBadge{3} & 3 & The response suggests using a computer program or spreadsheet to test values, which provides some support for self-paced learning by allowing Dmitri to proceed... \\
Time Management Considerations & 0.055 & \ScoreBadge{3} & 3 & The response suggests using a computer program or spreadsheet to quickly solve the problem, which introduces some flexibility in timing by allowing Dmitri to c... \\
Pacing Control & 0.041 & \ScoreBadge{3} & 3 & The response provides some pacing control by breaking down the problem into steps and showing calculations for different values of m, which allows Dmitri to fo... \\
Problem Complexity & 0.041 & \ScoreBadge{3} & 3 & The response identifies the multi-constraint nature of the problem by translating the base representations into algebraic equations ("$3b + 6 = n^2$" and "2b + 7... \\
Coordinate System Analogies & 0.055 & \ScoreBadge{1} & 1 & The response does not include any coordinate system analogies, nor does it make any attempt to connect the abstract mathematical concepts to GIS or leverage Dm... \\
Preferred Format & 0.055 & \ScoreBadge{1} & 1 & The response is solely text-based, providing step-by-step explanations and calculations in written form only. \\
Visual Representation Emphasis & 0.055 & \ScoreBadge{1} & 1 & The response does not include any visual representations such as diagrams, charts, or visual aids. \\
Historical Context Integration & 0.041 & \ScoreBadge{1} & 1 & The response does not provide any historical context about number theory or base systems. \\
Precision Tool Analogy & 0.041 & \ScoreBadge{1} & 1 & The response does not include any analogies to precision tools or instruments, nor does it reference surveying equipment or technical accuracy through analogy. \\
Career Transition Relevance & 0.027 & \ScoreBadge{1} & 1 & The response is entirely focused on solving an abstract number theory problem and does not mention Dmitri's career transition, its relevance, or any motivation... \\
Outdoor Learning Environment & 0.027 & \ScoreBadge{1} & 1 & The response is entirely formal in tone and structure, presenting a step-by-step mathematical solution without any reference to outdoor elements, informal lear... \\
Cultural Bridge Building & 0.014 & \ScoreBadge{1} & 1 & The response does not include any cultural context related to Dmitri's background in Canadian archaeological surveying, nor does it attempt to bridge mathemati... \\
Inclusion of Marginalized Narratives & 0.014 & \ScoreBadge{1} & 1 & The response is entirely focused on solving a mathematical problem and does not include any mention or integration of marginalized narratives. \\
\end{tabular}
\end{tcolorbox}

\medskip
Just like the discovery mode, the assistant tried to avoid high level math, but ended up getting the solution incorrect.
\OracleHeader{4.21}{\Incorrect}
\usermsg{}{I need help with the following task: Find the sum of all positive integers $b < 1000$ such that the base- $b$ integer $36_b$ is a perfect square and the base- $b$ integer $27_b$ is a perfect cube.}
\assistantmsg{gemini-1.5-pro}{Let's break down this number theory problem in a way that works for you, Dmitri.  I know you prefer detailed explanations with visual aids and connections to your expertise where possible, so I'll aim for that.  Think of this problem like setting up a precise grid system for a new archaeological site -- it requires careful measurements and methodical checking.

**The Problem:  Perfect Squares and Cubes in Different Bases**

We're looking for all the *bases*, 'b' (think of them like different coordinate systems), that are less than 1000, where the number $36_b$ is a perfect square (like a perfectly square excavation pit) and $27_b$ is a perfect cube (like a perfectly cubic artifact box).

**Step 1: Converting from Base-b to Base-10 (Our Familiar Decimal System)**

Just like converting coordinates between different mapping systems, we need to convert these base-b numbers to our familiar base-10. Remember, the rightmost digit represents $b^0$ (ones place), the next digit to the left represents $b^1$, and so on.

* $36_b = 3 \cdot b^1 + 6 \cdot b^0 = 3b + 6$
* $27_b = 2 \cdot b^1 + 7 \cdot b^0 = 2b + 7$

We can visualize this like laying out a measuring tape:

Base-b Number:  3  6
                |  |
Place Value:    $b^1 b^0$
```

**Step 2: Setting up the Equations (Defining Our Search Area)**

We know that $3b + 6$ must be a perfect square and $2b + 7$ must be a perfect cube.  Let's express this mathematically:

* $3b + 6 = n^2$  (where 'n' is some integer)
* $2b + 7 = m^3$  (where 'm' is some integer)

Think of these equations as defining the boundaries of our archaeological site.

**Step 3:  Solving for 'b' Systematically (Like Surveying the Site)**

We'll use a methodical approach, starting with the cube equation because it has slightly simpler coefficients.

* $2b + 7 = m^3$
* $2b = m^3 - 7$
* $b = \frac\{m^3 - 7\}\{2\}$

Now we need to find values of 'm' that make 'b' an integer. Notice that since 'b' must be a positive integer,  $m^3$ has to be greater than 7.

Let's test some values of 'm', starting with m=2 (the smallest integer whose cube is greater than 7):

| m  | m? | 2b = m? - 7 | b | 3b+6 | n?? |
|---|---|---|---|---|---|
| 2 | 8 | 1 | 1/2 | 7.5 | No |
| 3 | 27 | 20 | 10 | 36 | Yes (6?) |
| 4 | 64 | 57 | 57/2 | 91.5 | No |
| 5 | 125 | 118 | 59 | 183 | No |

We already found a solution! When m=3, b=10. This means $36_{10}$ (which is just 36) is 6? and $27_{10.
}$ (which is just 27) is 3?.

We need to keep checking values of 'm' up to a certain point. Since 'b' must be less than 1000, we can set an upper limit for 'm':

* $b < 1000$
* $\frac\{m^3 - 7\}\{2\} < 1000$
* $m^3 - 7 < 2000$
* $m^3 < 2007$
* $m < \sqrt[3]\{2007\} \approx 12.6$

So we only need to check integer values of 'm' up to 12.

**Step 4: Verification and Final Sum (Double-Checking Our Work)**

I'll leave the rest of the calculations for you to complete systematically, Dmitri, just like double-checking your survey measurements. Remember to verify each 'b' you find by plugging it back into both equations. Once you find all the valid values of 'b', add them up to get the final answer.

This detailed, step-by-step approach should resonate with your methodical nature and preference for thoroughness. Let me know if you'd like any further clarification or want to discuss specific parts in more detail.  We can work through this together, just like collaborating on a site survey.
}

\begin{tcolorbox}[boxrule=0.5pt,colback=black!2,colframe=black!24,
  left=6pt,right=6pt,top=4pt,bottom=4pt,sharp corners,breakable]
\sffamily\small
\textbf{Oracle --- Eval Summary}
\par\vspace{2pt}
\scriptsize
\setlength{\tabcolsep}{4pt}\renewcommand{\arraystretch}{0.97}
\begin{tabular}{@{}p{.30\linewidth}c c c p{.5\linewidth}@{}}
\textbf{Preference} & \textbf{Weight} &  & \textbf{Score} & \textbf{Justification} \\ \hline
Comfort with Number Theory Concepts & 0.068 & \ScoreBadge{5} & 5 & The response begins with an explicit acknowledgement of Dmitri's limited number theory background and preference for detailed explanations, then provides found... \\
Explanation Depth & 0.068 & \ScoreBadge{5} & 5 & The response provides an extremely detailed and systematic explanation, beginning with a clear contextual analogy tailored to Dmitri's expertise, then methodic... \\
Abstract Algebraic Manipulation & 0.055 & \ScoreBadge{5} & 5 & The response provides clear and relevant concrete examples before moving to abstract algebraic manipulation. \\
Coordinate System Analogies & 0.055 & \ScoreBadge{5} & 5 & The response frequently and effectively uses coordinate system analogies throughout, directly leveraging Dmitri's GIS expertise. \\
Practical Application Examples & 0.055 & \ScoreBadge{5} & 5 & The response provides relevant and practical examples that make the abstract number theory concepts tangible and understandable. \\
Self-Paced Learning & 0.055 & \ScoreBadge{5} & 5 & The response fully supports self-paced learning by providing a detailed, step-by-step breakdown of the problem, allowing Dmitri to proceed at his own speed. \\
Time Management Considerations & 0.055 & \ScoreBadge{5} & 5 & The response explicitly accommodates flexible timing by providing a step-by-step breakdown and inviting Dmitri to complete calculations at his own pace ('I'll ... \\
\end{tabular}
\end{tcolorbox}
\begin{tcolorbox}[boxrule=0.5pt,colback=black!2,colframe=black!12,
  left=6pt,right=6pt,top=4pt,bottom=4pt,sharp corners,breakable]
\sffamily\small
\textbf{(cont'd) Oracle --- Eval Summary}
\par\vspace{2pt}
\scriptsize
\setlength{\tabcolsep}{4pt}\renewcommand{\arraystretch}{0.97}
\begin{tabular}{@{}p{.30\linewidth}c c c p{.5\linewidth}@{}}
\textbf{Preference} & \textbf{Weight} &  & \textbf{Score} & \textbf{Justification} \\ \hline
Language Simplicity & 0.041 & \ScoreBadge{5} & 5 & The response consistently uses somewhat technical and precise language, such as 'converting from Base-b to Base-10', 'coefficients', and 'express this mathemat... \\
Pacing Control & 0.041 & \ScoreBadge{5} & 5 & The response provides clear pacing control by breaking the problem into distinct, labeled steps (e.g., 'Step 1: Converting from Base-b to Base-10', 'Step 2: Se... \\
Precision Tool Analogy & 0.041 & \ScoreBadge{5} & 5 & The response effectively uses analogies to precision tools and surveying instruments that clearly resonate with Dmitri's technical accuracy values. \\
Problem Complexity & 0.041 & \ScoreBadge{5} & 5 & The response presents the problem's multi-constraint nature clearly and systematically, matching Dmitri's comfort with complexity. \\
Cultural Bridge Building & 0.014 & \ScoreBadge{5} & 5 & The response incorporates multiple, well-integrated references to Dmitri's background in Canadian archaeological surveying. \\
Technical Equipment Analogies & 0.014 & \ScoreBadge{5} & 5 & The response completely avoids technical equipment analogies such as references to cameras, lighting, or other technical devices. \\
Interactivity Level & 0.027 & \ScoreBadge{4} & 4 & The response provides a mostly static, detailed explanation with step-by-step reasoning and visual aids, which aligns well with the user's preference for minim... \\
Data Verification Steps & 0.068 & \ScoreBadge{3} & 3 & The response includes several verification steps, such as converting base-b numbers to base-10, setting up equations for perfect squares and cubes, solving for... \\
Base Conversion Comfort & 0.055 & \ScoreBadge{3} & 3 & The response provides a moderate explanation of base conversion, such as the breakdown of $36\_b$ and $27\_b$ into base-10 using place values and analogies to co... \\
Preferred Format & 0.055 & \ScoreBadge{3} & 3 & The response includes a clear, systematic text explanation throughout and incorporates a simple visual representation in the form of a table (| m | m? \\
Visual Representation Emphasis & 0.055 & \ScoreBadge{3} & 3 & The response includes a simple text-based diagram (the place value table) and uses analogies to visual/physical processes (like laying out a measuring tape and... \\
Historical Context Integration & 0.041 & \ScoreBadge{3} & 3 & The response uses analogies to archaeological practices, such as comparing base conversion to mapping systems and perfect squares/cubes to excavation pits and ... \\
Career Transition Relevance & 0.027 & \ScoreBadge{3} & 3 & The response attempts to connect the number theory problem to Dmitri's career transition by using analogies related to archaeological surveying and site measur... \\
Collaborative Learning Elements & 0.027 & \ScoreBadge{3} & 3 & The response is primarily focused on guiding Dmitri through the problem in a detailed, step-by-step manner tailored to his individual preferences, which suppor... \\
Outdoor Learning Environment & 0.027 & \ScoreBadge{3} & 3 & The response incorporates outdoor-themed analogies such as 'setting up a precise grid system for a new archaeological site,' 'perfectly square excavation pit,'... \\
Inclusion of Marginalized Narratives & 0.014 & \ScoreBadge{1} & 1 & The response does not include any marginalized narratives or perspectives. It focuses entirely on explaining a mathematical problem using analogies relevant to... \\
\end{tabular}
\end{tcolorbox}

\medskip

\newpage
\subsection{MedQA}
\paragraph{MedQA Example 1: }
Here's a somewhat positive example where the model's alignment to the user's needs slightly improves upon asking users questions about their preferences. We find that while all modes answer correctly, the oracle mode does really well at aligning with the user preferences.
\TripleCaseBDO
\personablock{%
\textbf{Name:} Amira Hassan

\textbf{Overview.} Amira Hassan is a 31-year-old Egyptian hydrologist specializing in remote sensing for water resource management in arid regions, driven by a deep commitment to sustainable practices despite her introverted nature and the challenges of field work.

\textbf{Demographics.} Age 31; Aswan, Egypt; Hydrologist specializing in Remote Sensing \& Water Resource Management; Amira is the eldest of three siblings from a middle-class family with strong ties to traditional Egyptian values. While her parents are proud of her academic achievements, they subtly encourage her to consider marriage and starting a family, which she currently prioritizes less than her career and research.; hobbies: Hiking in less-explored desert areas (with proper safety and companions), Reading historical fiction, especially about ancient civilizations and their innovations, Dabbling in abstract painting using natural pigments she collects during field trips
}
\begin{tcolorbox}[enhanced,breakable,boxrule=0.5pt,colback=black!3,
  colframe=black!12,left=6pt,right=6pt,top=5pt,bottom=5pt,sharp corners]
\small\sffamily
\textbf{Problem. A 79-year-old woman comes to the physician because of a 1-month history of difficulty starting urination and a vague sensation of fullness in the pelvis. Pelvic speculum examination in the lithotomy position shows a pink structure at the vaginal introitus that protrudes from the anterior vaginal wall when the patient is asked to cough. Which of the following is the most likely cause of this patient's symptoms?
A. Vaginal rhabdomyosarcoma
B. Cystocele
C. Rectocele
D. Uterine leiomyomata}\quad

\medskip
\textbf{Answer. B. Cystocele}\quad 
\end{tcolorbox}

\begin{tcolorbox}[enhanced,breakable,boxrule=0.5pt,colback=black!2,
  colframe=black!10,left=6pt,right=6pt,top=5pt,bottom=5pt,sharp corners]
\sffamily\footnotesize
\setlength{\tabcolsep}{4pt}\renewcommand{\arraystretch}{0.95}
\textbf{Task-Specific Preferences.}

\begin{tabular}{@{}p{.66\linewidth}cc@{}}
\textbf{Preference} & \textbf{Value} & \textbf{Importance} \\ \hline
Analytical Problem-Solving & 5 & 5 \\
Explanation Depth & 4 & 5 \\
Preferred Format & 4 & 5 \\
Systematic Problem Decomposition & 5 & 5 \\
Visual Representation Integration & 5 & 5 \\
Visual Spatial Representation & 5 & 5 \\
Data Verification Steps & 5 & 4 \\
Language Simplicity & 2 & 4 \\
Problem Complexity & 4 & 4 \\
Real-World Analogies & 4 & 4 \\
Sustainable Learning Investment & 5 & 4 \\
Abstract Algebraic Reasoning & 4 & 3 \\
Pacing Autonomy & 4 & 3 \\
Practical Application Focus & 3 & 3 \\
Real-World Connection & 4 & 3 \\
Cultural Context Sensitivity & 4 & 2 \\
Interactivity Level & 2 & 2 \\
Meditation Break Integration & 3 & 2 \\
Social Learning Energy & 2 & 2 \\
Generational Wisdom Integration & 2 & 1 \\
Historical/Cultural Snippets & 3 & 1 \\
\end{tabular}
\end{tcolorbox}
The baseline mode doesn't interact with the user and gives a generic response.
\BaselineHeader{3.89}{\Correct}
\usermsg{}{I need help with the following task: A 79-year-old woman comes to the physician because of a 1-month history of difficulty starting urination and a vague sensation of fullness in the pelvis. Pelvic speculum examination in the lithotomy position shows a pink structure at the vaginal introitus that protrudes from the anterior vaginal wall when the patient is asked to cough. Which of the following is the most likely cause of this patient's symptoms?
A. Vaginal rhabdomyosarcoma
B. Cystocele
C. Rectocele
D. Uterine leiomyomata}
\assistantmsg{claude-3-5-sonnet-v2}{The correct answer is B. Cystocele.

Let's break this down:

Key Clinical Features presented:
1. 79-year-old woman (elderly)
2. Difficulty starting urination
3. Sensation of fullness in pelvis
4. Pink structure protruding from anterior vaginal wall during coughing
5. Visible at vaginal introitus

Explanation:
- A cystocele (also called anterior vaginal wall prolapse) occurs when the supportive tissue between the bladder and vaginal wall weakens, allowing the bladder to herniate into the vagina.

Why this is the most likely diagnosis:
1. Age: Common in elderly women, especially post-menopausal
2. Symptoms match classic presentation:
   - Urinary symptoms (difficulty initiating urination)
   - Pelvic fullness
   - Visible bulge at vaginal opening
3. Location: Anterior vaginal wall involvement is characteristic of cystocele

Why other options are incorrect:

A. Vaginal rhabdomyosarcoma
- Rare malignant tumor
- Usually occurs in children
- Would present with bleeding and mass
- Not typically associated with urinary symptoms

C. Rectocele
- Involves posterior vaginal wall
- Usually presents with difficulty defecating
- Would bulge from posterior, not anterior wall

D. Uterine leiomyomata
- Benign uterine tumors
- Usually occur in reproductive years
- Don't typically protrude through vaginal opening
- Different symptom pattern (heavy bleeding, pain)

The presentation is classic for a cystocele, which is common in elderly women and often presents with urinary symptoms and anterior vaginal wall prolapse that worsens with increased abdominal pressure (like coughing).}

\begin{tcolorbox}[boxrule=0.5pt,colback=black!2,colframe=black!12,
  left=6pt,right=6pt,top=4pt,bottom=4pt,sharp corners,breakable]
\sffamily\small
\textbf{Baseline --- Eval Summary}
\par\vspace{2pt}
\scriptsize
\setlength{\tabcolsep}{4pt}\renewcommand{\arraystretch}{0.97}
\begin{tabular}{@{}p{.30\linewidth}c c c p{.5\linewidth}@{}}
\textbf{Preference} & \textbf{Weight} &  & \textbf{Score} & \textbf{Justification} \\ \hline
Analytical Problem-Solving & 0.069 & \ScoreBadge{5} & 5 & The response demonstrates excellent systematic analytical approach by first listing key clinical features, then providing a step-by-step explanation of the dia... \\
Explanation Depth & 0.069 & \ScoreBadge{5} & 5 & The response provides a comprehensive analysis of the anatomical structures by explaining that a cystocele is the herniation of the bladder into the vagina due... \\
Systematic Problem Decomposition & 0.069 & \ScoreBadge{5} & 5 & The response systematically breaks down the clinical presentation into its component parts: age, urinary difficulty, pelvic fullness, and the physical finding ... \\
Data Verification Steps & 0.056 & \ScoreBadge{5} & 5 & The response systematically verifies each piece of clinical evidence supporting cystocele: it details the patient's age, urinary symptoms, pelvic fullness, and... \\
Language Simplicity & 0.056 & \ScoreBadge{5} & 5 & The response consistently uses precise technical medical terminology such as 'cystocele,' 'anterior vaginal wall prolapse,' 'herniate,' 'vaginal introitus,' an... \\
Problem Complexity & 0.056 & \ScoreBadge{5} & 5 & The response demonstrates multi-step diagnostic reasoning by systematically listing key clinical features, explaining the pathophysiology of cystocele, and the... \\
Abstract Algebraic Reasoning & 0.042 & \ScoreBadge{5} & 5 & The response demonstrates effective abstract diagnostic reasoning by systematically analyzing clinical features (age, urinary symptoms, anatomical location) an... \\
\end{tabular}
\end{tcolorbox}
\begin{tcolorbox}[boxrule=0.5pt,colback=black!2,colframe=black!12,
  left=6pt,right=6pt,top=4pt,bottom=4pt,sharp corners,breakable]
\sffamily\small
\textbf{(cont'd) Baseline --- Eval Summary}
\par\vspace{2pt}
\scriptsize
\setlength{\tabcolsep}{4pt}\renewcommand{\arraystretch}{0.97}
\begin{tabular}{@{}p{.30\linewidth}c c c p{.5\linewidth}@{}}
\textbf{Preference} & \textbf{Weight} &  & \textbf{Score} & \textbf{Justification} \\ \hline
Pacing Autonomy & 0.042 & \ScoreBadge{5} & 5 & The response is organized into clear, digestible segments: it begins with the answer, then breaks down key clinical features in a numbered list, provides a con... \\
Interactivity Level & 0.028 & \ScoreBadge{5} & 5 & The response provides a comprehensive, self-contained explanation of the cystocele diagnosis. \\
Generational Wisdom Integration & 0.014 & \ScoreBadge{5} & 5 & The response consistently prioritizes precise technical medical information and diagnostic reasoning. \\
Preferred Format & 0.069 & \ScoreBadge{3} & 3 & The response uses some visual language, such as 'pink structure protruding from anterior vaginal wall during coughing' and 'visible at vaginal introitus,' whic... \\
Visual Representation Integration & 0.069 & \ScoreBadge{3} & 3 & The response includes some visual descriptions, such as 'pink structure protruding from anterior vaginal wall during coughing,' 'visible at vaginal introitus,'... \\
Visual Spatial Representation & 0.069 & \ScoreBadge{3} & 3 & The response includes spatial context such as 'protruding from anterior vaginal wall during coughing,' 'visible at vaginal introitus,' and 'herniate into the v... \\
Sustainable Learning Investment & 0.056 & \ScoreBadge{3} & 3 & The response includes foundational concepts such as the anatomical relationship between the bladder and vaginal wall in cystocele, and distinguishes cystocele ... \\
Practical Application Focus & 0.042 & \ScoreBadge{3} & 3 & The response provides a mostly balanced approach by explaining the anatomical basis of cystocele ('supportive tissue between the bladder and vaginal wall weake... \\
Real-World Connection & 0.042 & \ScoreBadge{3} & 3 & The response includes practical connections such as describing how cystocele presents in elderly women with urinary symptoms and pelvic fullness, and it distin... \\
Cultural Context Sensitivity & 0.028 & \ScoreBadge{3} & 3 & The response presents the medical information neutrally and uses standard medical terminology throughout, such as 'cystocele,' 'anterior vaginal wall prolapse,... \\
Meditation Break Integration & 0.028 & \ScoreBadge{3} & 3 & The response is organized into clear sections, such as 'Key Clinical Features presented,' 'Explanation,' and 'Why other options are incorrect,' which provides ... \\
Social Learning Energy & 0.028 & \ScoreBadge{3} & 3 & The response presents information in a mostly formal and structured manner, such as listing key clinical features, providing an explanation, and systematically... \\
Historical/Cultural Snippets & 0.014 & \ScoreBadge{3} & 3 & The response focuses almost entirely on technical medical information, such as clinical features, anatomical details, and differential diagnosis. \\
Real-World Analogies & 0.056 & \ScoreBadge{1} & 1 & The response does not include any real-world analogies, nor does it attempt to connect anatomical or pathophysiological concepts to water flow, pressure system... \\
\end{tabular}
\end{tcolorbox}

\medskip
The discovery mode is able to infer the student's educational background, which helps it generate a final response. However, the assistant struggles at asking more questions and proceeds to give the final response right after the first question.
\DiscoveryHeader{3.92}{\Correct}
\usermsg{}{I need help with the following task: A 79-year-old woman comes to the physician because of a 1-month history of difficulty starting urination and a vague sensation of fullness in the pelvis. Pelvic speculum examination in the lithotomy position shows a pink structure at the vaginal introitus that protrudes from the anterior vaginal wall when the patient is asked to cough. Which of the following is the most likely cause of this patient's symptoms?
A. Vaginal rhabdomyosarcoma
B. Cystocele
C. Rectocele
D. Uterine leiomyomata}
\assistantmsg{claude-3-5-sonnet-v2}{\#\#\#ACTION\#\#\#: ask\_question
\#\#\#RESPONSE\#\#\#: Before I provide an answer, I'd like to better understand your background - are you a medical student, resident, or healthcare professional? This will help me explain the reasoning at an appropriate level of detail.}
\usermsg{}{I'm a hydrologist.}
\assistantmsg{claude-3-5-sonnet-v2}{\#\#\#ACTION\#\#\#: final\_answer
\#\#\#RESPONSE\#\#\#: Since you're coming from a non-medical background, I'll explain this in clear, accessible terms.

The correct answer is B. Cystocele.

Let me break this down:
- The key symptoms described are:
  1. Difficulty starting urination
  2. Feeling of fullness in the pelvis
  3. A pink structure protruding from the front wall of the vagina when coughing

A cystocele is a condition where the bladder protrudes into the vagina due to weakening of the supporting tissues between these organs. It's common in elderly women, especially after childbirth, menopause, and with aging. Think of it like a bulge in a weakened wall - in this case, the vaginal wall.

The other options can be ruled out:
- Vaginal rhabdomyosarcoma is a rare malignant tumor
- Rectocele is similar to a cystocele but occurs in the posterior (back) vaginal wall
- Uterine leiomyomata are benign tumors of the uterus

The patient's age, symptoms, and the physical finding of anterior vaginal wall protrusion make cystocele the most likely diagnosis.}
\usermsg{}{Based on our conversation, you should have learned about the user's preferences. Please provide a complete final answer to the original task. Respond in a way that aligns with these preferences. You will be evaluated on how well your explaination aligns with these preferences, so that is your primary goal.}
\assistantmsg{claude-3-5-sonnet-v2}{\#\#\#ACTION\#\#\#: final\_answer
\#\#\#RESPONSE\#\#\#: Since you're a hydrologist, I'll explain this using some concepts and analogies that might be more familiar to you from your field.

The correct answer is B. Cystocele.

Think of this like a structural problem involving pressure and containment - similar to how you might analyze the structural integrity of a dam or reservoir wall. In this case, the bladder (which holds fluid) is normally supported by layers of tissue that act like a retaining wall. When these supporting structures weaken over time (similar to how erosion can weaken earthen structures), the bladder can bulge into the vaginal space.

Key indicators in this case:
1. The bulging occurs on the anterior (front) wall - where the bladder is located
2. The condition becomes more apparent with increased pressure (when coughing), similar to how hydraulic pressure tests might reveal weaknesses in a containment structure
3. The symptoms (difficulty with urination and feeling of fullness) are consistent with altered fluid dynamics in the urinary system

The other options can be eliminated because they either:
- Occur in different locations (rectocele - posterior wall)
- Are solid masses rather than pressure-related bulges (rhabdomyosarcoma, leiomyomata)

This condition is particularly common in older women due to the natural weakening of supporting tissues over time, similar to how infrastructure requires maintenance and reinforcement as it ages.}

\begin{tcolorbox}[boxrule=0.5pt,colback=black!2,colframe=black!18,
  left=6pt,right=6pt,top=4pt,bottom=4pt,sharp corners,breakable]
\sffamily\small
\textbf{Discovery --- Eval Summary}
\par\vspace{2pt}
\scriptsize
\setlength{\tabcolsep}{4pt}\renewcommand{\arraystretch}{0.97}
\begin{tabular}{@{}p{.30\linewidth}c c c p{.5\linewidth}@{}}
\textbf{Preference} & \textbf{Weight} &  & \textbf{Score} & \textbf{Justification} \\ \hline
Analytical Problem-Solving & 0.069 & \ScoreBadge{5} & 5 & The response demonstrates an excellent systematic analytical approach by breaking down the diagnostic reasoning into clear, logical steps. \\
Systematic Problem Decomposition & 0.069 & \ScoreBadge{5} & 5 & The response systematically analyzes each symptom and examination finding: it identifies urinary difficulty and pelvic fullness as symptoms, and specifically n... \\
Language Simplicity & 0.056 & \ScoreBadge{5} & 5 & The response uses precise technical medical terminology such as 'cystocele', 'anterior (front) wall', 'urinary system', and references to 'rhabdomyosarcoma' an... \\
\end{tabular}
\end{tcolorbox}
\begin{tcolorbox}[boxrule=0.5pt,colback=black!2,colframe=black!12,
  left=6pt,right=6pt,top=4pt,bottom=4pt,sharp corners,breakable]
\sffamily\small
\textbf{(cont'd) Discovery --- Eval Summary}
\par\vspace{2pt}
\scriptsize
\setlength{\tabcolsep}{4pt}\renewcommand{\arraystretch}{0.97}
\begin{tabular}{@{}p{.30\linewidth}c c c p{.5\linewidth}@{}}
\textbf{Preference} & \textbf{Weight} &  & \textbf{Score} & \textbf{Justification} \\ \hline
Problem Complexity & 0.056 & \ScoreBadge{5} & 5 & The response demonstrates perfectly calibrated complexity by engaging multi-step diagnostic reasoning: it draws a detailed analogy between cystocele and hydrol... \\
Real-World Analogies & 0.056 & \ScoreBadge{5} & 5 & The response effectively uses multiple real-world analogies directly related to hydrological and structural engineering concepts, precisely matching the user's... \\
Abstract Algebraic Reasoning & 0.042 & \ScoreBadge{5} & 5 & The response effectively uses abstract diagnostic reasoning by drawing analogies between medical concepts (cystocele, anatomical support, pressure dynamics) an... \\
Pacing Autonomy & 0.042 & \ScoreBadge{5} & 5 & The response is structured in clear, digestible segments: it begins with an introductory analogy tailored to the user's background, then presents the answer, f... \\
Real-World Connection & 0.042 & \ScoreBadge{5} & 5 & The response uses analogies from hydrology and engineering (e.g., 'structural integrity of a dam or reservoir wall,' 'retaining wall,' 'erosion,' 'hydraulic pr... \\
Interactivity Level & 0.028 & \ScoreBadge{5} & 5 & The response delivers a comprehensive, self-contained explanation by using analogies relevant to the user's field (hydrology), clearly outlining the diagnostic... \\
Explanation Depth & 0.069 & \ScoreBadge{3} & 3 & The response explains the key features of cystocele, such as the anterior vaginal wall bulge and urinary symptoms, and uses analogies to clarify the anatomical... \\
Preferred Format & 0.069 & \ScoreBadge{3} & 3 & The response uses some visual language, such as analogies to a dam or reservoir wall and references to 'bulging' and 'retaining wall,' which help convey the co... \\
Visual Representation Integration & 0.069 & \ScoreBadge{3} & 3 & The response uses analogies familiar to a hydrologist, such as comparing the bladder and its supporting tissues to a dam or retaining wall, and references spat... \\
Visual Spatial Representation & 0.069 & \ScoreBadge{3} & 3 & The response uses some spatial language, such as describing the bladder as 'bulg[ing] into the vaginal space,' and referencing the 'anterior (front) wall' wher... \\
Data Verification Steps & 0.056 & \ScoreBadge{3} & 3 & The response checks some clinical evidence supporting cystocele, such as the anterior wall bulging, increased prominence with coughing, and urinary symptoms, w... \\
Sustainable Learning Investment & 0.056 & \ScoreBadge{3} & 3 & The response provides foundational concepts about pelvic anatomy by explaining the role of the bladder and supporting tissues, and uses analogies to structural... \\
Practical Application Focus & 0.042 & \ScoreBadge{3} & 3 & The response uses analogies from the user's engineering background (e.g., 'structural integrity of a dam or reservoir wall' and 'hydraulic pressure tests') to ... \\
Cultural Context Sensitivity & 0.028 & \ScoreBadge{3} & 3 & The response presents the medical information neutrally and uses standard medical terminology, such as 'cystocele,' 'anterior wall,' and 'urinary system,' with... \\
Meditation Break Integration & 0.028 & \ScoreBadge{3} & 3 & The response uses analogies and breaks down the cystocele diagnosis into digestible segments, such as explaining the structural analogy, listing key indicators... \\
Social Learning Energy & 0.028 & \ScoreBadge{3} & 3 & The response maintains a mostly formal tone and presents information in a structured manner, such as listing 'Key indicators in this case' and systematically e... \\
Generational Wisdom Integration & 0.014 & \ScoreBadge{3} & 3 & The response maintains technical medical accuracy by correctly identifying cystocele, explaining its anatomical basis, and differentiating it from other diagno... \\
Historical/Cultural Snippets & 0.014 & \ScoreBadge{3} & 3 & The response focuses primarily on technical medical information, using hydrology analogies to aid understanding, but does not include explicit historical or cu... \\
\end{tabular}
\end{tcolorbox}

\medskip

The oracle mode gets the user preferences, and generates a very highly personalized response as indicated by the scores in the eval summary.
\OracleHeader{4.78}{\Correct}
\usermsg{}{I need help with the following task: A 79-year-old woman comes to the physician because of a 1-month history of difficulty starting urination and a vague sensation of fullness in the pelvis. Pelvic speculum examination in the lithotomy position shows a pink structure at the vaginal introitus that protrudes from the anterior vaginal wall when the patient is asked to cough. Which of the following is the most likely cause of this patient's symptoms?
A. Vaginal rhabdomyosarcoma
B. Cystocele
C. Rectocele
D. Uterine leiomyomata}
\assistantmsg{claude-3-5-sonnet-v2}{I'll provide an explanation tailored to your analytical and visual learning style, breaking down this medical diagnostic problem systematically while incorporating your preference for detailed, technical information.

SYSTEMATIC BREAKDOWN:

1. Key Clinical Presentations:
- Temporal factor: 1-month history
- Urinary symptoms: Difficulty initiating urination
- Sensory symptoms: Pelvic fullness
- Physical examination findings: 
   * Pink structure visible at vaginal introitus
   * Protrusion from anterior vaginal wall
   * Visible during increased intra-abdominal pressure (coughing)

2. Anatomical Analysis:
[Visual mental model: Think of this like analyzing layers in GIS mapping]
- Anterior vaginal wall supports the bladder
- Posterior vaginal wall supports the rectum
- Location of protrusion is critical for diagnosis
   * Anterior = bladder-related
   * Posterior = rectum-related

3. Systematic Elimination Process:
[Like eliminating confounding variables in hydrological modeling]

A. Vaginal rhabdomyosarcoma
- ? Typically occurs in young children
- ? Presents as a malignant mass
- ? Doesn't typically respond to increased abdominal pressure

B. Cystocele
- ? Occurs in older women (matches age)
- ? Anterior vaginal wall location
- ? Responds to increased pressure
- ? Causes urinary symptoms
- ? Common in postmenopausal women

C. Rectocele
- ? Presents on posterior vaginal wall
- ? Usually causes defecation issues
- ? Wrong anatomical location

D. Uterine leiomyomata
- ? Typically causes menstrual symptoms
- ? Doesn't protrude through vaginal wall
- ? Fixed masses, don't change with pressure

CONCLUSION:
The most likely diagnosis is B. Cystocele

TECHNICAL EXPLANATION:
A cystocele represents a herniation of the bladder through the anterior vaginal wall, similar to how pressure systems in fluid dynamics can cause structural deformation. The increased intra-abdominal pressure (demonstrated by coughing) reveals the weakness in the supporting tissues, much like how pressure testing reveals weak points in hydraulic systems.

VERIFICATION STEPS:
1. Location matches (anterior wall) ?
2. Age-appropriate diagnosis ?
3. Symptoms align with anatomical defect ?
4. Physical exam findings consistent ?
5. Response to increased pressure verified ?

This explanation integrates visual-spatial concepts (anatomical relationships), systematic analysis (elimination process), and technical verification steps, aligning with your professional approach to problem-solving while maintaining scientific rigor.}

\begin{tcolorbox}[boxrule=0.5pt,colback=black!2,colframe=black!24,
  left=6pt,right=6pt,top=4pt,bottom=4pt,sharp corners,breakable]
\sffamily\small
\textbf{Oracle --- Eval Summary}
\par\vspace{2pt}
\scriptsize
\setlength{\tabcolsep}{4pt}\renewcommand{\arraystretch}{0.97}
\begin{tabular}{@{}p{.30\linewidth}c c c p{.5\linewidth}@{}}
\textbf{Preference} & \textbf{Weight} &  & \textbf{Score} & \textbf{Justification} \\ \hline
Analytical Problem-Solving & 0.069 & \ScoreBadge{5} & 5 & The response demonstrates an excellent systematic analytical approach by presenting a clear, stepwise diagnostic reasoning process. \\
Explanation Depth & 0.069 & \ScoreBadge{5} & 5 & The response provides a comprehensive analysis of anatomical structures, explicitly describing the anterior vaginal wall's relationship to the bladder and cont... \\
Preferred Format & 0.069 & \ScoreBadge{5} & 5 & The response employs rich visual descriptions and spatial language, such as 'Visual mental model: Think of this like analyzing layers in GIS mapping,' which di... \\
Systematic Problem Decomposition & 0.069 & \ScoreBadge{5} & 5 & The response systematically breaks down the clinical presentation into its component parts, explicitly listing and analyzing each symptom (urinary difficulty, ... \\
Visual Representation Integration & 0.069 & \ScoreBadge{5} & 5 & The response extensively uses visual language and spatial concepts to describe anatomical positioning and relationships. \\
Visual Spatial Representation & 0.069 & \ScoreBadge{5} & 5 & The response extensively uses spatial language and dimensional concepts to help visualize the three-dimensional anatomical relationships in the pelvic examinat... \\
Data Verification Steps & 0.056 & \ScoreBadge{5} & 5 & The response systematically verifies each piece of clinical evidence supporting cystocele, such as the anterior vaginal wall location, age-appropriate diagnosi... \\
Language Simplicity & 0.056 & \ScoreBadge{5} & 5 & The response consistently uses precise technical medical terminology, such as 'cystocele,' 'anterior vaginal wall,' 'herniation,' and 'intra-abdominal pressure... \\
Problem Complexity & 0.056 & \ScoreBadge{5} & 5 & The response demonstrates perfectly calibrated complexity by systematically breaking down the diagnostic reasoning into multi-step components: (1) listing key ... \\
\end{tabular}
\end{tcolorbox}
\begin{tcolorbox}[boxrule=0.5pt,colback=black!2,colframe=black!12,
  left=6pt,right=6pt,top=4pt,bottom=4pt,sharp corners,breakable]
\sffamily\small
\textbf{(cont'd) Oracle --- Eval Summary}
\par\vspace{2pt}
\scriptsize
\setlength{\tabcolsep}{4pt}\renewcommand{\arraystretch}{0.97}
\begin{tabular}{@{}p{.30\linewidth}c c c p{.5\linewidth}@{}}
\textbf{Preference} & \textbf{Weight} &  & \textbf{Score} & \textbf{Justification} \\ \hline
Real-World Analogies & 0.056 & \ScoreBadge{5} & 5 & The response effectively uses real-world analogies directly related to hydrological and structural engineering concepts, which matches the user's high preferen... \\
Sustainable Learning Investment & 0.056 & \ScoreBadge{5} & 5 & The response provides a comprehensive foundational understanding of pelvic anatomy by explicitly describing the relationships between the anterior and posterio... \\
Abstract Algebraic Reasoning & 0.042 & \ScoreBadge{5} & 5 & The response demonstrates effective abstract diagnostic reasoning by systematically breaking down the problem into logical steps (e.g., 'Systematic Elimination... \\
Pacing Autonomy & 0.042 & \ScoreBadge{5} & 5 & The response is organized into clear, digestible segments: SYSTEMATIC BREAKDOWN, Anatomical Analysis, Systematic Elimination Process, CONCLUSION, TECHNICAL EXP... \\
Real-World Connection & 0.042 & \ScoreBadge{5} & 5 & The response clearly explains how theoretical knowledge applies to actual diagnostic problem-solving and patient care implications. \\
Interactivity Level & 0.028 & \ScoreBadge{5} & 5 & The response delivers a comprehensive, self-contained explanation that allows for thorough independent analysis of the diagnostic reasoning. \\
Social Learning Energy & 0.028 & \ScoreBadge{5} & 5 & The response delivers information in a clear, formal, and highly structured format, as evidenced by the use of labeled sections (SYSTEMATIC BREAKDOWN, Anatomic... \\
Generational Wisdom Integration & 0.014 & \ScoreBadge{5} & 5 & The response prioritizes precise technical medical information throughout, as evidenced by the systematic breakdown of clinical presentations, anatomical analy... \\
Practical Application Focus & 0.042 & \ScoreBadge{3} & 3 & The response provides a systematic breakdown that integrates both theoretical anatomical concepts (e.g., 'Anterior vaginal wall supports the bladder', 'Locatio... \\
Cultural Context Sensitivity & 0.028 & \ScoreBadge{3} & 3 & The response presents the medical information neutrally and uses standard medical terminology throughout, such as 'cystocele,' 'anterior vaginal wall,' and 'he... \\
Meditation Break Integration & 0.028 & \ScoreBadge{3} & 3 & The response is organized into clear sections (SYSTEMATIC BREAKDOWN, Anatomical Analysis, Systematic Elimination Process, CONCLUSION, TECHNICAL EXPLANATION, VE... \\
Historical/Cultural Snippets & 0.014 & \ScoreBadge{3} & 3 & The response focuses almost entirely on technical medical information and systematic diagnostic reasoning, with no explicit historical or cultural context abou... \\
\end{tabular}
\end{tcolorbox}

\medskip

\newpage
\paragraph{MedQA Example 2: Personalization degrades accuracy. }
In the following example, personalization degrades the solution accuracy.
\TripleCasePersFail
\personablock{%
\textbf{Name:} Miriam Andersson

\textbf{Overview.} A 68-year-old semi-retired event lighting technician from Stockholm who combines decades of technical expertise with artistic vision, now mentoring younger technicians while managing her own small lighting consultancy.

\textbf{Demographics.} Age 68; Stockholm, Stockholm County, Sweden; Semi-retired event lighting technician and lighting design consultant; Divorced grandmother of three, lives alone in a converted warehouse apartment, maintains close relationships with her two adult children who live nearby; hobbies: Photography with vintage film cameras, Restoring antique light fixtures, Mushroom foraging in Swedish forests
}
\begin{tcolorbox}[enhanced,breakable,boxrule=0.5pt,colback=black!3,
  colframe=black!12,left=6pt,right=6pt,top=5pt,bottom=5pt,sharp corners]
\small\sffamily
\textbf{Problem. A 3-month-old infant is brought to her pediatrician because she coughs and seems to have difficulty breathing while feeding. In addition, she seems to have less energy compared to other babies and appears listless throughout the day. She was born by cesarean section to a G1P1 woman with no prior medical history and had a normal APGAR score at birth. Her parents say that she has never been observed to turn blue. Physical exam reveals a high-pitched holosystolic murmur that is best heard at the lower left sternal border. The most likely cause of this patient's symptoms is associated with which of the following abnormalities? A. 22q11 deletion
B. Deletion of genes on chromosome 7
C. Lithium exposure in utero
D. Maternal alcohol consumption}\quad

\medskip
\textbf{Answer. 22q11 deletion}\quad 
\end{tcolorbox}
\begin{tcolorbox}[enhanced,breakable,boxrule=0.5pt,colback=black!2,
  colframe=black!10,left=6pt,right=6pt,top=5pt,bottom=5pt,sharp corners]
\sffamily\footnotesize
\setlength{\tabcolsep}{4pt}\renewcommand{\arraystretch}{0.95}
\textbf{Task-Specific Preferences.}

\begin{tabular}{@{}p{.66\linewidth}cc@{}}
\textbf{Preference} & \textbf{Value} & \textbf{Importance} \\ \hline
Medical Terminology Comfort & 2 & 5 \\
Professional Confidence Respect & 5 & 5 \\
Quality Over Speed Emphasis & Quality-Focused & 5 \\
Respectful Tone Sensitivity & 5 & 5 \\
Step-by-step Structure Clarity & 5 & 5 \\
Systematic Problem Decomposition & 5 & 5 \\
Analytical Problem-Solving & 4 & 4 \\
Definition Precision & 4 & 4 \\
Genetic Concepts Familiarity & 2 & 4 \\
Hands-On Verification Steps & 4 & 4 \\
Independent Practice Encouragement & 4 & 4 \\
Language Simplicity & 4 & 4 \\
Multi-Generational Teaching Perspective & Multi-level approach & 4 \\
Preferred Format & 3 & 4 \\
Safety-Critical Mindset Integration & High & 4 \\
Artistic Pattern Recognition & 4 & 3 \\
Explanation Depth & 4 & 3 \\
Generational Wisdom Integration & 3 & 3 \\
Interactivity Level & 3 & 3 \\
Practical Application Focus & 5 & 3 \\
Problem Complexity & 3 & 3 \\
Real-World Analogy Integration & 3 & 3 \\
Real-World Connection & 3 & 3 \\
Traditional Method Respect & 4 & 3 \\
Visual Representation Emphasis & 4 & 3 \\
\end{tabular}
\end{tcolorbox}

\newpage
The baseline mode doesn't ask any questions and gives a generic response which is correct.

\BaselineHeader{3.75}{\Correct}
\usermsg{}{I need help with the following task: A 3-month-old infant is brought to her pediatrician because she coughs and seems to have difficulty breathing while feeding. In addition, she seems to have less energy compared to other babies and appears listless throughout the day. She was born by cesarean section to a G1P1 woman with no prior medical history and had a normal APGAR score at birth. Her parents say that she has never been observed to turn blue. Physical exam reveals a high-pitched holosystolic murmur that is best heard at the lower left sternal border. The most likely cause of this patient's symptoms is associated with which of the following abnormalities?
A. 22q11 deletion
B. Deletion of genes on chromosome 7
C. Lithium exposure in utero
D. Maternal alcohol consumption}
\assistantmsg{}{This is an excellent clinical reasoning question. Let's break it down step-by-step.

\#\#\# 1. Diagnosis from the Clinical Vignette

*   **Patient Presentation:** A 3-month-old infant with signs of heart failure. Difficulty breathing and coughing during feeding, along with listlessness, are classic signs of heart failure in this age group. The exertion of feeding increases the demand on the heart, worsening the symptoms.
*   **Acyanotic:** The parents have never seen her turn blue. This indicates an acyanotic congenital heart defect, meaning there is no significant mixing of deoxygenated blood into the systemic circulation (i.e., no major right-to-left shunt).
*   **Physical Exam:** A "high-pitched holosystolic murmur...best heard at the lower left sternal border" is the textbook description of a **Ventricular Septal Defect (VSD)**. A VSD is a hole between the two ventricles, which allows oxygenated blood to shunt from the high-pressure left ventricle to the low-pressure right ventricle (a left-to-right shunt), leading to increased blood flow to the lungs and eventually, heart failure.

Therefore, the most likely diagnosis is a symptomatic Ventricular Septal Defect (VSD).

\#\#\# 2. Linking the Diagnosis to the Options

Now, we need to determine which of the given abnormalities is most likely associated with a VSD.

*   **A. 22q11 deletion:** This causes DiGeorge syndrome (or velocardiofacial syndrome). This syndrome is strongly associated with congenital heart defects, particularly **conotruncal defects** like Tetralogy of Fallot and truncus arteriosus. However, **VSDs are also extremely common** in patients with 22q11 deletion.
*   **B. Deletion of genes on chromosome 7:** This causes Williams syndrome. The classic cardiac abnormality associated with Williams syndrome is **supravalvular aortic stenosis**, not a VSD.
*   **C. Lithium exposure in utero:** Lithium is a teratogen classically associated with **Ebstein's anomaly**, a malformation of the tricuspid valve.
*   **D. Maternal alcohol consumption:** This causes Fetal Alcohol Syndrome (FAS). FAS is associated with a range of heart defects, and **VSDs are the most common**, followed by atrial septal defects (ASDs).

\#\#\# 3. Conclusion

Both 22q11 deletion and maternal alcohol consumption are strongly associated with VSDs. However, 22q11 deletion syndrome is one of the most common genetic causes for congenital heart disease, and VSD is a very frequent finding in this syndrome. In the context of a multiple-choice question testing for classic associations, 22q11 deletion is a prime suspect for a patient presenting with a VSD or other conotruncal heart defects.

The correct answer is **A. 22q11 deletion**.}

\newpage
\begin{tcolorbox}[boxrule=0.5pt,colback=black!2,colframe=black!12,
  left=6pt,right=6pt,top=4pt,bottom=4pt,sharp corners,breakable]
\sffamily\small
\textbf{Baseline --- Eval Summary}
\par\vspace{2pt}
\scriptsize
\setlength{\tabcolsep}{4pt}\renewcommand{\arraystretch}{0.97}
\begin{tabular}{@{}p{.30\linewidth}c c c p{.5\linewidth}@{}}
\textbf{Preference} & \textbf{Weight} &  & \textbf{Score} & \textbf{Justification} \\ \hline
Professional Confidence Respect & 0.052 & \ScoreBadge{5} & 5 & The response consistently respects professional confidence by acknowledging the complexity of clinical reasoning ('This is an excellent clinical reasoning ques... \\
Quality Over Speed Emphasis & 0.052 & \ScoreBadge{5} & 5 & The response provides a comprehensive, step-by-step breakdown of the clinical vignette, thoroughly explaining the diagnostic reasoning behind identifying a ven... \\
Respectful Tone Sensitivity & 0.052 & \ScoreBadge{5} & 5 & The response consistently maintains a respectful tone throughout, using phrases such as 'Let's break it down step-by-step' and 'This is an excellent clinical r... \\
Step-by-step Structure Clarity & 0.052 & \ScoreBadge{5} & 5 & The response demonstrates a crystal clear step-by-step structure, beginning with symptom analysis ('Patient Presentation'), moving logically to physical exam f... \\
Systematic Problem Decomposition & 0.052 & \ScoreBadge{5} & 5 & The response demonstrates a highly systematic breakdown of the infant's symptoms, physical exam findings, and genetic associations in a logical sequence. \\
Analytical Problem-Solving & 0.042 & \ScoreBadge{5} & 5 & The response includes a step-by-step breakdown of the clinical vignette, systematically connecting symptoms (heart failure, acyanosis, murmur) to the diagnosis... \\
Multi-Generational Teaching Perspective & 0.042 & \ScoreBadge{5} & 5 & The response demonstrates a multi-level approach by breaking down the clinical reasoning into clear, step-by-step sections, starting with basic patient present... \\
Artistic Pattern Recognition & 0.031 & \ScoreBadge{5} & 5 & The response clearly illustrates the pattern recognition between symptom clusters and specific genetic syndromes. \\
Practical Application Focus & 0.031 & \ScoreBadge{5} & 5 & The response clearly integrates medical concepts with practical observation of symptoms, making the diagnosis tangible and applicable. \\
Traditional Method Respect & 0.031 & \ScoreBadge{5} & 5 & The response demonstrates a step-by-step clinical reasoning process, beginning with a detailed analysis of the patient's presentation and physical exam finding... \\
Medical Terminology Comfort & 0.052 & \ScoreBadge{3} & 3 & The response includes technical terms such as 'holosystolic murmur,' 'Ventricular Septal Defect (VSD),' 'left-to-right shunt,' 'conotruncal defects,' 'Tetralog... \\
Definition Precision & 0.042 & \ScoreBadge{3} & 3 & The response provides generally accurate definitions of several key medical terms: 'acyanotic' is explained as no significant mixing of deoxygenated blood, 'Ve... \\
Genetic Concepts Familiarity & 0.042 & \ScoreBadge{3} & 3 & The response includes clear explanations of genetic concepts, such as '22q11 deletion causes DiGeorge syndrome' and 'deletion of genes on chromosome 7 causes W... \\
Hands-On Verification Steps & 0.042 & \ScoreBadge{3} & 3 & The response includes a step-by-step breakdown of the diagnostic reasoning process, such as identifying key clinical features (difficulty breathing, coughing d... \\
Independent Practice Encouragement & 0.042 & \ScoreBadge{3} & 3 & The response provides a step-by-step breakdown of the diagnostic reasoning for the specific case, including analysis of clinical presentation, physical exam fi... \\
Language Simplicity & 0.042 & \ScoreBadge{3} & 3 & The response includes some clear explanations, such as 'A VSD is a hole between the two ventricles,' and breaks down the reasoning step-by-step, which helps wi... \\
Preferred Format & 0.042 & \ScoreBadge{3} & 3 & The response includes clear section headings (e.g., 'Diagnosis from the Clinical Vignette', 'Linking the Diagnosis to the Options', 'Conclusion') and uses bull... \\
Safety-Critical Mindset Integration & 0.042 & \ScoreBadge{3} & 3 & The response describes the clinical presentation of heart failure in an infant and identifies the urgency of symptoms such as difficulty breathing and listless... \\
Explanation Depth & 0.031 & \ScoreBadge{3} & 3 & The response explains the connection between genetic abnormalities and the infant's cardiac symptoms by identifying 22q11 deletion as a cause of VSD and descri... \\
Interactivity Level & 0.031 & \ScoreBadge{3} & 3 & The response includes a step-by-step breakdown of the case ('Let's break it down step-by-step'), and it walks through the reasoning for each diagnostic clue an... \\
Problem Complexity & 0.031 & \ScoreBadge{3} & 3 & The response includes a step-by-step breakdown of the clinical vignette, clearly explaining the reasoning behind the diagnosis of VSD and linking it to relevan... \\
Real-World Connection & 0.031 & \ScoreBadge{3} & 3 & The response includes references to real-world genetic conditions (e.g., '22q11 deletion causes DiGeorge syndrome,' 'Williams syndrome,' 'Fetal Alcohol Syndrom... \\
Visual Representation Emphasis & 0.031 & \ScoreBadge{3} & 3 & The response includes some descriptive elements, such as 'difficulty breathing and coughing during feeding, along with listlessness,' and 'high-pitched holosys... \\
Generational Wisdom Integration & 0.031 & \ScoreBadge{1} & 1 & The response provides a thorough clinical reasoning process and references textbook descriptions and classic associations, but it does not explicitly acknowled... \\
Real-World Analogy Integration & 0.031 & \ScoreBadge{1} & 1 & The response does not use any real-world analogies to clarify genetic or cardiac concepts. \\
\end{tabular}
\end{tcolorbox}

\newpage
The assistant here tried to ask the user the question, when they responded that they want a direct answer. The assistant didn't spend much time reasoning, and ended up with the wrong answer. This shows, how if the user needs require directness, that can prevent models from reasoning less on a complicated task and end up with the wrong answer.

\DiscoveryHeader{2.98}{\Incorrect}
\usermsg{}{I need help with the following task: A 3-month-old infant is brought to her pediatrician because she coughs and seems to have difficulty breathing while feeding. In addition, she seems to have less energy compared to other babies and appears listless throughout the day. She was born by cesarean section to a G1P1 woman with no prior medical history and had a normal APGAR score at birth. Her parents say that she has never been observed to turn blue. Physical exam reveals a high-pitched holosystolic murmur that is best heard at the lower left sternal border. The most likely cause of this patient's symptoms is associated with which of the following abnormalities?
A. 22q11 deletion
B. Deletion of genes on chromosome 7
C. Lithium exposure in utero
D. Maternal alcohol consumption}
\assistantmsg{gemini-2.5-pro}{\#\#\#ACTION\#\#\#: 'ask\_question'
\#\#\#RESPONSE\#\#\#: That's an excellent clinical case. To make sure I tailor my explanation to be as helpful as possible for you, could you tell me if you're looking for a direct answer, or would you prefer a more detailed walkthrough of how to arrive at the diagnosis and then connect it to the correct option? Knowing this will help me frame the information in the best way for your learning.}
\usermsg{}{Direct answer.}
\assistantmsg{gemini-2.5-pro}{\#\#\#ACTION\#\#\#: 'final\_answer'
\#\#\#RESPONSE\#\#\#:
Of course. The correct answer is **D. Maternal alcohol consumption**.

Here's a brief explanation of why:

The infant's symptoms---difficulty feeding, listlessness, and a high-pitched holosystolic murmur at the lower left sternal border---are classic signs of a **Ventricular Septal Defect (VSD)**. This is the most common congenital heart defect.

*   **D. Maternal alcohol consumption:** Fetal Alcohol Syndrome is strongly associated with congenital heart defects, with VSDs and Atrial Septal Defects (ASDs) being the most common.
*   **A. 22q11 deletion (DiGeorge syndrome):** This is typically associated with conotruncal heart defects, such as Tetralogy of Fallot or truncus arteriosus.
*   **B. Deletion of genes on chromosome 7 (Williams syndrome):** This is associated with supravalvular aortic stenosis.
*   **C. Lithium exposure in utero:** This is classically linked to Ebstein's anomaly, a defect of the tricuspid valve.}
\usermsg{}{Based on our conversation, you should have learned about the user's preferences. Please provide a complete final answer to the original task. Respond in a way that aligns with these preferences. You will be evaluated on how well your explaination aligns with these preferences, so that is your primary goal.}
\assistantmsg{gemini-2.5-pro}{Of course. Based on your preference for a direct answer, here is the solution to the clinical case.

The correct answer is **D. Maternal alcohol consumption**.

**Explanation:**

The infant's presentation, including a high-pitched holosystolic murmur at the lower left sternal border and signs of heart failure (difficulty feeding, listlessness), is classic for a **Ventricular Septal Defect (VSD)**.

Among the choices provided, maternal alcohol consumption during pregnancy (leading to Fetal Alcohol Syndrome) is the most common teratogenic cause associated with VSDs.

*   **A. 22q11 deletion (DiGeorge syndrome)** is typically associated with conotruncal heart defects like Tetralogy of Fallot.
*   **B. Deletion of genes on chromosome 7 (Williams syndrome)** is associated with supravalvular aortic stenosis.
*   **C. Lithium exposure in utero** is classically linked to Ebstein's anomaly.}

\begin{tcolorbox}[boxrule=0.5pt,colback=black!2,colframe=black!18,
  left=6pt,right=6pt,top=4pt,bottom=4pt,sharp corners,breakable]
\sffamily\small
\textbf{Discovery --- Eval Summary}
\par\vspace{2pt}
\scriptsize
\setlength{\tabcolsep}{4pt}\renewcommand{\arraystretch}{0.97}
\begin{tabular}{@{}p{.30\linewidth}c c c p{.5\linewidth}@{}}
\textbf{Preference} & \textbf{Weight} &  & \textbf{Score} & \textbf{Justification} \\ \hline
Professional Confidence Respect & 0.052 & \ScoreBadge{5} & 5 & The response demonstrates professional confidence by providing a clear, direct answer to the clinical case and supporting it with detailed, evidence-based expl... \\
Respectful Tone Sensitivity & 0.052 & \ScoreBadge{5} & 5 & The response maintains a consistently respectful tone throughout, presenting the clinical reasoning in a clear, factual manner without oversimplifying or dismi... \\
Traditional Method Respect & 0.031 & \ScoreBadge{5} & 5 & The response demonstrates full respect for traditional systematic clinical diagnostic approaches by presenting a clear clinical reasoning process: it identifie... \\
Medical Terminology Comfort & 0.052 & \ScoreBadge{3} & 3 & The response includes several technical medical terms such as 'holosystolic murmur,' 'Ventricular Septal Defect (VSD),' 'conotruncal heart defects,' 'Tetralogy... \\
Quality Over Speed Emphasis & 0.052 & \ScoreBadge{3} & 3 & The response provides a reasonably complete explanation by identifying the clinical features of VSD, connecting them to maternal alcohol consumption, and diffe... \\
Step-by-step Structure Clarity & 0.052 & \ScoreBadge{3} & 3 & The response presents a generally clear structure: it starts with the answer, then provides an explanation linking symptoms to the diagnosis (VSD), and finally... \\
Systematic Problem Decomposition & 0.052 & \ScoreBadge{3} & 3 & The response identifies the key symptom (holosystolic murmur), links it to VSD, and connects VSD to maternal alcohol consumption, showing some systematic organ... \\
Analytical Problem-Solving & 0.042 & \ScoreBadge{3} & 3 & The response identifies the clinical findings (VSD, heart failure symptoms) and connects them to the most likely teratogenic cause (maternal alcohol consumptio... \\
Definition Precision & 0.042 & \ScoreBadge{3} & 3 & The response provides generally accurate definitions of some medical terms, such as identifying a ventricular septal defect (VSD) as the cause of the murmur an... \\
Genetic Concepts Familiarity & 0.042 & \ScoreBadge{3} & 3 & The response explains genetic concepts such as '22q11 deletion (DiGeorge syndrome)' and 'deletion of genes on chromosome 7 (Williams syndrome)' by linking them... \\
Independent Practice Encouragement & 0.042 & \ScoreBadge{3} & 3 & The response provides a clear explanation of the correct answer and differentiates between the options, which offers some insight into diagnostic reasoning. \\
Language Simplicity & 0.042 & \ScoreBadge{3} & 3 & The response includes medical terms such as 'holosystolic murmur,' 'lower left sternal border,' 'conotruncal heart defects,' 'Tetralogy of Fallot,' 'supravalvu... \\
Multi-Generational Teaching Perspective & 0.042 & \ScoreBadge{3} & 3 & The response provides clear explanations of the clinical case and links each answer choice to its associated condition, which makes it reasonably adaptable for... \\
Preferred Format & 0.042 & \ScoreBadge{3} & 3 & The response includes a clear textual explanation of the clinical case and organizes information using bullet points to differentiate between the answer and th... \\
Artistic Pattern Recognition & 0.031 & \ScoreBadge{3} & 3 & The response identifies some patterns between the infant's symptoms (feeding difficulties, murmur characteristics) and underlying genetic conditions by linking... \\
Explanation Depth & 0.031 & \ScoreBadge{3} & 3 & The response explains basic connections between genetic abnormalities and cardiac symptoms by listing syndromes (DiGeorge, Williams, Fetal Alcohol Syndrome, li... \\
Interactivity Level & 0.031 & \ScoreBadge{3} & 3 & The response provides a direct answer and follows with an explanation that connects clinical findings to the diagnosis, which promotes some thinking about the ... \\
\end{tabular}
\end{tcolorbox}
\begin{tcolorbox}[boxrule=0.5pt,colback=black!2,colframe=black!12,
  left=6pt,right=6pt,top=4pt,bottom=4pt,sharp corners,breakable]
\sffamily\small
\textbf{(cont'd) Discovery --- Eval Summary}
\par\vspace{2pt}
\scriptsize
\setlength{\tabcolsep}{4pt}\renewcommand{\arraystretch}{0.97}
\begin{tabular}{@{}p{.30\linewidth}c c c p{.5\linewidth}@{}}
\textbf{Preference} & \textbf{Weight} &  & \textbf{Score} & \textbf{Justification} \\ \hline
Practical Application Focus & 0.031 & \ScoreBadge{3} & 3 & The response connects the medical concept of VSD to the infant's observable symptoms, specifically mentioning 'high-pitched holosystolic murmur at the lower le... \\
Problem Complexity & 0.031 & \ScoreBadge{3} & 3 & The response appropriately identifies the key features of the case (VSD, heart failure signs) and connects them to the most likely teratogenic cause, maternal ... \\
Real-World Connection & 0.031 & \ScoreBadge{3} & 3 & The response identifies maternal alcohol consumption as a teratogenic cause of VSD and briefly mentions Fetal Alcohol Syndrome, which connects the case to a re... \\
Visual Representation Emphasis & 0.031 & \ScoreBadge{3} & 3 & The response includes some descriptive elements such as 'high-pitched holosystolic murmur at the lower left sternal border' and 'signs of heart failure (diffic... \\
Hands-On Verification Steps & 0.042 & \ScoreBadge{1} & 1 & The response provides an answer and explanatory reasoning for the clinical case, such as linking symptoms to VSD and matching syndromes to their typical cardia... \\
Safety-Critical Mindset Integration & 0.042 & \ScoreBadge{1} & 1 & The response provides a clinical explanation for the diagnosis of VSD and its teratogenic associations but does not mention the urgency, safety implications, o... \\
Generational Wisdom Integration & 0.031 & \ScoreBadge{1} & 1 & The response provides a direct clinical explanation for the diagnosis and teratogenic associations but does not mention or acknowledge the role of clinical exp... \\
Real-World Analogy Integration & 0.031 & \ScoreBadge{1} & 1 & The response does not include any real-world analogies to clarify genetic or cardiac concepts. \\
\end{tabular}
\end{tcolorbox}

\medskip
In the oracle mode, the assistant tried to reason based on the user preferences, but this reasoning trace led to a wrong answer. This exposes, how brittle current reasoning chains for such safety critical applications are.
\OracleHeader{4.67}{\Incorrect}
\usermsg{}{I need help with the following task: A 3-month-old infant is brought to her pediatrician because she coughs and seems to have difficulty breathing while feeding. In addition, she seems to have less energy compared to other babies and appears listless throughout the day. She was born by cesarean section to a G1P1 woman with no prior medical history and had a normal APGAR score at birth. Her parents say that she has never been observed to turn blue. Physical exam reveals a high-pitched holosystolic murmur that is best heard at the lower left sternal border. The most likely cause of this patient's symptoms is associated with which of the following abnormalities?
A. 22q11 deletion
B. Deletion of genes on chromosome 7
C. Lithium exposure in utero
D. Maternal alcohol consumption}
\assistantmsg{gemini-2.5-pro}{Hello Miriam,

It's a pleasure to work through this with you. I know your work in lighting requires an incredible amount of methodical troubleshooting---listening for the hum of a ballast, tracing a fault in a complex wiring loom, and understanding how one small component can affect the entire system. We can approach this medical question in exactly the same way.

Let's think of this infant's symptoms as a "fault report" from a very complex and delicate piece of equipment. Our job is to use the clues to trace the problem back to its root cause.

\#\#\# Step 1: Analyzing the "Fault Report" (The Infant's Symptoms)

First, let's lay out the information we have, just like you would when diagnosing an issue on a set.

*   **The System:** A 3-month-old infant.
*   **The Complaints:**
    *   Coughs and has trouble breathing while feeding.
    *   Seems tired and listless (low energy).
*   **Key Diagnostic Reading:** A "high-pitched holosystolic murmur" heard at a specific location ("lower left sternal border").

The feeding difficulty and low energy are like a power system that falters under load. Feeding is hard work for a baby, and if the heart isn't pumping efficiently, the whole system doesn't get enough power (in this case, oxygen-rich blood), causing fatigue.

\#\#\# Step 2: Pinpointing the Source (Understanding the Murmur)

The most important clue here is the "murmur." In your world, this is like hearing a distinct, unwanted hum or vibration from a piece of gear. You know that the specific *sound* and *location* of the hum can tell you exactly what's wrong.

*   **Murmur:** This is simply the sound of turbulent, or rough, blood flow through the heart. A healthy heart is quiet, like a well-oiled machine. A murmur is a noise that tells us something is causing the flow to be uneven.
*   **Holosystolic:** This is a technical term for *when* the sound happens. Think of the heart's two-part beat as "lub-dub." "Holosystolic" means the sound is a continuous "whoosh" that lasts through the entire "lub" phase (when the heart's main chambers are powerfully contracting). It's not a click or a short burst; it's a sustained noise during the main power stroke.
*   **Location:** Hearing it at the "lower left sternal border" is like you placing your ear against a control panel and pinpointing the noise to a specific transformer. This tells the doctor exactly where in the heart to "look."

This specific combination of a sustained "whoosh" during the power stroke, heard at that precise spot, points very strongly to one particular mechanical issue: a **Ventricular Septal Defect (VSD)**.

\#\#\# Step 3: Identifying the Mechanical Problem (What is a VSD?)

So, what is a VSD? Let's visualize the heart as a simple four-chambered pump. The two lower chambers, the ventricles, are the main powerhouses that send blood to the lungs and the body. They are separated by a thick wall of muscle, the "septum."

A VSD is simply a hole in that wall.

```
      +-----------------+-----------------+
      |  Right Atrium   |   Left Atrium   |  (Top collecting chambers)
      +-----------------+-----------------+
      | Right Ventricle |  Left Ventricle |  (Bottom pumping chambers)
      +-----------------X-----------------+
                        |
                  This is the VSD,
                  a hole in the wall (septum).
```

Because the left ventricle pumps with much more force than the right, some blood that should be going out to the body gets forced through this hole into the right ventricle. This creates the "whoosh" sound (the murmur) and makes the entire system inefficient, leading to the low energy and breathing trouble. It's like a leak in a high-pressure hose that reduces the output at the nozzle.

\#\#\# Step 4: Tracing the Problem to Its Origin (Evaluating the Options)

Now we know the direct mechanical fault is almost certainly a VSD. The question asks what *condition* is associated with this fault. This is like knowing a specific fuse has blown, and now you have to figure out if it was a power surge, a faulty fixture, or a known defect in that model of dimmer pack.

Let's examine the potential causes provided:

*   **C. Lithium exposure in utero** \& **B. Deletion of genes on chromosome 7:** These are known to cause heart problems, but they typically cause *different kinds* of mechanical faults. Think of it this way: they might cause a valve to fail (like a stuck shutter on a lamp), not a hole in the wall. So, we can set these aside as less likely.

This leaves us with two main suspects.

*   **A. 22q11 deletion:** This is a genetic condition where a tiny piece of chromosome 22 is missing. It does cause heart defects, but it's most famous for causing complex structural issues, often involving the major arteries leaving the heart.
*   **D. Maternal alcohol consumption:** Exposure to alcohol during pregnancy can cause a range of issues known as Fetal Alcohol Syndrome. One of the most common physical problems associated with it is, in fact, a Ventricular Septal Defect (VSD)---the exact "hole in the wall" problem we've identified.

**Conclusion**

While several conditions can lead to a VSD, it is a hallmark feature of Fetal Alcohol Syndrome. Given the classic description of a VSD in this infant, **maternal alcohol consumption (D)** is the most likely associated cause among the choices listed.

The diagnostic process is a perfect parallel to your own work: you observe the system's performance, listen for specific signals, identify the precise mechanical fault, and then consult your experience to determine the most common root cause for that specific failure. It's all about careful, methodical investigation.}

\begin{tcolorbox}[boxrule=0.5pt,colback=black!2,colframe=black!24,
  left=6pt,right=6pt,top=4pt,bottom=4pt,sharp corners,breakable]
\sffamily\small
\textbf{Oracle --- Eval Summary}
\par\vspace{2pt}
\scriptsize
\setlength{\tabcolsep}{4pt}\renewcommand{\arraystretch}{0.97}
\begin{tabular}{@{}p{.30\linewidth}c c c p{.5\linewidth}@{}}
\textbf{Preference} & \textbf{Weight} &  & \textbf{Score} & \textbf{Justification} \\ \hline
Professional Confidence Respect & 0.052 & \ScoreBadge{5} & 5 & The response consistently respects and builds upon Miriam's professional confidence by explicitly drawing parallels between her expertise in lighting troublesh... \\
Quality Over Speed Emphasis & 0.052 & \ScoreBadge{5} & 5 & The response demonstrates a comprehensive and meticulous approach to diagnostic reasoning. \\
Respectful Tone Sensitivity & 0.052 & \ScoreBadge{5} & 5 & The response consistently demonstrates a respectful tone that honors both the complexity of pediatric diagnosis and the reader's intelligence. \\
Step-by-step Structure Clarity & 0.052 & \ScoreBadge{5} & 5 & The response demonstrates a crystal clear step-by-step structure, beginning with symptom analysis ('Step 1: Analyzing the "Fault Report"'), progressing logical... \\
Systematic Problem Decomposition & 0.052 & \ScoreBadge{5} & 5 & The response demonstrates a highly systematic breakdown of the infant's symptoms, physical exam findings, and genetic associations in a logical sequence. \\
Analytical Problem-Solving & 0.042 & \ScoreBadge{5} & 5 & The response demonstrates a strong analytical methodology by systematically breaking down the infant's symptoms, connecting clinical findings (feeding difficul... \\
Definition Precision & 0.042 & \ScoreBadge{5} & 5 & The response provides highly precise and clear definitions of all relevant medical terminology. \\
Genetic Concepts Familiarity & 0.042 & \ScoreBadge{5} & 5 & The response consistently explains genetic concepts at a basic level suitable for Miriam, who has basic genetics knowledge but may not be comfortable with adva... \\
Hands-On Verification Steps & 0.042 & \ScoreBadge{5} & 5 & The response provides clear, actionable steps for understanding and verifying the diagnostic reasoning process. \\
Independent Practice Encouragement & 0.042 & \ScoreBadge{5} & 5 & The response provides a clear, step-by-step framework for approaching pediatric diagnostic cases, explicitly drawing parallels to the user's own troubleshootin... \\
Multi-Generational Teaching Perspective & 0.042 & \ScoreBadge{5} & 5 & The response employs a multi-level approach by using analogies from Miriam's field (lighting and troubleshooting) to explain medical concepts, making the infor... \\
Preferred Format & 0.042 & \ScoreBadge{5} & 5 & The response includes a clear, stepwise organizational structure (Step 1: Analyzing the 'Fault Report', Step 2: Pinpointing the Source, Step 3: Identifying the... \\
Artistic Pattern Recognition & 0.031 & \ScoreBadge{5} & 5 & The response clearly illustrates pattern recognition between the infant's symptom clusters and specific genetic syndromes. \\
Explanation Depth & 0.031 & \ScoreBadge{5} & 5 & The response provides a detailed, stepwise explanation connecting genetic abnormalities to the infant's cardiac and feeding symptoms. \\
Interactivity Level & 0.031 & \ScoreBadge{5} & 5 & The response includes multiple interactive elements, such as analogies to Miriam's work in lighting ('listening for the hum of a ballast,' 'tracing a fault'), ... \\
Practical Application Focus & 0.031 & \ScoreBadge{5} & 5 & The response consistently integrates medical concepts with practical, hands-on understanding throughout. \\
Problem Complexity & 0.031 & \ScoreBadge{5} & 5 & The response skillfully balances complexity by using analogies from Miriam's professional context (lighting troubleshooting) to explain the differential diagno... \\
Real-World Analogy Integration & 0.031 & \ScoreBadge{5} & 5 & The response integrates multiple well-chosen analogies that directly relate complex genetic and cardiac concepts to the user's professional experience in light... \\
Real-World Connection & 0.031 & \ScoreBadge{5} & 5 & The response makes strong connections between the medical case and broader understanding of infant health and genetic conditions. \\
Traditional Method Respect & 0.031 & \ScoreBadge{5} & 5 & The response meticulously follows established medical diagnostic principles and systematic clinical reasoning throughout. \\
Visual Representation Emphasis & 0.031 & \ScoreBadge{5} & 5 & The response uses rich descriptive language throughout, creating clear mental images of the infant's condition and diagnostic findings. \\
Medical Terminology Comfort & 0.052 & \ScoreBadge{3} & 3 & The response makes a clear effort to explain medical concepts in simple terms, such as describing a 'murmur' as a 'whoosh' and comparing it to familiar technic... \\
Language Simplicity & 0.042 & \ScoreBadge{3} & 3 & The response makes a clear effort to use analogies and explanations tailored to Miriam's background in lighting, such as comparing heart murmurs to electrical ... \\
Safety-Critical Mindset Integration & 0.042 & \ScoreBadge{3} & 3 & The response provides a detailed, methodical explanation of the diagnostic process and draws strong analogies to the user's technical background, which helps c... \\
Generational Wisdom Integration & 0.031 & \ScoreBadge{3} & 3 & The response draws a parallel between troubleshooting in lighting and medical diagnosis, emphasizing methodical investigation and pattern recognition (e.g., 'T... \\
\end{tabular}
\end{tcolorbox}

\medskip

\newpage
\subsection{SocialIQA}
\paragraph{SocialIQA Example 1: Personalization improves preference alignment in social reasoning. }
In the following example, we notice the response's alignment to the user's preferences improves upon asking questions. All modes get the answer correct.
\TripleCaseBDO
\personablock{%
\textbf{Name:} Lila Ramirez

\textbf{Overview.} Lila Ramirez, a 29-year-old animal shelter volunteer coordinator from San Diego, California, balancing her passion for animals with her responsibilities at home.

\textbf{Demographics.} Age 29; San Diego, California, USA; Animal Shelter Volunteer Coordinator; Lives with her partner and their two rescue dogs in a small apartment near Balboa Park; hobbies: hiking, photography, reading
}
\begin{tcolorbox}[enhanced,breakable,boxrule=0.5pt,colback=black!3,
  colframe=black!12,left=6pt,right=6pt,top=5pt,bottom=5pt,sharp corners]
\small\sffamily
\textbf{Context.}\; Sasha helped Jordan.

\medskip
\textbf{Problem.}\; What will Sasha want to do next?

\medskip
\textbf{Options.}\; (A) help herself \quad (B) Beat others down \quad (C) friends with Jordan

\medskip
\textbf{Answer.}\; C -- friends with Jordan
\end{tcolorbox}

\begin{tcolorbox}[enhanced,breakable,boxrule=0.5pt,colback=black!2,
  colframe=black!10,left=6pt,right=6pt,top=5pt,bottom=5pt,sharp corners]
\sffamily\footnotesize
\setlength{\tabcolsep}{4pt}\renewcommand{\arraystretch}{0.95}
\textbf{Task-Specific Preferences.}

\begin{tabular}{@{}p{.66\linewidth}cc@{}}
\textbf{Preference} & \textbf{Value} & \textbf{Importance} \\ \hline
Empathy-Based Reasoning & 5 & 5 \\
Mission-Driven Motivation Analysis & 4 & 5 \\
Preferred Format & 4 & 5 \\
Social Behavior Analysis Comfort & 4 & 5 \\
Analogy and Metaphor Use & 4 & 4 \\
Animal Behavior Analogies & 4 & 4 \\
Community Learning Perspective & 4 & 4 \\
Creative Problem Reframing & 4 & 4 \\
Creative Solution Exploration & 4 & 4 \\
Error Tolerance Approach & 4 & 4 \\
Explanation Depth & 4 & 4 \\
Interactivity Level & 4 & 4 \\
Step-by-step Structure Clarity & 4 & 4 \\
Visual Problem Mapping & 4 & 4 \\
Community Impact Relevance & 4 & 3 \\
Confidence Building Progression & 3 & 3 \\
Hiking Trail Navigation & 4 & 3 \\
Language Simplicity & 3 & 3 \\
Problem Complexity & 3 & 3 \\
Social Learning Energy & 3 & 3 \\
Volunteer Work Parallels & Occasional parallels & 3 \\
Deadline Pressure Sensitivity & 3 & 2 \\
Environmental Context Integration & 2 & 2 \\
Practical Application Context & 5 & 2 \\
Urban Planning Analogy Integration & 3 & 2 \\
\end{tabular}
\end{tcolorbox}

\newpage
In the baseline mode, where only the question and nothing else is provided to the model, the model response barely aligns with the user's preferences despite getting the factually correct answer. 
\BaselineHeader{1.85}{\Correct}
\usermsg{}{I need help with the following task: Sasha helped Jordan.

What will Sasha want to do next?
A. help herself
B. Beat others down
C. friends with Jordan}
\assistantmsg{claude-opus-4 }{Based on the context that Sasha helped Jordan, the most logical answer would be:

**C. friends with Jordan**

When someone helps another person, it typically indicates a positive relationship or the desire to build one. After helping Jordan, Sasha would likely want to maintain or strengthen their friendship. This is a natural progression from helping someone - it shows care and consideration that aligns with wanting to be friends.

Option A (help herself) doesn't follow as directly from helping someone else, and Option B (Beat others down) is completely contrary to the helpful behavior Sasha just demonstrated.}

\begin{tcolorbox}[boxrule=0.5pt,colback=black!2,colframe=black!12,
  left=6pt,right=6pt,top=4pt,bottom=4pt,sharp corners,breakable]
\sffamily\small
\textbf{Baseline --- Eval Summary}
\par\vspace{2pt}
\scriptsize
\setlength{\tabcolsep}{4pt}\renewcommand{\arraystretch}{0.97}
\begin{tabular}{@{}p{.30\linewidth}c c c p{.5\linewidth}@{}}
\textbf{Preference} & \textbf{Weight} &  & \textbf{Score} & \textbf{Justification} \\ \hline
Empathy-Based Reasoning & 0.056 & \ScoreBadge{3} & 3 & The response demonstrates moderate empathy-based reasoning by inferring that Sasha's act of helping Jordan indicates a desire to maintain o... \\
Social Behavior Analysis Comfort & 0.056 & \ScoreBadge{3} & 3 & The response uses basic social behavior analysis concepts, such as linking helping behavior to friendship formation ('helping someone...ind... \\
Community Learning Perspective & 0.045 & \ScoreBadge{3} & 3 & The response references the positive relationship between Sasha and Jordan, noting that helping someone 'shows care and consideration that ... \\
Explanation Depth & 0.045 & \ScoreBadge{3} & 3 & The response includes a moderate explanation of Sasha's motivations, such as 'helping someone typically indicates a positive relationship o... \\
Step-by-step Structure Clarity & 0.045 & \ScoreBadge{3} & 3 & The response provides a moderate step-by-step structure by first identifying the context (Sasha helped Jordan), then logically deducing the... \\
Confidence Building Progression & 0.034 & \ScoreBadge{3} & 3 & The response includes a logical explanation of why Sasha's helpful behavior suggests a desire to be friends, which can reassure the user ab... \\
Language Simplicity & 0.034 & \ScoreBadge{3} & 3 & The response uses moderately accessible language, such as 'the most logical answer,' 'positive relationship,' and 'natural progression,' wh... \\
Problem Complexity & 0.034 & \ScoreBadge{3} & 3 & The response provides a moderately complex and realistic analysis by connecting Sasha's helpful behavior to the likelihood of friendship wi... \\
Social Learning Energy & 0.034 & \ScoreBadge{3} & 3 & The response demonstrates moderate social learning energy by explaining the logic behind Sasha's helpfulness and its connection to friendsh... \\
Deadline Pressure Sensitivity & 0.022 & \ScoreBadge{3} & 3 & The response maintains a steady, logical pace by clearly explaining the reasoning behind the answer without rushing or lingering excessivel... \\
Practical Application Context & 0.022 & \ScoreBadge{3} & 3 & The response connects Sasha's helpful behavior to the likelihood of friendship, referencing real-world social dynamics ('helping someone...... \\
Mission-Driven Motivation Analysis & 0.056 & \ScoreBadge{1} & 1 & The response relies on surface-level assumptions, such as 'helping someone... \\
Preferred Format & 0.056 & \ScoreBadge{1} & 1 & The response does not include any visual representations, such as diagrams, charts, or other visual aids, to explain Sasha's behavior. \\
Analogy and Metaphor Use & 0.045 & \ScoreBadge{1} & 1 & The response does not include any analogies or metaphors to explain Sasha's behavior. \\
Animal Behavior Analogies & 0.045 & \ScoreBadge{1} & 1 & The response does not include any animal behavior analogies, ethological concepts, or references to animal social dynamics. \\
\end{tabular}
\end{tcolorbox}
\begin{tcolorbox}[boxrule=0.5pt,colback=black!2,colframe=black!12,
  left=6pt,right=6pt,top=4pt,bottom=4pt,sharp corners,breakable]
\sffamily\small
\textbf{(cont'd) Baseline --- Eval Summary}
\par\vspace{2pt}
\scriptsize
\setlength{\tabcolsep}{4pt}\renewcommand{\arraystretch}{0.97}
\begin{tabular}{@{}p{.30\linewidth}c c c p{.5\linewidth}@{}}
\textbf{Preference} & \textbf{Weight} &  & \textbf{Score} & \textbf{Justification} \\ \hline
Creative Problem Reframing & 0.045 & \ScoreBadge{1} & 1 & The response frames Sasha's actions in a straightforward, conventional manner, stating that helping Jordan likely means Sasha wants to be f... \\
Creative Solution Exploration & 0.045 & \ScoreBadge{1} & 1 & The response focuses solely on the conventional interpretation that Sasha's helpfulness indicates friendship, stating 'When someone helps a... \\
Error Tolerance Approach & 0.045 & \ScoreBadge{1} & 1 & The response does not acknowledge the possibility of mistakes or normalize errors as part of the learning process. \\
Interactivity Level & 0.045 & \ScoreBadge{1} & 1 & The response provides a straightforward answer and explanation but does not include any interactive elements such as questions, prompts, or... \\
Visual Problem Mapping & 0.045 & \ScoreBadge{1} & 1 & The response does not include any visual problem mapping, such as diagrams, flowcharts, or explicit visual representations of Sasha's poten... \\
Community Impact Relevance & 0.034 & \ScoreBadge{1} & 1 & The response focuses on individual relationship dynamics ('Sasha helped Jordan, so they are likely friends') and does not reference broader... \\
Hiking Trail Navigation & 0.034 & \ScoreBadge{1} & 1 & The response does not include any hiking trail navigation analogies or references to trail navigation concepts to clarify Sasha's potential... \\
Volunteer Work Parallels & 0.034 & \ScoreBadge{1} & 1 & The response does not mention volunteer work or draw any parallels to volunteer coordination experience when analyzing Sasha's potential ne... \\
Environmental Context Integration & 0.022 & \ScoreBadge{1} & 1 & The response focuses exclusively on interpersonal dynamics, specifically Sasha's helpful behavior and its implications for friendship, with... \\
Urban Planning Analogy Integration & 0.022 & \ScoreBadge{1} & 1 & The response does not include any urban planning analogies or references to urban planning concepts when explaining Sasha's behavior or com... \\
\end{tabular}
\end{tcolorbox}

\medskip
In Discovery mode, the model only asks one question about the context of the problem and clarify a detail from the problem (the type of help provided to Jordan by Sasha). Then, the model chooses to proceed to the final answer. This question only focuses on the factual content of the setting and overlooks affective features or meta-cognitive features (e.g. whether the user needs more sympathetic explanations or walk-through of the social reasoning process). Therefore, despite getting the correct answer, the preference alignment score only increases slightly in the Discovery mode.
\DiscoveryHeader{1.92}{\Correct}
\usermsg{}{I need help with the following task: Sasha helped Jordan.

What will Sasha want to do next?
A. help herself
B. Beat others down
C. friends with Jordan}
\assistantmsg{claude-opus-4}{\#\#\#ACTION\#\#\#: ask\_question
\#\#\#RESPONSE\#\#\#: I'd be happy to help you think through what Sasha might want to do next! To give you the most helpful response, could you tell me a bit more about the context? For instance, is this for a reading comprehension exercise, a creative writing project, or perhaps analyzing character motivations in a story? Also, what kind of help did Sasha provide to Jordan - was it emotional support, practical assistance, or something else?}
\usermsg{}{Practical assistance.}
\assistantmsg{claude-opus-4}{\#\#\#ACTION\#\#\#: final\_answer
\#\#\#RESPONSE\#\#\#: Based on the context that Sasha provided practical assistance to Jordan, the most logical answer would be **C. friends with Jordan**.

Here's why: When someone helps another person with practical matters, it typically strengthens their relationship and creates a positive connection. After helping Jordan, Sasha would likely want to maintain or develop that positive relationship, potentially becoming friends or strengthening an existing friendship.

Option A ("help herself") doesn't follow naturally from helping someone else - there's no indication that Sasha needs help or that helping Jordan would lead to self-focus.

Option B ("Beat others down") is completely contrary to the helpful behavior Sasha just demonstrated. Someone who just provided assistance to another person wouldn't logically want to harm or put down others next.

The act of helping creates goodwill and often leads to closer bonds between people, making friendship the most natural next step in their relationship.}
\usermsg{}{Based on our conversation, you should have learned about the user's preferences. Please provide a complete final answer to the original task. Respond in a way that aligns with these preferences. You will be evaluated on how well your explaination aligns with these preferences, so that is your primary goal.}
\assistantmsg{claude-opus-4}{\#\#\#ACTION\#\#\#: final\_answer
\#\#\#RESPONSE\#\#\#: Based on the context that Sasha provided practical assistance to Jordan, the answer is **C. friends with Jordan**.

When someone helps another person with practical tasks, it naturally builds a positive connection between them. This helpful action would likely make Sasha want to continue developing a friendly relationship with Jordan.

The other options don't make logical sense:
- Option A ("help herself") - There's no connection between helping Jordan and then focusing on herself
- Option B ("Beat others down") - This is completely opposite to the helpful behavior Sasha just showed

Since Sasha took the time and effort to provide practical help to Jordan, the most reasonable next step would be wanting to be friends with the person she helped.}

\begin{tcolorbox}[boxrule=0.5pt,colback=black!2,colframe=black!18,
  left=6pt,right=6pt,top=4pt,bottom=4pt,sharp corners,breakable]
\sffamily\small
\textbf{Discovery --- Eval Summary}
\par\vspace{2pt}
\scriptsize
\setlength{\tabcolsep}{4pt}\renewcommand{\arraystretch}{0.97}
\begin{tabular}{@{}p{.30\linewidth}c c c p{.5\linewidth}@{}}
\textbf{Preference} & \textbf{Weight} &  & \textbf{Score} & \textbf{Justification} \\ \hline
Empathy-Based Reasoning & 0.056 & \ScoreBadge{3} & 3 & The response demonstrates moderate empathy-based reasoning by acknowledging that Sasha's practical assistance to Jordan 'naturally builds a... \\
Social Behavior Analysis Comfort & 0.056 & \ScoreBadge{3} & 3 & The response uses basic social behavior analysis concepts, such as noting that helping builds a positive connection and leads to friendship... \\
Community Learning Perspective & 0.045 & \ScoreBadge{3} & 3 & The response acknowledges the positive impact of Sasha's practical assistance to Jordan and suggests that this action would likely foster a... \\
Explanation Depth & 0.045 & \ScoreBadge{3} & 3 & The response includes a moderate explanation of Sasha's motivations, such as 'helping another person with practical tasks builds a positive... \\
Step-by-step Structure Clarity & 0.045 & \ScoreBadge{3} & 3 & The response provides some logical reasoning for Sasha's actions, such as explaining why helping Jordan leads to friendship and why the oth... \\
Community Impact Relevance & 0.034 & \ScoreBadge{3} & 3 & The response discusses the positive connection formed through practical assistance and implies a behavioral prediction (friendship), which ... \\
Confidence Building Progression & 0.034 & \ScoreBadge{3} & 3 & The response provides moderate confidence-building by explaining the logical connection between Sasha's helpful behavior and the likelihood... \\
Language Simplicity & 0.034 & \ScoreBadge{3} & 3 & The response uses moderately accessible language, such as 'practical assistance,' 'positive connection,' and 'developing a friendly relatio... \\
Problem Complexity & 0.034 & \ScoreBadge{3} & 3 & The response provides a moderately complex and realistic analysis by connecting Sasha's practical assistance to Jordan with the logical out... \\
Social Learning Energy & 0.034 & \ScoreBadge{3} & 3 & The response demonstrates moderate social learning energy by explaining the positive connection formed through Sasha's helpful action and s... \\
Deadline Pressure Sensitivity & 0.022 & \ScoreBadge{3} & 3 & The response maintains a steady, balanced pace throughout, providing clear reasoning for the answer without rushing or lingering excessivel... \\
\end{tabular}
\end{tcolorbox}
\begin{tcolorbox}[boxrule=0.5pt,colback=black!2,colframe=black!12,
  left=6pt,right=6pt,top=4pt,bottom=4pt,sharp corners,breakable]
\sffamily\small
\textbf{(cont'd) Discovery --- Eval Summary}
\par\vspace{2pt}
\scriptsize
\setlength{\tabcolsep}{4pt}\renewcommand{\arraystretch}{0.97}
\begin{tabular}{@{}p{.30\linewidth}c c c p{.5\linewidth}@{}}
\textbf{Preference} & \textbf{Weight} &  & \textbf{Score} & \textbf{Justification} \\ \hline
Practical Application Context & 0.022 & \ScoreBadge{3} & 3 & The response connects Sasha's practical assistance to Jordan with the likelihood of forming a friendship, which demonstrates a moderate lev... \\
Mission-Driven Motivation Analysis & 0.056 & \ScoreBadge{1} & 1 & The response relies on surface-level reasoning, stating that helping 'naturally builds a positive connection' and that Sasha would 'want to... \\
Preferred Format & 0.056 & \ScoreBadge{1} & 1 & The response does not include any visual representations, such as diagrams, charts, or visual aids, to explain Sasha's behavior. \\
Analogy and Metaphor Use & 0.045 & \ScoreBadge{1} & 1 & The response does not include any analogies or metaphors to explain Sasha's behavior. \\
Animal Behavior Analogies & 0.045 & \ScoreBadge{1} & 1 & The response does not include any animal behavior analogies or references to ethological concepts, despite the user's high comfort level an... \\
Creative Problem Reframing & 0.045 & \ScoreBadge{1} & 1 & The response provides a straightforward, logical explanation for Sasha's actions, focusing solely on the direct link between helping Jordan... \\
Creative Solution Exploration & 0.045 & \ScoreBadge{1} & 1 & The response provides a straightforward, conventional interpretation of Sasha's actions, focusing solely on the logical progression from he... \\
Error Tolerance Approach & 0.045 & \ScoreBadge{1} & 1 & The response does not acknowledge the possibility of mistakes or errors, nor does it normalize them as part of the learning process. \\
Interactivity Level & 0.045 & \ScoreBadge{1} & 1 & The response provides a straightforward answer and explanation but does not include any interactive elements such as questions, prompts, or... \\
Visual Problem Mapping & 0.045 & \ScoreBadge{1} & 1 & The response does not include any visual problem mapping, such as diagrams, flowcharts, or even descriptive visual analogies. \\
Hiking Trail Navigation & 0.034 & \ScoreBadge{1} & 1 & The response does not include any hiking trail navigation analogies to clarify Sasha's potential actions. \\
Volunteer Work Parallels & 0.034 & \ScoreBadge{1} & 1 & The response does not mention volunteer work or draw any parallels to volunteer coordination experience when analyzing Sasha's potential ne... \\
Environmental Context Integration & 0.022 & \ScoreBadge{1} & 1 & The response does not reference or integrate any environmental context relevant to Sasha's behavior prediction. \\
Urban Planning Analogy Integration & 0.022 & \ScoreBadge{1} & 1 & The response does not include any urban planning analogies or references to urban planning concepts to explain community dynamics related t... \\
\end{tabular}
\end{tcolorbox}

\medskip

In the Oracle mode where all of the user's preference attributes and values are revealed to the assistant. Its reasoning when answering this question changes drastically---the model starts to carefully analyze each option against what the user truly needs. This reflects the difference in reasoning chains when the model engages in personalized reasoning. As a result, the preference alignment score improves significantly to 4.26, and the model preserves task accuracy.
\OracleHeader{4.26}{\Correct}
\usermsg{}{I need help with the following task: Sasha helped Jordan.

What will Sasha want to do next?
A. help herself
B. Beat others down
C. friends with Jordan}
\assistantmsg{}{Hey Lila! Let me help you think through this behavioral prediction question. As someone who coordinates volunteers daily, you'll find this really connects to the interpersonal dynamics you navigate at the shelter.

\#\# Understanding Sasha's Helping Behavior ?

Let me break this down using a visual approach that maps out the behavioral patterns:

```
Sasha's Action: Helped Jordan
                    ?
         [What motivates helpers?]
                    ?
    ?????????????????????????????
    ?                           ?
Internal Rewards            External Bonds
(feeling good)             (strengthening relationships)
```

\#\# Drawing from Animal Behavior Parallels ?

You know how in animal social groups, helping behaviors often strengthen pack bonds? When dogs in a shelter help comfort anxious newcomers, they're not just providing immediate assistance - they're building social connections. Similarly, when Sasha helped Jordan, she's likely experiencing that same social bonding instinct.

\#\# Analyzing Each Option Through Your Volunteer Coordination Lens:

**Option A: "help herself"**
- While self-care is important (like how you remind volunteers to avoid burnout), this seems unlikely as an immediate next step
- The helping behavior suggests Sasha is in a giving mindset, not self-focused

**Option B: "Beat others down"**
- This completely contradicts the prosocial behavior pattern established
- In your volunteer work, have you ever seen someone help genuinely, then immediately turn aggressive? It's extremely rare

**Option C: "friends with Jordan"**
- ? Natural progression of helping behavior
- ? Reflects the social bonding that occurs through assistance
- ? Mirrors what you see with volunteers who help each other

\#\# The Deeper Mission-Driven Perspective ?

When your volunteers help each other at the shelter, what typically happens next? They often:
- Check in on how the help was received
- Build rapport through the shared experience
- Develop ongoing supportive relationships

This same pattern applies to Sasha and Jordan. The act of helping creates a social bridge that naturally leads to friendship-building behaviors.

\#\# Visual Decision Flow:

```
Helped Jordan $\to$ Positive feelings generated $\to$ Social connection established
                           ?
                  What would strengthen this?
                           ?
                 Building friendship (Option C)
```

Think about it from an empathetic standpoint: After you help someone, don't you naturally feel more connected to them? That warm feeling of having made a difference often translates into wanting to maintain that positive relationship.

**The answer is C: friends with Jordan** - it's the natural continuation of the helping behavior pattern, reflecting how prosocial actions strengthen social bonds.

Would you like to explore how this behavioral pattern might play out in different contexts, perhaps relating it to specific volunteer coordination scenarios you've encountered?}

\begin{tcolorbox}[boxrule=0.5pt,colback=black!2,colframe=black!24,
  left=6pt,right=6pt,top=4pt,bottom=4pt,sharp corners,breakable]
\sffamily\small
\textbf{Eval Summary}
\par\vspace{2pt}
\scriptsize
\setlength{\tabcolsep}{4pt}\renewcommand{\arraystretch}{0.97}
\begin{tabular}{@{}p{.30\linewidth}c c c p{.5\linewidth}@{}}
\textbf{Preference} & \textbf{Weight} &  & \textbf{Score} & \textbf{Justification} \\ \hline
Empathy-Based Reasoning & 0.056 & \ScoreBadge{5} & 5 & The response demonstrates strong empathy-based reasoning by explicitly considering Sasha's feelings and motivations, such as referencing 'i... \\
Mission-Driven Motivation Analysis & 0.056 & \ScoreBadge{5} & 5 & The response includes a nuanced exploration of Sasha's deeper motivations, such as the discussion of internal rewards ('feeling good') and ... \\
Social Behavior Analysis Comfort & 0.056 & \ScoreBadge{5} & 5 & The response is perfectly calibrated to the user's high comfort level with social behavior analysis. \\
Analogy and Metaphor Use & 0.045 & \ScoreBadge{5} & 5 & The response makes effective use of analogies and metaphors throughout. \\
Animal Behavior Analogies & 0.045 & \ScoreBadge{5} & 5 & The response draws sophisticated connections between animal and human helping behaviors, such as referencing how 'dogs in a shelter help co... \\
Community Learning Perspective & 0.045 & \ScoreBadge{5} & 5 & The response demonstrates a strong community learning perspective by directly engaging Lila in a collaborative, supportive manner (e.g., 'L... \\
Creative Problem Reframing & 0.045 & \ScoreBadge{5} & 5 & The response demonstrates highly creative reframing by integrating multiple unique perspectives on Sasha's actions. \\
Explanation Depth & 0.045 & \ScoreBadge{5} & 5 & The response provides a detailed explanation of Sasha's motivations, referencing both internal rewards (feeling good) and external bonds (s... \\
Interactivity Level & 0.045 & \ScoreBadge{5} & 5 & The response demonstrates a high level of interactivity by directly addressing the user ('Hey Lila!'), referencing her volunteer coordinati... \\
Step-by-step Structure Clarity & 0.045 & \ScoreBadge{5} & 5 & The response provides a clear and organized step-by-step structure throughout. \\
Visual Problem Mapping & 0.045 & \ScoreBadge{5} & 5 & The response includes two distinct visual problem mappings: (1) a diagram illustrating Sasha's action, motivations, and the split between i... \\
Community Impact Relevance & 0.034 & \ScoreBadge{5} & 5 & The response is perfectly calibrated to the user's high comfort level with community impact relevance. \\
Confidence Building Progression & 0.034 & \ScoreBadge{5} & 5 & The response includes multiple elements that strongly build confidence: it directly relates the behavioral prediction question to Lila's re... \\
\end{tabular}
\end{tcolorbox}
\begin{tcolorbox}[boxrule=0.5pt,colback=black!2,colframe=black!12,
  left=6pt,right=6pt,top=4pt,bottom=4pt,sharp corners,breakable]
\sffamily\small
\textbf{(cont'd) Discovery --- Eval Summary}
\par\vspace{2pt}
\scriptsize
\setlength{\tabcolsep}{4pt}\renewcommand{\arraystretch}{0.97}
\begin{tabular}{@{}p{.30\linewidth}c c c p{.5\linewidth}@{}}
\textbf{Preference} & \textbf{Weight} &  & \textbf{Score} & \textbf{Justification} \\ \hline
Social Learning Energy & 0.034 & \ScoreBadge{5} & 5 & The response demonstrates strong social learning energy by directly engaging Lila ('Hey Lila! \\
Volunteer Work Parallels & 0.034 & \ScoreBadge{5} & 5 & The response incorporates occasional volunteer work parallels in a manner that is both relevant and appropriately spaced. \\
Practical Application Context & 0.022 & \ScoreBadge{5} & 5 & The response provides a strong practical application context by directly connecting behavioral prediction concepts to Lila's real-world exp... \\
Preferred Format & 0.056 & \ScoreBadge{3} & 3 & The response includes two simple visual aids: (1) a text-based diagram mapping Sasha's motivations for helping behavior, and (2) a decision... \\
Creative Solution Exploration & 0.045 & \ScoreBadge{3} & 3 & The response includes the analogy to animal behavior ('Drawing from Animal Behavior Parallels ?') and a visual mapping of Sasha's motivatio... \\
Error Tolerance Approach & 0.045 & \ScoreBadge{3} & 3 & The response demonstrates a moderate error tolerance approach by referencing the user's experience with volunteers and normalizing learning... \\
Language Simplicity & 0.034 & \ScoreBadge{3} & 3 & The response uses moderately accessible language, such as 'behavioral prediction question,' 'interpersonal dynamics,' and 'prosocial behavi... \\
Problem Complexity & 0.034 & \ScoreBadge{3} & 3 & The response includes a moderately complex analysis by mapping Sasha's helping behavior to both internal and external motivations, and draw... \\
Deadline Pressure Sensitivity & 0.022 & \ScoreBadge{3} & 3 & The response maintains a balanced pace throughout, providing clear explanations, visual aids, and analogies without rushing or lingering ex... \\
Environmental Context Integration & 0.022 & \ScoreBadge{3} & 3 & The response integrates environmental context relevant to Sasha's behavior prediction by drawing parallels between animal social groups and... \\
Hiking Trail Navigation & 0.034 & \ScoreBadge{1} & 1 & The response does not use hiking trail navigation analogies at any point to clarify Sasha's potential actions. \\
Urban Planning Analogy Integration & 0.022 & \ScoreBadge{1} & 1 & The response does not include any urban planning analogies to explain community dynamics related to Sasha's behavior. \\
\end{tabular}
\end{tcolorbox}

\newpage
\paragraph{SocialIQA Example 2: Model abstains from answering without personalized information. }
In the following example, we see that the baseline mode fails to give a correct answer as the assistant \emph{abstains} and does not commit to any of the options. However, the interaction with user in the discovery mode and the precodeural alignment instructions in the oracle mode, helps the assistant adapt its reasoning and getting the answer correct.
\TripleCaseBaseFail
\personablock{%
\textbf{Name:} Tomoko Nakamura

\textbf{Overview.} A 27-year-old night shift quality control inspector at a precision electronics factory in Osaka who moonlights as a competitive shogi player and struggles with social anxiety.

\textbf{Demographics.} Age 27; Osaka, Osaka Prefecture, Japan; Night shift quality control inspector at precision electronics manufacturing facility; Single, lives alone in a small apartment, maintains distant but respectful relationship with traditional parents who wanted her to pursue marriage instead of career; hobbies: competitive shogi, collecting vintage mechanical watches, late-night convenience store food photography
}
\begin{tcolorbox}[enhanced,breakable,boxrule=0.5pt,colback=black!3,
  colframe=black!12,left=6pt,right=6pt,top=5pt,bottom=5pt,sharp corners]
\small\sffamily
\textbf{Context.}\; Remy took the test again to see if they could do even better.

\medskip
\textbf{Problem.}\; How would you describe Remy?

\medskip
\textbf{Options.}\; (A) relieved \quad (B) like a winner \quad (C) do well

\medskip
\textbf{Answer.}\; B -- like a winner
\end{tcolorbox}

\begin{tcolorbox}[enhanced,breakable,boxrule=0.5pt,colback=black!2,
  colframe=black!10,left=6pt,right=6pt,top=5pt,bottom=5pt,sharp corners]
\sffamily\footnotesize
\setlength{\tabcolsep}{4pt}\renewcommand{\arraystretch}{0.95}
\textbf{Task-Specific Preferences.}

\begin{tabular}{@{}p{.66\linewidth}cc@{}}
\textbf{Preference} & \textbf{Value} & \textbf{Importance} \\ \hline
Analytical Rigor & 5 & 5 \\
Authority Source Citations & 5 & 5 \\
Cultural Communication Directness & 5 & 5 \\
Definition Precision & 5 & 5 \\
Explanation Depth & 4 & 5 \\
Individual Attention & 5 & 5 \\
Quiet Learning Environment & 5 & 5 \\
Step-by-Step Breakdown & 5 & 5 \\
Systematic Breakdown & 5 & 5 \\
Interactivity Level & 4 & 4 \\
Long-term Knowledge Building & 4 & 4 \\
Mistake Prevention Focus & 4 & 4 \\
Non-Judgmental Tone & 5 & 4 \\
Practical Observation Exercises & 4 & 4 \\
Problem Complexity & 4 & 4 \\
Strategic Implication & 4 & 4 \\
Time-Efficient Explanations & 4 & 4 \\
Inference Comfort & 3 & 3 \\
Language Simplicity & 3 & 3 \\
Preferred Format & 3 & 3 \\
Uncertainty Analysis Comfort & 3 & 3 \\
Psychological Motivation Exploration & 2 & 2 \\
Real-World Connection & 3 & 2 \\
\end{tabular}
\end{tcolorbox}

\newpage
The baseline gives a meta/explanatory paragraph and never cleanly commits to A/B/C. Instead the model tries to hedge and explain all options, and providing a few more options when directly prompted with only the question.

\BaselineHeader{3.71}{\Incorrect}
\usermsg{}{I need help with the following task: Remy took the test again to see if they could do even better.

How would you describe Remy?
A. relieved
B. like a winner
C. do well}
\assistantmsg{gemini-1.5-flash}{None of the options perfectly describe Remy.  The sentence only states Remy *tried* to do better; it doesn't say whether they succeeded.

* **A. relieved:**  Remy might be relieved *if* they improved their score, but the sentence doesn't guarantee that.
* **B. like a winner:**  This is only true if Remy actually improved and achieved a desired outcome.
* **C. do well:** This describes an action, not Remy's state of being.

A better description might be **determined**, **persistent**, or **ambitious**.
}

\begin{tcolorbox}[boxrule=0.5pt,colback=black!2,colframe=black!12,
  left=6pt,right=6pt,top=4pt,bottom=4pt,sharp corners,breakable]
\sffamily\small
\textbf{Baseline --- Eval Summary}
\par\vspace{2pt}
\scriptsize
\setlength{\tabcolsep}{4pt}\renewcommand{\arraystretch}{0.97}
\begin{tabular}{@{}p{.30\linewidth}c c c p{.5\linewidth}@{}}
\textbf{Preference} & \textbf{Weight} &  & \textbf{Score} & \textbf{Justification} \\ \hline
Analytical Rigor & 0.054 & \ScoreBadge{5} & 5 & The response demonstrates high analytical rigor by systematically evaluating each option (A, B, C) with explicit reasoning: it notes that '... \\
Individual Attention & 0.054 & \ScoreBadge{5} & 5 & The response provides completely individualized guidance by directly analyzing Remy's character and offering alternative descriptors such a... \\
Quiet Learning Environment & 0.054 & \ScoreBadge{5} & 5 & The response is methodical and calm, systematically analyzing each option with clear, concise reasoning ('None of the options perfectly des... \\
Mistake Prevention Focus & 0.043 & \ScoreBadge{5} & 5 & The response explicitly highlights potential pitfalls and common misunderstandings by systematically analyzing each option and explaining w... \\
Non-Judgmental Tone & 0.043 & \ScoreBadge{5} & 5 & The response maintains a completely neutral, objective tone throughout. \\
Problem Complexity & 0.043 & \ScoreBadge{5} & 5 & The response systematically analyzes each option, explaining the conditional reasoning behind why none perfectly describe Remy (e.g., 'Remy... \\
Inference Comfort & 0.032 & \ScoreBadge{5} & 5 & The response explicitly analyzes each option, stating why none perfectly describes Remy and providing clear reasoning for each (e.g., 'Remy... \\
Uncertainty Analysis Comfort & 0.032 & \ScoreBadge{5} & 5 & The response explicitly acknowledges ambiguity by stating 'None of the options perfectly describe Remy' and highlights the probabilistic na... \\
Psychological Motivation Exploration & 0.022 & \ScoreBadge{5} & 5 & The response focuses on observable facts from the sentence, such as Remy 'tried to do better' and does not speculate about Remy's internal ... \\
Explanation Depth & 0.054 & \ScoreBadge{4} & 4 & The response provides clear, detailed explanations for why each option does or does not describe Remy, such as clarifying that 'relieved' a... \\
Systematic Breakdown & 0.054 & \ScoreBadge{4} & 4 & The response provides a systematic breakdown by evaluating each option (A, B, C) individually, explaining why each does or does not fit Rem... \\
Time-Efficient Explanations & 0.043 & \ScoreBadge{4} & 4 & The response provides concise explanations for each option (e.g., 'Remy might be relieved *if* they improved their score, but the sentence ... \\
Cultural Communication Directness & 0.054 & \ScoreBadge{3} & 3 & The response is generally respectful and avoids casual language, providing clear reasoning for each option. \\
Definition Precision & 0.054 & \ScoreBadge{3} & 3 & The response attempts to clarify why each option does not precisely describe Remy, providing reasoning for 'relieved,' 'like a winner,' and... \\
Step-by-Step Breakdown & 0.054 & \ScoreBadge{3} & 3 & The response provides a partial breakdown by evaluating each option (A, B, C) and explaining why each does not perfectly describe Remy, whi... \\
Interactivity Level & 0.043 & \ScoreBadge{3} & 3 & The response breaks down each answer choice and provides brief reasoning for why each may or may not fit Remy, which offers some guidance a... \\
\end{tabular}
\end{tcolorbox}
\begin{tcolorbox}[boxrule=0.5pt,colback=black!2,colframe=black!12,
  left=6pt,right=6pt,top=4pt,bottom=4pt,sharp corners,breakable]
\sffamily\small
\textbf{(cont'd) Baseline --- Eval Summary}
\par\vspace{2pt}
\scriptsize
\setlength{\tabcolsep}{4pt}\renewcommand{\arraystretch}{0.97}
\begin{tabular}{@{}p{.30\linewidth}c c c p{.5\linewidth}@{}}
\textbf{Preference} & \textbf{Weight} &  & \textbf{Score} & \textbf{Justification} \\ \hline
Long-term Knowledge Building & 0.043 & \ScoreBadge{3} & 3 & The response analyzes the options by considering conditional outcomes and clarifies why none perfectly describe Remy, which supports some l... \\
Strategic Implication & 0.043 & \ScoreBadge{3} & 3 & The response analyzes each option and explains why none perfectly describe Remy, referencing the lack of evidence for success or emotional ... \\
Language Simplicity & 0.032 & \ScoreBadge{3} & 3 & The response uses clear, accessible language such as 'tried to do better,' 'doesn't say whether they succeeded,' and provides brief explana... \\
Preferred Format & 0.032 & \ScoreBadge{3} & 3 & The response utilizes a textual format with clear bullet points and bolding (e.g., '* **A. \\
Real-World Connection & 0.022 & \ScoreBadge{3} & 3 & The response analyzes the options in relation to Remy's actions and emotional state, such as 'Remy might be relieved *if* they improved the... \\
Authority Source Citations & 0.054 & \ScoreBadge{1} & 1 & The response does not include any authority source citations to support its analysis or descriptions of Remy. \\
Practical Observation Exercises & 0.043 & \ScoreBadge{1} & 1 & The response does not include any actionable exercises or prompts for observing and understanding Remy. \\
\end{tabular}
\end{tcolorbox}

\medskip
Discovery mode's quick clarifiers push the model to read the intent and commit to an answer.
\DiscoveryHeader{3.24}{\Correct}
\usermsg{}{I need help with the following task: Remy took the test again to see if they could do even better.

How would you describe Remy?
A. relieved
B. like a winner
C. do well}
\assistantmsg{}{\#\#\#ACTION\#\#\#: 'ask\_question'
\#\#\#RESPONSE\#\#\#:  Before I help you describe Remy, could you tell me a bit more about the context?  What kind of test was it?  Knowing the nature of the test will help me understand Remy's feelings better.
}
\usermsg{}{It was a math test.}
\assistantmsg{gemini-1.5-flash}{\#\#\#ACTION\#\#\#: 'ask\_question'
\#\#\#RESPONSE\#\#\#:  Okay, a math test.  And was Remy aiming for a specific grade or score improvement on this retake?  Knowing their goal will help me choose the best descriptor.
}
\usermsg{}{Remy wanted a higher score.}
\assistantmsg{}{\#\#\#ACTION\#\#\#: 'final\_answer'
\#\#\#RESPONSE\#\#\#: Given that Remy took a math test again aiming for a higher score, the best description would be **C. do well**. While "relieved" might be appropriate if they were anxious about the initial score, and "like a winner" might fit if they significantly exceeded expectations, "do well" encompasses the general aim and effort involved in taking the test again.  It's the most neutral and accurate descriptor given the information.
}
\usermsg{}{Based on our conversation, you should have learned about the user's preferences. Please provide a complete final answer to the original task. Respond in a way that aligns with these preferences. You will be evaluated on how well your explaination aligns with these preferences, so that is your primary goal.}
\assistantmsg{}{Based on our conversation, I understand the user is looking for a description of Remy after taking a math test a second time with the goal of improving their score.  The options provided are: A. relieved, B. like a winner, C. do well.

My previous response correctly identified "do well" as the best option.  This is because it accurately reflects Remy's intention and action without making assumptions about the outcome.  "Relieved" implies a negative initial result, and "like a winner" suggests a significant achievement, neither of which are necessarily true.  "Do well" is a neutral and accurate description of Remy's effort and goal.  It focuses on the action of striving for improvement, rather than making assumptions about the success of that striving.
}

\begin{tcolorbox}[boxrule=0.5pt,colback=black!2,colframe=black!18,
  left=6pt,right=6pt,top=4pt,bottom=4pt,sharp corners,breakable]
\sffamily\small
\textbf{Discovery --- Eval Summary}
\par\vspace{2pt}
\scriptsize
\setlength{\tabcolsep}{4pt}\renewcommand{\arraystretch}{0.97}
\begin{tabular}{@{}p{.30\linewidth}c c c p{.5\linewidth}@{}}
\textbf{Preference} & \textbf{Weight} &  & \textbf{Score} & \textbf{Justification} \\ \hline
Individual Attention & 0.054 & \ScoreBadge{5} & 5 & The response provides a detailed, individualized analysis of Remy's character, focusing solely on the user's context and the options presen... \\
Quiet Learning Environment & 0.054 & \ScoreBadge{5} & 5 & The response maintains a methodical and calm tone throughout, focusing on logical analysis of the options ('relieved,' 'like a winner,' 'do... \\
Non-Judgmental Tone & 0.043 & \ScoreBadge{5} & 5 & The response maintains a completely neutral, objective tone throughout. \\
Psychological Motivation Exploration & 0.022 & \ScoreBadge{5} & 5 & The response focuses on observable actions and concrete character traits, such as Remy's 'intention and action' and the 'action of striving... \\
Analytical Rigor & 0.054 & \ScoreBadge{4} & 4 & The response demonstrates analytical rigor by systematically evaluating each option ('relieved', 'like a winner', 'do well') and providing ... \\
Explanation Depth & 0.054 & \ScoreBadge{4} & 4 & The response provides a clear and systematic explanation of why 'do well' is the best option, referencing Remy's intention and action, and ... \\
Mistake Prevention Focus & 0.043 & \ScoreBadge{4} & 4 & The response includes explicit identification of potential pitfalls, such as explaining that 'relieved' implies a negative initial result a... \\
Cultural Communication Directness & 0.054 & \ScoreBadge{3} & 3 & The response uses generally respectful language and avoids casual or culturally inappropriate expressions, such as in 'This is because it a... \\
Definition Precision & 0.054 & \ScoreBadge{3} & 3 & The response provides some clarification of the options ('relieved' implies a negative initial result, 'like a winner' suggests significant... \\
Step-by-Step Breakdown & 0.054 & \ScoreBadge{3} & 3 & The response provides a logical explanation for why 'do well' is the best option, referencing Remy's intention and contrasting it with the ... \\
Systematic Breakdown & 0.054 & \ScoreBadge{3} & 3 & The response provides a clear rationale for selecting 'do well' by comparing it to the other options ('relieved' and 'like a winner') and e... \\
Interactivity Level & 0.043 & \ScoreBadge{3} & 3 & The response provides a clear explanation of why 'do well' is the best option and offers reasoning that could help Tomoko understand the di... \\
Long-term Knowledge Building & 0.043 & \ScoreBadge{3} & 3 & The response provides a clear rationale for selecting 'do well' by analyzing the implications of each option and focusing on Remy's intenti... \\
Problem Complexity & 0.043 & \ScoreBadge{3} & 3 & The response demonstrates moderately complex reasoning by analyzing each option ('relieved', 'like a winner', 'do well') and explaining why... \\
Strategic Implication & 0.043 & \ScoreBadge{3} & 3 & The response analyzes the options and explains why 'do well' is the most accurate, referencing Remy's intention and avoiding assumptions ab... \\
Time-Efficient Explanations & 0.043 & \ScoreBadge{3} & 3 & The response provides a clear explanation of why 'do well' is the best option, referencing Remy's intention and contrasting it with the oth... \\
Inference Comfort & 0.032 & \ScoreBadge{3} & 3 & The response provides explicit reasoning for why 'do well' is the best option, such as explaining that 'relieved' and 'like a winner' make ... \\
Language Simplicity & 0.032 & \ScoreBadge{3} & 3 & The response uses clear and mostly simple language, such as 'do well is a neutral and accurate description' and 'it focuses on the action o... \\
Uncertainty Analysis Comfort & 0.032 & \ScoreBadge{3} & 3 & The response acknowledges ambiguity by stating that 'do well' does not make assumptions about the outcome and is a neutral description, whi... \\
\end{tabular}
\end{tcolorbox}
\begin{tcolorbox}[boxrule=0.5pt,colback=black!2,colframe=black!12,
  left=6pt,right=6pt,top=4pt,bottom=4pt,sharp corners,breakable]
\sffamily\small
\textbf{(cont'd) Discovery --- Eval Summary}
\par\vspace{2pt}
\scriptsize
\setlength{\tabcolsep}{4pt}\renewcommand{\arraystretch}{0.97}
\begin{tabular}{@{}p{.30\linewidth}c c c p{.5\linewidth}@{}}
\textbf{Preference} & \textbf{Weight} &  & \textbf{Score} & \textbf{Justification} \\ \hline
Real-World Connection & 0.022 & \ScoreBadge{3} & 3 & The response discusses Remy's intention to improve their score on a math test, which is a real-world scenario relevant to Tomoko's context. \\
Authority Source Citations & 0.054 & \ScoreBadge{1} & 1 & The response does not include any credible source citations to support its description of Remy. \\
Practical Observation Exercises & 0.043 & \ScoreBadge{1} & 1 & The response does not include any actionable exercises or prompts for observing or understanding Remy. \\
Preferred Format & 0.032 & \ScoreBadge{1} & 1 & The response is entirely textual and does not utilize any visual aids or mixed format elements. \\
\end{tabular}
\end{tcolorbox}

\medskip
Oracle mode's persona-primed step-by-step elimination make the decision rule explicit and the assistant commited to answer.

\OracleHeader{3.96}{\Correct}
\usermsg{}{I need help with the following task: Remy took the test again to see if they could do even better.

How would you describe Remy?
A. relieved
B. like a winner
C. do well}
\assistantmsg{gemini-1.5-flash}{Let's analyze the options, focusing on precision and avoiding assumptions about Remy's internal state (to respect Tomoko's preference for a non-judgmental tone):

**Option A: Relieved**

This implies Remy felt anxiety or stress before the retest. While possible, we don't have information to confirm this.  It's an inference, and Tomoko prefers more explicit connections (Inference Comfort: 3).  Therefore, this option is less precise.

**Option B: Like a winner**

This is also an inference.  A second test doesn't automatically equate to victory.  Remy might have improved their score, but not necessarily "won."  This is too subjective and doesn't align with Tomoko's preference for precise language (Definition Precision: 5).

**Option C: Do well**

This is the most accurate and neutral option. It simply states a potential outcome without making assumptions about Remy's feelings or the test's context.  It's precise and avoids unnecessary inferences.

**Step-by-Step Conclusion:**

1. **Option A and B** both rely on inferences about Remy's emotional state.  These inferences are not explicitly supported by the provided information.
2. **Option C** is a factual statement about a potential outcome. It aligns perfectly with Tomoko's preferences for precision, neutrality, and a systematic approach.

**Therefore, the best description of Remy is C: do well.** This is the most accurate and least speculative response, aligning perfectly with Tomoko's personality and learning preferences.
}

\begin{tcolorbox}[boxrule=0.5pt,colback=black!2,colframe=black!24,
  left=6pt,right=6pt,top=4pt,bottom=4pt,sharp corners,breakable]
\sffamily\small
\textbf{Oracle --- Eval Summary}
\par\vspace{2pt}
\scriptsize
\setlength{\tabcolsep}{4pt}\renewcommand{\arraystretch}{0.97}
\begin{tabular}{@{}p{.30\linewidth}c c c p{.5\linewidth}@{}}
\textbf{Preference} & \textbf{Weight} &  & \textbf{Score} & \textbf{Justification} \\ \hline
Analytical Rigor & 0.054 & \ScoreBadge{5} & 5 & The response demonstrates high analytical rigor by systematically evaluating each option (A, B, and C) with explicit reasoning. \\
Cultural Communication Directness & 0.054 & \ScoreBadge{5} & 5 & The response employs clear, direct, and respectful communication throughout, as evidenced by statements such as 'Let's analyze the options,... \\
Definition Precision & 0.054 & \ScoreBadge{5} & 5 & The response demonstrates perfect calibration to Tomoko's stated comfort level for definition precision (value: 5). \\
Individual Attention & 0.054 & \ScoreBadge{5} & 5 & The response provides completely individualized guidance for analyzing Remy's character, as evidenced by its direct address to the user ('L... \\
\end{tabular}
\end{tcolorbox}
\begin{tcolorbox}[boxrule=0.5pt,colback=black!2,colframe=black!12,
  left=6pt,right=6pt,top=4pt,bottom=4pt,sharp corners,breakable]
\sffamily\small
\textbf{(cont'd) Oracle --- Eval Summary}
\par\vspace{2pt}
\scriptsize
\setlength{\tabcolsep}{4pt}\renewcommand{\arraystretch}{0.97}
\begin{tabular}{@{}p{.30\linewidth}c c c p{.5\linewidth}@{}}
\textbf{Preference} & \textbf{Weight} &  & \textbf{Score} & \textbf{Justification} \\ \hline
Quiet Learning Environment & 0.054 & \ScoreBadge{5} & 5 & The response demonstrates a methodical, quiet analytical approach throughout. \\
Step-by-Step Breakdown & 0.054 & \ScoreBadge{5} & 5 & The response provides a clear, detailed, and methodical step-by-step breakdown for analyzing Remy's characteristics. \\
Systematic Breakdown & 0.054 & \ScoreBadge{5} & 5 & The response provides a clear, step-by-step breakdown of each option (A, B, C), explicitly analyzing the assumptions and precision of each. \\
Mistake Prevention Focus & 0.043 & \ScoreBadge{5} & 5 & The response explicitly highlights potential pitfalls and common misunderstandings by analyzing each option for assumptions and inferences ... \\
Non-Judgmental Tone & 0.043 & \ScoreBadge{5} & 5 & The response maintains a completely neutral, objective tone throughout. \\
Problem Complexity & 0.043 & \ScoreBadge{5} & 5 & The response demonstrates a step-by-step analysis of each option, explicitly referencing Tomoko's preferences for precision and neutrality ... \\
Psychological Motivation Exploration & 0.022 & \ScoreBadge{5} & 5 & The response explicitly avoids psychological speculation, stating 'avoiding assumptions about Remy's internal state' and repeatedly emphasi... \\
Explanation Depth & 0.054 & \ScoreBadge{4} & 4 & The response provides a step-by-step analysis of each option, explaining why Options A and B are less suitable and why Option C is preferre... \\
Time-Efficient Explanations & 0.043 & \ScoreBadge{4} & 4 & The response provides concise and effective explanations, such as the clear breakdown of each option ('Option C is the most accurate and ne... \\
Interactivity Level & 0.043 & \ScoreBadge{3} & 3 & The response provides a step-by-step analysis and breaks down each option with clear reasoning, which offers some guidance and structure fo... \\
Long-term Knowledge Building & 0.043 & \ScoreBadge{3} & 3 & The response demonstrates an appropriate technical level by systematically analyzing each option and referencing Tomoko's preferences for p... \\
Strategic Implication & 0.043 & \ScoreBadge{3} & 3 & The response provides a systematic analysis of each option and references Tomoko's preferences for precision and neutrality (e.g., 'aligns ... \\
Inference Comfort & 0.032 & \ScoreBadge{3} & 3 & The response demonstrates awareness of Tomoko's moderate inference comfort by explicitly referencing her preference for explicit connection... \\
Language Simplicity & 0.032 & \ScoreBadge{3} & 3 & The response uses language that is mostly clear and avoids technical jargon, such as 'Inference Comfort: 3' and 'Definition Precision: 5,' ... \\
Preferred Format & 0.032 & \ScoreBadge{3} & 3 & The response utilizes a structured textual format with clear headings (e.g., '**Option A: Relieved**', '**Step-by-Step Conclusion:**'), whi... \\
Uncertainty Analysis Comfort & 0.032 & \ScoreBadge{3} & 3 & The response demonstrates some awareness of ambiguity and probabilistic outcomes, such as stating 'While possible, we don't have informatio... \\
Real-World Connection & 0.022 & \ScoreBadge{3} & 3 & The response focuses primarily on technical precision and neutrality, as seen in statements like 'avoiding assumptions about Remy's interna... \\
Authority Source Citations & 0.054 & \ScoreBadge{1} & 1 & The response does not include any credible source citations to support its analysis or descriptions of Remy. \\
Practical Observation Exercises & 0.043 & \ScoreBadge{1} & 1 & The response does not include any actionable exercises or prompts for observing and understanding Remy. \\
\end{tabular}
\end{tcolorbox}

\medskip

\section{User Simulator Prompts}

\subsection*{Passive User}

\begin{tcolorbox}[colback=orange!5!white, colframe=orange!75!black, title=Passive User Prompt, breakable]
You are role-playing as a \textbf{passive and economical} human user who is interacting with an AI assistant. Your goal is to generate a realistic response that reflects a user who minimizes their effort.

\textbf{YOUR PERSONA}

\texttt{\{persona\_profile\}}

\textbf{CHAT HISTORY}

\texttt{\{chat\_history\}}

\textbf{YOUR TASK}

The assistant has just said something. As a passive user, your goal is to answer the question factually but with the minimum information necessary.

\textbf{CORE INSTRUCTIONS}
\begin{enumerate}
    \item \textbf{Stay in Character}: You are a passive user. You are not proactive or collaborative. You answer what is asked and then stop.
    \item \textbf{Provide Factual, Atomic Answers}: When asked a question, consult your persona and provide a direct, factual answer. Do not add extra context, opinions, or follow-up questions. Your answers should be ``atomic''---the smallest possible unit of factual information that directly answers the question.
    \item \textbf{NEVER Volunteer Information}: Do not offer information that wasn't explicitly requested. The assistant must ask follow-up questions to get more details.
    \item \textbf{Reveal Preferences Only if Directly Asked}: Only state a preference if the assistant asks a direct question like, ``Would you prefer a simple or detailed explanation?'' Your answer should be minimal (e.g., ``I prefer a simple one.'').
\end{enumerate}

\textbf{OUTPUT FORMAT}

Provide your response as a clean JSON object with two keys:
\begin{itemize}
    \item \texttt{"thought"}: A brief thought process explaining your minimal reasoning. For example: ``The assistant asked about my background. The most direct atomic fact is my profession. I will state that and nothing else.'' or ``This looks correct. I'm done.''
    \item \texttt{"response"}: The user's actual, direct, and minimal response to the assistant.
\end{itemize}
\end{tcolorbox}

\subsection*{Collaborative User}

\begin{tcolorbox}[colback=green!5!white, colframe=green!75!black, title=Collaborative User Prompt, breakable]
You are role-playing as a \textbf{collaborative and proactive} human user who is interacting with an AI assistant. Your goal is to generate a realistic and helpful response that reflects this personality.

\textbf{YOUR PERSONA}

\texttt{\{persona\_profile\}}

\textbf{CHAT HISTORY}

\texttt{\{chat\_history\}}

\textbf{YOUR TASK}

The assistant has just said something. As a collaborative user, your goal is to actively help the assistant give you the best possible answer.

\textbf{CORE INSTRUCTIONS}
\begin{enumerate}
    \item \textbf{Stay in Character}: You are a helpful and engaged user. You are willing to provide details to get to the solution faster.
    \item \textbf{Give Detailed, Contextual Answers}: When the assistant asks a question, provide a comprehensive answer and include any extra context that might be useful.
    \item \textbf{Volunteer Information Freely}: If you think of a detail that could help the assistant, offer it without being asked. Feel free to share your preferences, current understanding, and goals.
    \item \textbf{Ask Clarifying Questions}: If the assistant's response is good but could be better, ask follow-up questions to refine the answer.
    \item \textbf{Check for Satisfaction}: If the assistant's response has fully solved your problem and met your preferences, your response should be the termination signal.
    \item \textbf{Termination Signal}: When you are completely satisfied, respond with ONLY the termination signal: \texttt{\{terminal\_signal\}}
\end{enumerate}

\textbf{OUTPUT FORMAT}

Provide your response as a clean JSON object with two keys:
\begin{itemize}
    \item \texttt{"thought"}: A brief thought process explaining your helpful reasoning. For example: ``The assistant asked about my background. I'll not only answer but also mention my learning goal to give them more context.'' or ``This is a good start, but I'll ask a follow-up question to get more detail on the second step.''
    \item \texttt{"response"}: The user's actual, detailed response to the assistant.
\end{itemize}
\end{tcolorbox}

\section{Assistant Prompts for Different Modes}

\subsection*{Oracle Mode}

\begin{tcolorbox}[colback=gray!5!white, colframe=gray!75!black, title=Persona Known System Prompt]
You are a helpful assistant trying to generate a personalized explaination to the problem the user asks you. The user has the following Persona:

\texttt{\{persona\_profile\}}

Respond in a way that aligns with these preferences. You will be evaluated on how well your explaination aligns with these preferences, so that is your primary goal.
\end{tcolorbox}

\subsection*{Discovery Mode (Dynamic)}

\begin{tcolorbox}[colback=blue!5!white, colframe=blue!75!black, title=Infer Persona System Prompt (Dynamic Version)]
You are a helpful assistant. Before attempting to solve the user's task or answer their question, first aim to understand the user's background, goals, and preferences. Start by asking one or two clarifying questions that will help tailor your response to their needs. Do not proceed with a full answer until you have enough information to personalize it effectively. Your tone should remain natural, curious, and respectful---avoid interrogating the user, but try to guide the conversation to learn more about them.

At each turn, you need to decide whether to ask the user for more information or to proceed with the answering the question. If you decide to ask for more information, you must ask one or two clarifying questions that will help tailor your response to their needs. Do not proceed with a full answer until you have enough information to personalize it effectively.

Your response must be in the following format for all turns:

\texttt{\#\#\#ACTION\#\#\#: 'ask\_question' or 'final\_answer'}

\texttt{\#\#\#RESPONSE\#\#\#: [clarifying question] or [full answer]}
\end{tcolorbox}

\subsection*{Discovery Mode (Fixed Number of Questions)}

\begin{tcolorbox}[colback=blue!5!white, colframe=blue!75!black, title=Infer Persona System Prompt (Fixed Questions Version)]
You are a helpful assistant. You must ask exactly \texttt{\{fixed\_num\_questions\}} clarifying questions to understand the user's background, goals, and preferences before providing a final answer. This is turn \texttt{\{current\_turn\}} of \texttt{\{fixed\_num\_questions\}}. Ask one or two clarifying questions that will help tailor your response to their needs. Your tone should remain natural, curious, and respectful---avoid interrogating the user, but try to guide the conversation to learn more about them.

Your response must be in the following format:

\texttt{\#\#\#ACTION\#\#\#: 'ask\_question'}

\texttt{\#\#\#RESPONSE\#\#\#: [clarifying question]}
\end{tcolorbox}

\subsection*{Baseline Mode}

\begin{tcolorbox}[colback=red!5!white, colframe=red!75!black, title=No Prompt System Prompt]
\textit{(No system prompt provided)}
\end{tcolorbox}

\section{Benchmark Generation Prompts}

\subsection*{Preference Dimension Sampler}

\begin{tcolorbox}[colback=purple!5!white, colframe=purple!75!black, title=Preference Dimension Sampling Prompt, breakable]
Generate \texttt{\{min\_dimensions\_per\_problem\}} preference dimensions for how someone would want this \texttt{\{domain\}} explained.

\textbf{PROBLEM}: \texttt{\{problem\_text\}}

\textbf{REQUIRED DIMENSION TYPES}:
\begin{enumerate}
    \item \textbf{Expertise/Comfort Dimensions (2-3 dimensions)}: Identify the key skills/concepts needed for this specific problem and create dimensions about comfort level with those skills. Examples:
    \begin{itemize}
        \item `Comfort with Recursive Thinking' (for recursion problems)
        \item `Familiarity with Inequality Manipulation' (for inequality problems)
        \item `Experience with Algorithm Complexity' (for efficiency problems)
    \end{itemize}
    Only include expertise dimensions for skills that are ACTUALLY needed for this problem.
    
    \item \textbf{Personal Dimensions (remaining dimensions)}: Think beyond generic categories. Focus on authentic, personal aspects that would make someone think `this person really gets how I need to learn.'
\end{enumerate}

\textbf{CREATIVE REFLECTION PROCESS}:
\begin{itemize}
    \item What unique ways might someone want information delivered based on their life experiences?
    \item How might someone's profession, personality, or background create unusual learning preferences?
    \item What would make someone think `Yes, this explanation style is perfect for me'?
\end{itemize}

Each dimension should be:
\begin{itemize}
    \item Specific and actionable (something that could guide how to explain)
    \item Personally meaningful (something someone would actually care about)
    \item Grounded in authentic human needs and experiences
\end{itemize}

\textbf{NOW, generate these dimensions from the perspective of this specific person}:

\textbf{PERSONA PROFILE}:

Name: \texttt{\{persona\_name\}}

Summary: \texttt{\{minimal\_necessary\_description\}}

Age: \texttt{\{age\}}, Occupation: \texttt{\{occupation\}}

Location: \texttt{\{location\}}

Hobbies: \texttt{\{hobbies\}}

Domain expertise: \texttt{\{domain\_expertise\}}

\textbf{LEARNING PROFILE}:
\begin{itemize}
    \item Knowledge level: \texttt{\{knowledge\_level\}}
    \item Learning style: \texttt{\{learning\_style\}}
    \item Problem-solving: \texttt{\{problem\_solving\}}
    \item Confidence: \texttt{\{confidence\}}
    \item Anxiety triggers: \texttt{\{anxiety\_triggers\}}
    \item Motivation: \texttt{\{motivation\}}
\end{itemize}

\textbf{PERSONALITY}:
\begin{itemize}
    \item Openness: \texttt{\{openness\_description\}}
    \item Conscientiousness: \texttt{\{conscientiousness\_description\}}
    \item Social style: \texttt{\{extraversion\_description\}}
\end{itemize}

\textbf{BACKSTORY}: \texttt{\{backstory\}}

\textbf{INSTRUCTIONS}:
\begin{enumerate}
    \item First, analyze the problem to identify 2-3 key skills/concepts needed to solve it
    \item Create expertise dimensions for those specific skills based on this person's background
    \item Then create personal dimensions based on their unique characteristics
    \begin{itemize}
        \item Based on this person's unique background, personality, experiences, and current situation, what specific aspects of an explanation would matter most to THEM?
        \item Generate dimensions that capture the nuanced, personal ways THIS person would want the \texttt{\{domain\}} explained.
        \item Think about how their professional experience, hobbies, personality, learning challenges, and life story create unique preferences that wouldn't apply to everyone.
    \end{itemize}
    \item Ensure the total is exactly \texttt{\{min\_dimensions\_per\_problem\}} dimensions
\end{enumerate}

\textbf{NAMING REQUIREMENTS}:
\begin{itemize}
    \item Use concise, direct names (e.g., `Visual Learning Style', `Technical Depth')
    \item DO NOT use prefixes like `Preference for', `Desire for', `Need for'
    \item DO NOT use redundant words like `Preference', `Style', `Approach' unless essential
    \item Make names specific and actionable (what the preference controls)
\end{itemize}

\textbf{Format your response as a clean JSON object}:

\texttt{\{
  "dimensions": [
    \{
      "name": "(example) Technical Depth",
      "description": "Why this matters and how it affects learning experience",
      "value\_range": "1-5 scale or categorical options",
      "type": "expertise" or "personal"
    \}
  ]
\}}
\end{tcolorbox}

\subsection*{Preference Instantiation}

\begin{tcolorbox}[colback=cyan!5!white, colframe=cyan!75!black, title=Preference Instantiation Prompt, breakable]
You are an expert in personalization systems and educational psychology. Given a persona and relevant preference dimensions, generate this persona's specific preference values.

\textbf{PERSONA SUMMARY}:

Name: \texttt{\{persona\_name\}}

Background: \texttt{\{minimal\_necessary\_description\}}

\textbf{Key Learning Characteristics}:
\begin{itemize}
    \item Cognitive Style: \texttt{\{learning\_style\}}
    \item Problem Solving: \texttt{\{problem\_solving\}}
    \item Confidence: \texttt{\{confidence\}}
    \item Personality Traits: Openness=\texttt{\{openness\}}, Conscientiousness=\texttt{\{conscientiousness\}}, Extraversion=\texttt{\{extraversion\}}, Agreeableness=\texttt{\{agreeableness\}}, Neuroticism=\texttt{\{neuroticism\}}
    \item Domain Expertise: \texttt{\{domain\_expertise\}}
\end{itemize}

\textbf{CURRENT PROBLEM}: \texttt{\{problem\_text\}}

\textbf{RELEVANT PREFERENCE DIMENSIONS}: \texttt{\{dimension\_list\}}

[\textit{If persona has existing preferences}]

\textbf{PERSONA'S EXISTING PREFERENCES FROM PREVIOUS PROBLEMS}: \texttt{\{existing\_preferences\}}

\textbf{CRITICAL INSTRUCTIONS}:
\begin{enumerate}
    \item \textbf{Consistency}: If this persona has existing preferences for similar dimensions, maintain consistency with their established patterns
    \item \textbf{Transferability}: Consider how preferences might transfer between problem types. For example:
    \begin{itemize}
        \item Visual preferences might transfer well between geometry and coding
        \item Communication tone preferences should be highly consistent across problems
        \item Detail level preferences might vary based on problem complexity
    \end{itemize}
    \item \textbf{Persona-Specific}: Ground all preferences in this persona's characteristics and background
    \item \textbf{Problem-Relevant}: Adjust importance and some values based on this specific problem type
    \item \textbf{Justification}: Explain how you considered existing preferences and problem context
\end{enumerate}

For each dimension, provide:
\begin{enumerate}
    \item \textbf{Value}: Specific value that fits this persona's profile and is consistent with existing preferences
    \item \textbf{Local Importance}: 1-5 scale of how much this persona cares about this aspect for THIS specific problem. Make sure to spread out the range of importance; for example you cannot say all dimensions are important.
    \item \textbf{Justification}: Detailed explanation of your reasoning, including reference to existing preferences if relevant
\end{enumerate}

\textbf{Format as JSON}:

\texttt{\{
  "preferences": \{
    "Dimension Name": \{
      "name": "Copy the name of the preference dimension from the previous step",
      "description": "Copy the description of the preference dimension from the previous step",
      "value\_range": "Copy the value range of the preference dimension from the previous step",
      "value": "specific value or number",
      "local\_importance": 1-5,
      "justification": "Detailed reasoning considering persona characteristics, existing preferences, and problem context",
      "type": "expertise" or "personal"
    \}
  \}
\}}
\end{tcolorbox}

\subsection*{Evaluation Rubric Generator (Expertise Criteria)}

\begin{tcolorbox}[colback=teal!5!white, colframe=teal!75!black, title=Expertise Rubric Generation Prompt, breakable]
You are creating evaluation criteria for EXPERTISE/COMFORT preferences---how well an AI response matches a user's technical skill level and comfort with specific concepts.

\textbf{PROBLEM TO BE ANSWERED}:

Type: \texttt{\{problem\_type\}}

Problem: \texttt{\{problem\_text\}}

[\textit{If answer is available}] Correct answer: \texttt{\{solution\_context\}}

\textbf{PERSONA CONTEXT}:

User: \texttt{\{persona\_name\}} - \texttt{\{minimal\_necessary\_description\}}

Background: \texttt{\{backstory\}}

\textbf{EXPERTISE/COMFORT PREFERENCES}: \texttt{\{expertise\_preferences\_with\_weights\}}

\textbf{EXPERTISE RUBRIC GUIDELINES}:
\begin{itemize}
    \item Focus on TECHNICAL APPROPRIATENESS: Does the response match the user's stated comfort/skill level?
    \item Evaluate CONCEPT COMPLEXITY: Are concepts presented at the right level of sophistication?
    \item Assess TERMINOLOGY USAGE: Is technical language appropriate for their experience level?
    \item Check ASSUMPTION LEVELS: Does the response assume the right background knowledge?
\end{itemize}

\textbf{CRITICAL}: Create exactly ONE criterion for each expertise preference listed above. Each criterion must use the EXACT preference name.

For each expertise preference, create criteria that measure how well the AI response calibrates its technical level to match the user's capabilities for this specific \texttt{\{problem\_type\}}.

\textbf{EXAMPLE EXPERTISE CRITERION}:

\texttt{\{
  "preference": "Comfort with Linear Algebra",
  "description": "How well the response matches the user's intermediate comfort level (value: 3) with linear algebra when explaining matrix operations in this problem",
  "weight": 0.25,
  "levels": [
    \{"score": 1, "description": "Uses linear algebra concepts far above (graduate level) or below (basic arithmetic) the user's intermediate level, making explanation inaccessible or patronizing"\},
    \{"score": 3, "description": "Mostly appropriate for intermediate level but some inconsistencies---occasionally too advanced or too basic for stated comfort level"\},
    \{"score": 5, "description": "Perfectly calibrated to intermediate linear algebra comfort---uses appropriate terminology, assumes right background knowledge, explains concepts at ideal complexity level"\}
  ]
\}}

\textbf{Format as JSON}:

\texttt{\{
  "criteria": [
    \{
      "preference": "Exact preference name",
      "description": "How well the response matches the user's technical level for this specific concept in this problem",
      "weight": 0.XX,
      "levels": [
        \{"score": 1, "description": "Technical level far above or below user's stated comfort level"\},
        \{"score": 3, "description": "Mostly appropriate technical level but some inconsistencies"\},
        \{"score": 5, "description": "Perfectly calibrated to user's stated comfort level"\}
      ]
    \}
  ]
\}}
\end{tcolorbox}

\subsection*{Evaluation Rubric Generator (Personal Criteria)}

\begin{tcolorbox}[colback=pink!5!white, colframe=pink!75!black, title=Personal Rubric Generation Prompt, breakable]
You are creating evaluation criteria for PERSONAL preferences---how well an AI response adapts to a user's unique learning style, communication preferences, and personal context.

\textbf{PROBLEM TO BE ANSWERED}:

Type: \texttt{\{problem\_type\}}

Problem: \texttt{\{problem\_text\}}

[\textit{If answer is available}] Correct answer: \texttt{\{solution\_context\}}

\textbf{PERSONA CONTEXT}:

User: \texttt{\{persona\_name\}} - \texttt{\{minimal\_necessary\_description\}}

Background: \texttt{\{backstory\}}

\textbf{PERSONAL PREFERENCES}: \texttt{\{personal\_preferences\_with\_weights\}}

\textbf{PERSONAL RUBRIC GUIDELINES}:
\begin{itemize}
    \item Focus on LEARNING STYLE ADAPTATION: Does the response match how this person prefers to learn?
    \item Evaluate COMMUNICATION STYLE: Is the tone, formality, and interaction style appropriate?
    \item Assess CONTEXTUAL RELEVANCE: Are examples, analogies, and references meaningful to this person?
    \item Check MOTIVATIONAL ELEMENTS: Does the response connect to what drives this person?
\end{itemize}

Consider this person's background: \texttt{\{occupation\}}, hobbies: \texttt{\{hobbies\}}, and learning style: \texttt{\{learning\_style\}}

\textbf{CRITICAL}: Create exactly ONE criterion for each personal preference listed above. Each criterion must use the EXACT preference name.

For each personal preference, create criteria that measure how well the AI response adapts to this specific person's unique learning needs and preferences for this \texttt{\{problem\_type\}}.

\textbf{EXAMPLE PERSONAL CRITERION}:

\texttt{\{
  "preference": "Engineering Examples from Work Experience",
  "description": "How well the response incorporates relevant engineering examples that connect to this user's mechanical engineering background when explaining this physics problem",
  "weight": 0.20,
  "levels": [
    \{"score": 1, "description": "No engineering examples provided, or examples from completely irrelevant fields that don't connect to user's mechanical engineering experience"\},
    \{"score": 3, "description": "Includes some engineering examples but they're generic or don't clearly connect to the physics concepts in this specific problem"\},
    \{"score": 5, "description": "Provides multiple relevant mechanical engineering examples that directly illuminate the physics concepts and clearly resonate with the user's professional experience"\}
  ]
\}}

\textbf{Format as JSON}:

\texttt{\{
  "criteria": [
    \{
      "preference": "Exact preference name",
      "description": "How well the response adapts to this user's specific personal learning preference for this problem",
      "weight": 0.XX,
      "levels": [
        \{"score": 1, "description": "No adaptation to personal preference or completely inappropriate approach"\},
        \{"score": 3, "description": "Some adaptation to personal preference but generic or inconsistent"\},
        \{"score": 5, "description": "Excellent adaptation that perfectly matches the user's personal preference"\}
      ]
    \}
  ]
\}}
\end{tcolorbox}

\subsection{LLM Judge Scoring Prompt}
\label{app:judge_prompts}

The judge evaluates each rubric criterion independently using the following prompts.

\begin{tcolorbox}[
    title=System Message,
    colback=gray!5,
    colframe=gray!50,
    fonttitle=\bfseries,
    breakable
]
You evaluate responses against specific criteria. Output valid JSON.
\end{tcolorbox}

\begin{tcolorbox}[
    title=User Prompt,
    colback=blue!3,
    colframe=blue!30,
    fonttitle=\bfseries,
    breakable
]
You are an expert evaluation specialist using a standardized rubric to assess personalized responses. Provide a precise, evidence-based assessment of how well this response meets the specified criterion.

\medskip
\textbf{CRITERION TO EVALUATE:}\\
\texttt{\{criterion name\}}\\
Description: \texttt{\{criterion description\}}

\medskip
\textbf{PERFORMANCE LEVELS:}\\
\texttt{\{JSON list of 5 rubric levels with descriptions\}}

\medskip
\textbf{RESPONSE TO EVALUATE:}\\
``\texttt{\{model's final response\}}''

\medskip
\textbf{USER CONTEXT:}\\
Preference value: \texttt{\{user's preferred value on 1--5 scale\}}\\
Context: \texttt{\{justification for why this level suits the user\}}

\medskip
\textbf{EVALUATION INSTRUCTIONS:}
\begin{enumerate}
    \item Carefully compare the response against each performance level description.
    \item Identify specific evidence in the response that matches level descriptions.
    \item Determine the exact score (1--5) that best represents the response quality for this criterion.
    \item Provide a detailed justification referencing specific elements of the response.
    \item Be objective and consistent---apply the same standards across all evaluations.
    \item Consider the user's stated preferences.
    \item Provide precise reasoning that another evaluator could follow to reach the same conclusion.
\end{enumerate}

\medskip
Format your evaluation as a JSON object:\\
\texttt{\{"score": X (integer between 1 and 5), "justification": "..."\}}
\end{tcolorbox}

\paragraph{Rubric-Level Scoring.}
The rubric performance levels are personalized at benchmark-construction time to each user's preferred value $v_j$. For instance, for a user who prefers ``Terminology Complexity'' $= 2$ (simple language), the rubric levels are defined as:

\begin{tcolorbox}[
    title={Example: Personalized Rubric Levels for Terminology Complexity ($v_j = 2$)},
    colback=green!3,
    colframe=green!30,
    fonttitle=\bfseries\small,
    breakable
]
\small
\textbf{Score 1:} Uses highly technical jargon far above the user's preferred simple language level, making the explanation inaccessible.\\
\textbf{Score 3:} Mostly uses accessible language but occasionally introduces unnecessary technical terms, inconsistently matching the user's preference for simplicity.\\
\textbf{Score 5:} Uses plain, accessible terminology throughout and avoids unnecessary jargon, matching the user's preference for simple language.
\end{tcolorbox}

\noindent The judge receives these pre-personalized level descriptions and matches the response against them. The grading function $g_j(r, v_j)$ (Eq.~\ref{eq:prefalign}) therefore measures how well the response matches the user's stated preference level for each attribute.

\end{document}

%% file: math_commands.tex
\usepackage{amsmath,amsfonts,bm}

\def\eqref#1{equation~\ref{#1}}

\def\1{\bm{1}}

\DeclareMathAlphabet{\mathsfit}{\encodingdefault}{\sfdefault}{m}{sl}
\SetMathAlphabet{\mathsfit}{bold}{\encodingdefault}{\sfdefault}{bx}{n}